\documentclass[sigconf]{acmart}
\usepackage{popets}

\usepackage{tikz}
\usetikzlibrary{positioning}
\usepackage{graphicx}

\usepackage{algorithm} 
\usepackage{algpseudocode} 

\newcommand{\tWaKA}{\text{t-WaKA}} 
\newcommand{\WaKA}{\text{WaKA}}
\newcommand{\DSV}{\text{DSV}}

\newcommand{\WaKAadd}{\operatorname{WaKA}_{\mathrm{add}}}
\newcommand{\WaKArem}{\operatorname{WaKA}_{\mathrm{rem}}}
\newtheorem{definition}{Definition}

\setcopyright{popets}
\copyrightyear{YYYY}

\acmYear{YYYY}
\acmVolume{YYYY}
\acmNumber{X}
\acmDOI{XXXXXXX.XXXXXXX}
\acmISBN{}
\acmConference{Proceedings on Privacy Enhancing Technologies}
\begin{document}

\title[WaKA: Data Attribution using K-Nearest Neighbors and Membership Privacy Principles]{WaKA: Data Attribution using K-Nearest Neighbors and Membership Privacy Principles}


\author{Patrick Mesana}
\affiliation{%
  \institution{HEC Montréal}
  \city{Montréal}
  \state{}
  \country{Canada}}
\email{patrick.mesana@hec.ca}

\author{Clément Bénesse}
\affiliation{%
  \institution{University du Québec à Montréal}
  \city{Montréal}
  \state{}
  \country{Canada}}
\email{clement.benesse@gmail.com}

\author{Hadrien Lautraite}
\affiliation{%
  \institution{University du Québec à Montréal}
  \city{Montréal}
  \state{}
  \country{Canada}}
\email{lautraite.hadrien@courrier.uqam.ca}

\author{Gilles Caporossi}
\affiliation{%
  \institution{HEC Montréal}
  \city{Montréal}
  \state{}
  \country{Canada}}
\email{gilles.caporossi@hec.ca}

\author{Sébastien Gambs}
\affiliation{%
  \institution{University du Québec à Montréal}
  \city{Montréal}
  \state{}
  \country{Canada}}
\email{gambs.sebastien@uqam.ca}


\renewcommand{\shortauthors}{Mesana et al.}

\begin{abstract}
In this paper, we introduce WaKA (\emph{Wasserstein K-nearest neighbors Attribution}), a novel attribution method that leverages principles from the LiRA (\emph{Likelihood Ratio Attack}) framework and $k$-nearest neighbors classifiers ($k$-NN). 
WaKA efficiently measures the contribution of individual data points to the model’s loss distribution, analyzing every possible $k$-NN that can be constructed using the training set, without requiring to sample subsets of the training set. 
WaKA is versatile and can be used \emph{a posteriori} as a membership inference attack (MIA) to assess privacy risks or \emph{a priori} for privacy influence measurement and data valuation.
Thus, WaKA can be seen as bridging the gap between data attribution and membership inference attack (MIA) by providing a unified framework to distinguish between a data point’s value and its privacy risk.
For instance, we have shown that self-attribution values are more strongly correlated with the attack success rate than the contribution of a point to the model generalization. 
WaKA’s different usages were also evaluated across diverse real-world datasets, demonstrating performance very close to LiRA when used as an MIA on $k$-NN classifiers, but with greater computational efficiency.
Additionally, WaKA shows greater robustness than Shapley Values for data minimization tasks (removal or addition) on imbalanced datasets.
\end{abstract}



\maketitle

\section{Introduction}
Data attribution methods have been developed originally to measure the contribution of individual data points in a training set to a model’s output.
These methods can serve different purposes depending on the context.
One key application is data valuation, in which the objective is to quantify the ``value'' of each data point with respect to its impact on the model’s ability to generalize. 
For example, Data Shapley Value (DSV), introduced in~\cite{ghorbani2019data} and~\cite{jia2019efficient}, is grounded in the game-theoretic Shapley Value framework and is often used for tasks such as data minimization via summarization, in which the objective is to remove a large fraction of data points while ensuring a high generalization performance of the model 
This approach aligns well with Article 5 of the General Data Protection Regulation (GDPR)~\cite{gdpr-info}, which emphasizes that personal data should be ``adequate, relevant, and limited to what is necessary'' in relation to the purposes for which they are processed.

\begin{figure*}[h!]
    \centering
    \includegraphics[width=10cm]{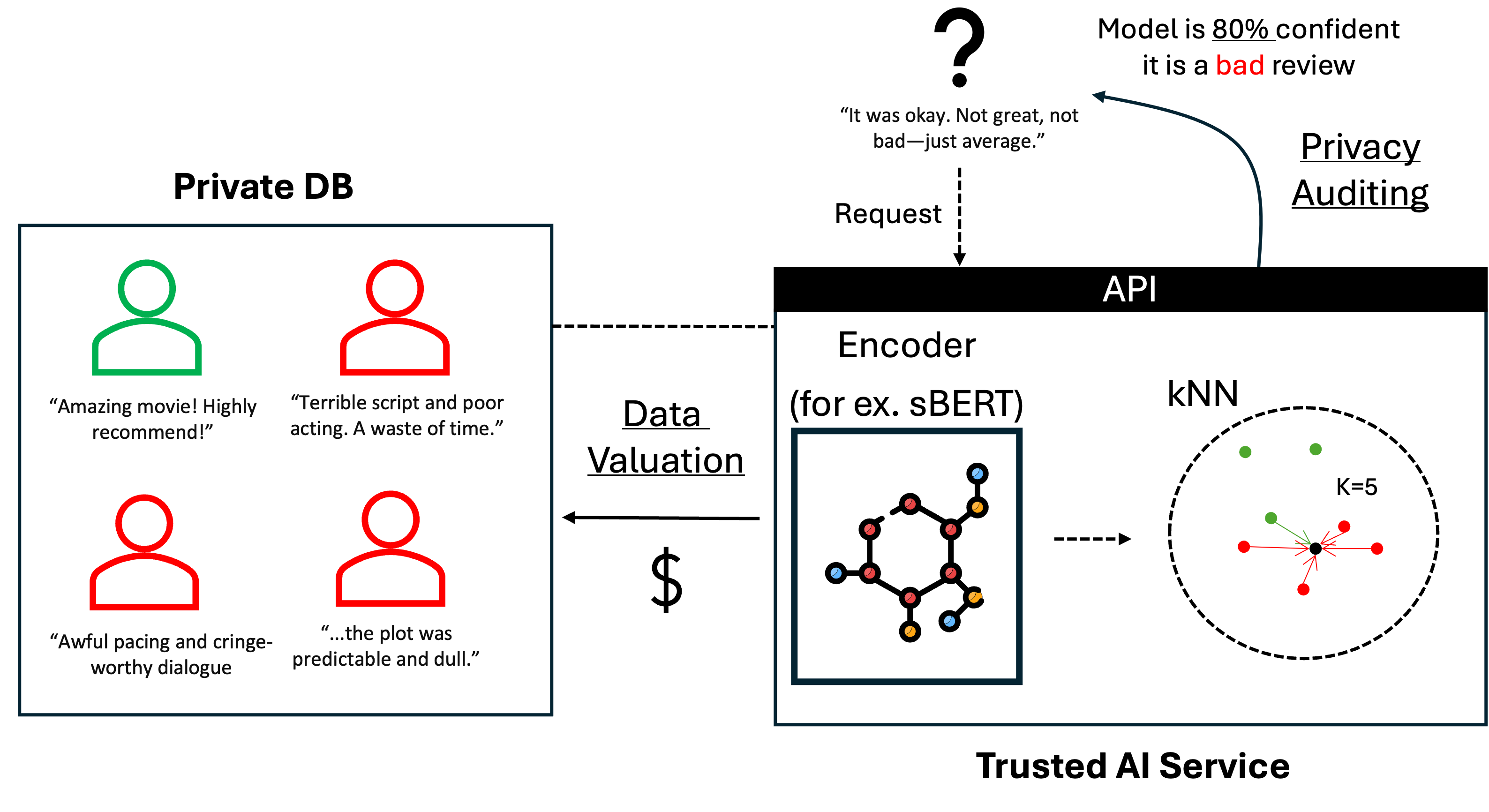} 
    \caption{Illustration of data attribution in a movie review classification scenario, highlighting the dual perspectives of data valuation (estimating data point value) and data privacy (measuring potential information leakage).}
    \label{fig:waka-perspectives}
\end{figure*}

As a motivating scenario, consider for instance an organization that collects a dataset and then trains a machine learning model over it, exposing its functionality via an API to monetize the access to the predictions of the model (\emph{e.g.}, for classifying movie reviews  or categorizing images). 
Data attribution can be used by this organization from two perspectives: data valuation and privacy with Figure~\ref{fig:waka-perspectives} illustrating both viewpoints. 
On one hand, data valuation can help estimate the contribution of each data point to the model, guiding decisions on redistributing a share of the model's value to individuals who provided data points. 
This valuation, which aligns with Shapley value, also helps the organization determine which data points bring no value or even negative value, potentially avoiding unnecessary costs by not acquiring these data points.

On the other hand, from the privacy perspective, the organization might be concerned with how much information about data points could potentially be leaked through the API, which is akin to measuring the privacy risk of data leakage. 
This is closely related to membership inference attacks (MIAs), in which an attacker aims to determine whether a particular data point was part of the model's training set. 
While data valuation and privacy concerns are often related, they are usually addressed through very different methods. 
To solve this issue, we introduce WaKA (for 1-Wasserstein $k$-NN Attribution), which provides a unified framework for addressing both aspects using $k$-nearest neighbors ($k$-NN) models.

While being simple in their design, $k$-NN models offer a clear and intuitive way to perform data attribution. 
In particular, they are recognized as an instance-based explainable method~\cite{molnar2020interpretable}, meaning its predictions are directly influenced by the training data without an explicit model abstraction.  
However, a well-known limitation is that they do not perform well on high-dimensional data such as textual documents or images due to the curse of dimensionality. 
In such cases, the distance between points becomes nearly identical, making the identification of neighbors ineffective~\cite{friedman1997bias}. 
To overcome this challenge and apply them to datasets like images and textual data, we first employ a neural network to learn embeddings suitable for $k$-NN classification. 
These learned embeddings capture meaningful representations of the data, making $k$-NN effective even in high-dimensional settings. 
This approach is also commonly used in the data valuation literature in which case it is usually referred to as ``surrogate models''~\cite{jia2021scalability}. 
Additionally, while the majority of research on data attribution and membership inference attacks has been centered on neural networks, $k$-NN-based studies are highly relevant in practice. 
In particular, in industrial applications, such as retrieval-augmented generation (RAG) pipelines, $k$-NN with embeddings is widely adopted as the use of learned representations mitigates the high-dimensionality challenges typically associated with nearest neighbor search.
Moreover, ~\cite{yadav2021behavior} demonstrated that interpretations of $k$-NN models using embeddings are comparable to a softmax layer in neural networks, reinforcing their relevance to modern machine learning workflows.

The term ``value'' in data valuation is semantically charged, as it suggests an intrinsic worth of data points. 
In this context, the term is grounded in the Shapley Value axioms, which provide a unique way of attributing the contribution of each data point to the model’s generalization performance.
Data valuation methods such as DSV seek to uncover the intrinsic properties of data in relation to the category of model being used, whether it be a type of neural network, a $k$-NN classifier or another machine learning model.

In contrast, membership inference ~\cite{shokri2017membership,yeom2018privacy} aims at determining whether a given data point was part of a specific model's training dataset.
For instance, LiRA~\cite{carlini2022membership} is a state-of-the-art approach for performing membership inference attacks (MIAs), which is based on the Likelihood Ratio Test (LRT), a statistical test that compares the likelihood of a model's prediction for a given data point when trained with and without this point. 
To realize this, LiRA requires the training of shadow models via sampling and provides a membership score for each point in the training set, facilitating a more detailed attribution analysis. 
Similarly to LiRA, WaKA assumes that the adversary has access to the underlying data distribution, enabling probability mass function (PMF) computation. 
This assumption is fundamental to our framework but also to LiRA.

\textbf{Related work.}
More recently, the intersection between data attribution and membership privacy has garnered significant attention as evidenced by a growing body of related literature~\cite{duddu2021shapr, ye2023leave, cohen2024membership}. 
One key concept is ``self-influence'', which has been investigated in differentiable models to measure the extent to which a data point influences its own prediction. 
This concept is particularly relevant in MIAs, as high self-influence scores often correlate with increased privacy risks~\cite{cohen2024membership}. 
Self-influence is computed using influence functions and measures how much the loss changes for a data point when it is upweighted. 
In particular, this method has been used for capturing how a point’s inclusion can lead to memorization~\cite{liu2021understanding}, which is in turn relates to privacy vulnerability. 
Throughout the paper, we adopt a similar notion, which we refer to as ``self-attribution'', whose objective is to address the question ``To what extent my data contribute to my own outcome?''. 
More precisely, it can be quantified by the marginal contribution of a point to the model’s prediction on that same point.

Beyond self-influence, several studies have analyzed how the model's performance relates to privacy risk. 
Prior work has shown that overfitting exacerbates MIAs~\cite{shokri2017membership, yeom2018privacy}, but recent findings indicate that privacy leakage can occur even in well-generalized models~\cite{nasr2019comprehensive, carlini2022membership}.
Additionally, per-record memorization suggests that not all training samples contribute equally to privacy risks—some points are more prone to memorization and thus more vulnerable to MIAs~\cite{carlini2019secret,feldman2020neural}. 
This suggests a complex relationship between the value of a data point and its privacy risk.
Recent work on Leave-One-Out Distinguishability (LOOD)~\cite{ye2023leave} provides a unified perspective on data attribution and privacy auditing by quantifying the statistical distance between a model’s outputs with and without a specific training point. 
This approach highlights the strong connection between influence, memorization and privacy leakage. 
While LOOD has been shown to predict MIA success and can serve as a privacy auditing tool, existing methods primarily focus on neural network-based models.

\textbf{Summary of contributions.}
Our main contributions can be summarized as follows. 
\begin{itemize}
\item We introduce the 1-Wasserstein $k$-NN Attribution (WaKA), a novel approach that leverages the Wasserstein distance from Optimal Transport, which has not been previously used in either membership inference attacks (MIAs) or data valuation. 
Unlike LiRA, which relies on hypothesis testing, WaKA accounts for the mass that a point moves positively or negatively. 
WaKA serves a dual purpose by providing a principled framework for both privacy risk assessment and data valuation, functioning as a general attribution method for data valuation while also offering privacy insights through self-attribution.
More precisely, it can be adapted into t-WaKA to effectively perform membership inference attacks.
\item In our experiments, we compare the performance of DSV and WaKA.
These experiments have been conducted on six diverse datasets—two tabular datasets (Adult and Bank), two textual datasets (IMDB and Yelp) and two image datasets (CIFAR-10 and CelebA) —to demonstrate the versatility and robustness of our method. 
The evaluation of these scenarios focuses on two key aspects: utility, through two data minimization tasks 
as well as privacy, by measuring the attack success rate (ASR) on all training points.
\item We explore the ``onion effect''~\cite{carlini2022privacy}, a phenomenon observed previously in neural networks in which removing data points incrementally reveals deeper layers of vulnerable privacy points in the sense that these points suffer from a higher ASR after the removal. 
To investigate this effect, we have replicated some of experiments of the original paper by eliminating 10\% of the training set using attribution methods, followed by a reassessment of privacy scores. 
More precisely, we have analyzed the relationship between privacy influences and WaKA influences, showing that they are correlated and can be used to predict, \emph{a priori}, whether removing a data point will impact the ASR on other points. 
\item Finally, we have also conducted experiments using t-WaKA as an MIA on specific training points for $k$-NN models. 
t-WaKA displays a similar performance as LiRA but uses significantly less resources as it relies on a single reusable $k$-NN model trained on the entire dataset, thus avoiding the need for shadow models. 
More precisely, once this $k$-NN model is trained, t-WaKA has a computational complexity of $O(\log N)$ for attacking a specific point, which is much faster than LiRA for $k$-NN.
\end{itemize}

\textbf{Outline.}
First in Section~\ref{sec_background}, we provide a brief overview of attribution methods (Leave one Out and Data Shapeley Value) for $k$-NN models as well as LiRA, before introducing the details of the Wasserstein k-NN Attribution (WaKA) method in Section~\ref{sec:waka}.
Afterwards, in Section~\ref{sec:exp}, we conduct an extensive evaluation of WaKA as a new attribution method and t-WaKA for assessing the success of MIAs before finally concluding with a discussion in Section~\ref{sec:discussion}.

\section{Background}
\label{sec_background}

In this section, we review the background notions necessary to the understanding of our work, namely the main existing attribution methods for \(k\)-NNs as well as the LiRA framework for conducting membership inference attacks.

\subsection{Attribution Methods for \(k\)-NNs}
\label{sec_loo_knn}

\textbf{Leave-One-Out (LOO)} attribution methods are commonly used to assess the contribution of individual data points to a model's performance by examining the effect of removing a point from the training set.
Although they can theoretically be applied with respect to any predictive target, they are typically computed on a test set \( D_{\text{test}} \) with the objective of measuring the contribution to the generalization performance. 
In this context, a positive contribution means reducing the generalization loss or increasing utility~\cite{jia2021scalability}. 

Consider a training set \( D = \{z_i\}_{i=1}^{N} \), in which each \( z_i \) represents a feature-label pair \( (x_i, y_i) \).
For \(k\)-NN models, the loss function $\ell$ for any data point \(z\) is generally defined as the fraction of neighbors that do not share the same label:
\begin{equation}
\ell(z; D, k) = 1 - \frac{1}{k} \sum_{j=1}^{k} \mathbf{1}(y_{\alpha_j} = y),
\end{equation}
in which $\alpha_j$ is the index of the \(j\)-th closest neighbor to \(z\) in the training set \(D\), and \(y\) is the true label of point \(z\). 
This is referred to as a ``loss'' because it measures the error introduced by the model's prediction, which also quantifies the model's disagreement with the ground truth. 
The utility function \( U(z_t; D) \) can be expressed as:
\[
U(z_t; D) = 1 - \ell(z_t; D) = \frac{1}{k} \sum_{j=1}^{k} \mathbf{1}(y_{\alpha_j} = y_t).
\]
Here, we evaluate utility using \(z_t\), which we refer to as a test point, because utility typically reflects the performance of the \(k\)-NN model on data points outside the training set. 
This approach aligns with the goal of assessing the model's ability to generalize to unseen data, ensuring that its utility is not biased by the training data.

The LOO attribution method computes the difference in utility with respect to a test point, with and without the point \( z_i \). 
Formally, the LOO attribution score for a point \( z_i \) in \(k\)-NN is given by:
\[
v_{\text{loo}}(z_i) = U(z_t; D, k) - U(z_t; D_{-z_i}, k),
\]
in which, \( D_{-z_i} \) represents the training set excluding \( z_i \). 
If we want to compute the LOO value for each \( z_i \) in \( D \), this approach requires evaluating \( N \) distinct neighborhood configurations.

\textbf{Data Shapley Value.} The Shapley Value, a concept from cooperative game theory, extends LOO by averaging the marginal contribution of each point across all possible subsets of the training set \cite{ghorbani2019data,jia2019efficient}.
The DSV is computed as the average contribution of each point \( z_i \) to the model’s utility across subsets \( S \subseteq D \):
\[
v_{\text{shap}}(z_i) = \frac{1}{N!} \sum_{S \subseteq D \setminus \{z_i\}} \frac{1}{\binom{N-1}{|S|}}  U(z_t; S \cup \{z_i\}) - U(z_t; S).
\]
The Shapley value satisfies four key axioms: \emph{Efficiency} (\emph{i.e.}, the total value is distributed among all players), \emph{Symmetry} (\emph{i.e.}, identical contributions are rewarded equally), \emph{Dummy} (\emph{i.e.}, players that contribute nothing receive zero value), and \emph{Linearity} (\emph{i.e.}, the Shapley values from two games can be combined linearly).

While this formulation provides an axiomatic and robust way of quantifying the importance of each data point, the computational complexity of directly computing Shapley values can be as high as \( O(2^N) \), making it infeasible for large datasets. 
Nonetheless, for \(k\)-NN classifiers, an exact formulation of the Shapley value exists, as shown by \cite{jia2019efficient}, which drastically reduces the complexity to \(O(N \log N)\).
More precisely, the exact Shapley value for \(k\)-NN can be computed as follows. 
For the farthest neighbor \(z_{\alpha_N}\), the Shapley value is:
\[
v_{\text{shap}}(z_{\alpha_N}) = \frac{1[y_{\alpha_N} = y_t]}{N}.
\]
Afterwards, for the remaining neighbors \( z_{\alpha_i} \) with \( i < N \), the Shapley value can be recursively computed as:

\[
v_{\text{shap}}(z_{\alpha_i}) = v_{\text{shap}}(z_{\alpha_{i+1}}) + \frac{1[y_{\alpha_i} = y_t] - 1[y_{\alpha_{i+1}} = y_t]}{k} \cdot \frac{\min(k, i)}{i}.
\]

By exploiting the structure of \(k\)-NN classifiers, this exact formulation avoids the need to evaluate individually all subsets, significantly reducing the computational cost.

Note that these attribution methods are agnostic to any specific model, unlike LiRA, which evaluates the impact of a point on a particular trained model. 
In contrast, DSV implicitly considers all possible \(k\)-NN models trained on subsets of the dataset, offering a more comprehensive measure of point importance across model variations.


\subsection{Likelihood Ratio Attack (LiRA)}

LiRA is a method for performing MIAs, which leverages a Likelihood Ratio Test (LRT) to compare the likelihood of a model when trained with and without a specific data point $z_i$~\cite{carlini2022membership}. 
More precisely for a given model $f$, the LiRA score for $z_i$ is defined as:
\begin{equation}
\Lambda(f; z_i) = \frac{P(f \mid \mathbb{Q})}{P(f \mid \mathbb{Q}_{-z_i})},
\end{equation}
in which $\mathbb{Q}$ and $\mathbb{Q}_{-z_i}$ represent the distributions of models trained on datasets that respectively include or exclude a training point $z_i$. 
By ``distributions of models'' we mean the set of possible models that can be obtained through training on different subsets of data. 
For instance for a neural network, this distribution refers to the space of model weights while for $k$-NN classifiers, the distribution of models corresponds to the distribution of subsets of $k$ data points from the training set $D$.
While theoretically, the number of unique possible combinations is $\binom{N}{k}$, in practice, combinations including the closest neighbors to a test point are more likely. 
We assume that the attacker does not have access to the training set and can only observe the final prediction or the loss of the model, similar to the adversary model used by~\cite{carlini2022membership}.

To compute the LiRA score with respect to the loss, the LRT becomes a one-dimensional statistic:
\begin{equation}
\Lambda(\ell; z_i) = \frac{P(\ell(z_i) \mid \mathbb{Q})}{P(\ell(z_i) \mid \mathbb{Q}_{z_-i})}.
\end{equation}

Estimating this ratio requires sampling various subsets of the training data, which can be computationally expensive. 
This motivated the development of our attribution method specifically designed for $k$-NNs, named WaKA. 
More precisely, WaKA relies on the same principle as LiRA, focusing on comparing the loss distributions for members and non-members of the training dataset. However, instead of relying on a statistical test to distinguish these distributions, WaKA measures the 1-Wasserstein distance between them, providing a more flexible and computationally efficient approach for $k$-NNs.
\section{Wasserstein $k$-NN Attribution (WaKA)}
\label{sec:waka}

\subsection{Attribution using 1-Wasserstein Distance}

In our approach, we quantify the importance of each data point in a \(k\)-NN model by measuring how the inclusion or exclusion of a data point affects the distribution of the model’s loss relative to a specific point \(z_t\).
Here, \(z_t\) represents the point at which we evaluate the model's loss. In the context of data valuation, \(z_t\) is typically a test point for which we aim to assess the influence of training data points on the model's performance. 
In the context of privacy, \(z_t\) could be the data point \(z_i\) itself, allowing us to analyze how the inclusion or exclusion of \(z_i\) affects the loss distribution relative to \(z_i\). 
To achieve this, we leverage the 1-Wasserstein distance \cite{Peyre_Cuturi_2018} between the loss distributions with and without the data point. More formally, the 1-Wasserstein distance (also known as the Earth Mover's Distance) between two probability distributions \(\mu\) and \(\nu\) can be defined as:
\[
W_1(\mu, \nu) = \int_{-\infty}^{\infty} \left| F_{\mu}(x) - F_{\nu}(x) \right| \, dx
\]
in which \(F_{\mu}\) and \(F_{\nu}\) are respectively the cumulative distribution functions (CDFs) of \(\mu\) and \(\nu\).

Let \(\mathbb{L}\) denote the distribution of loss values relative to \(z_t\) using \(k\)-NN models trained on all possible subsets of \(D\), and \(\mathbb{L}_{-z_i}\) refer to the corresponding distribution when \(z_i\) is excluded from \(D\).
Our goal is to compute the 1-Wasserstein distance between these two distributions to assess the impact of data point \(z_i\). 
Since the loss values in a \(k\)-NN classifier are discrete and belong to the finite set \(\mathcal{L} = \left\{ 0, \frac{1}{k}, \frac{2}{k}, \dots, 1 \right\}\), the 1-Wasserstein distance can be computed directly by summing the absolute difference of cumulative distribution functions (CDFs) of the loss distributions.

We define \(\mathcal{F} := \{\text{\(k\)-NN trained on subsets of } D\}\) and \(\mathcal{F}_{-z_i} := \{\text{\(k\)-NN trained on subsets of } D \setminus \{z_i\}\}\), the two spaces of models that correspond to all \(k\)-NN models trained with or without the point \(z_i\).
In this setting, let \(\mathbb{Q}\) denote the uniform distribution over models in \(\mathcal{F}\), and \(\mathbb{Q}_{-z_i}\) be the uniform distribution over models in \(\mathcal{F}_{-z_i}\).
Let \(\ell : \mathcal{F} \to \mathbb{R}\) denote the loss function mapping a model \(f \in \mathcal{F}\) to a real-valued loss \(\ell(f)\) evaluated at the point \(z_t\), \emph{i.e.}, \(\ell(f) = \ell(z_t; f)\). The distributions \(\mathbb{L}\) and \(\mathbb{L}_{-z_i}\) represent the pushforward distributions of \(\mathbb{Q}\) and \(\mathbb{Q}_{-z_i}\) through the loss function \(\ell\), denoted with the \(\#\) symbol:
\[
\mathbb{L} = \ell_\# \mathbb{Q}, \quad \mathbb{L}_{-z_i} = \ell_\# \mathbb{Q}_{-z_i}.
\]
More precisely, \(\mathbb{L}\) is the distribution of losses evaluated at \(z_t\) induced by models drawn from \(\mathbb{Q}\), and \(\mathbb{L}_{-z_i}\) is the distribution of losses evaluated at \(z_t\) induced by models drawn from \(\mathbb{Q}_{-z_i}\).

\begin{definition}[WaKA]
We define the 1-Wasserstein \(k\)-NN attribution for data point \(z_i\) relative to any point \(z_t\) as follows:

\begin{align}
W_1(\mathbb{L}, \mathbb{L}_{-z_i}) = \frac{1}{k} \sum_{l_{\text{min}} \leq l \leq l_{\text{max}}} \left| F_{\mathbb{\ell_\#Q}}(l) - F_{\mathbb{\ell_\#Q}_{-z_i}}(l) \right|,
\end{align}
in which \(\mathbb{L}\) and \(\mathbb{L}_{-z_i}\) are the loss distributions relative to \(z_t\), restricted to the range \([l_{\text{min}}, l_{\text{max}}]\). 
The cumulative distribution functions (CDFs) \(F_{\mathbb{\ell_\#Q}}(l)\) and \(F_{\mathbb{\ell_\#Q}_{-z_i}}(l)\) are evaluated for these restricted distributions at the discrete loss values \(l \in \mathcal{L} = \left\{ 0, \frac{1}{k}, \frac{2}{k}, \dots, 1 \right\}\). Here, \(l_{\text{min}} \geq 0\) and \(l_{\text{max}} \leq 1\) specify the bounds of the restricted range.
\end{definition}

 While DSV provides a unique set of values, WaKA introduces flexibility by focusing on how probability masses are moved positively or negatively, relative to particular loss values.
 For instance, when \(z_t\) and \(z_i\) share the same label (\(y_t = y_i\)), WaKA can measure how much loss mass is moved to improve the loss distribution. 
 Conversely, when \(z_t\) and \(z_i\) have different labels (\(y_t \neq y_i\)), WaKA can measure how much \(z_i\) worsens the loss distribution. Additionally, we can refine this analysis by incorporating the decision threshold of the \(k\)-NN classifier.
 Typically, the decision threshold is set to \(1/2\), but in imbalanced datasets, it can be adjusted to favor the minority class. 
 WaKA can identify points skewed positively or negatively relative to the decision threshold, making it especially useful for tasks like data removal or data addition. 
 Figure~\ref{fig:synthetic-exp} illustrates four types of points that WaKA can identify well.

\begin{figure*}[h!]
\centering
\caption{Illustrating the importance of data points in data valuation tasks. The plots highlight: 
(1) \textbf{Bad Outlier} (Top-Left): Outlier point that negatively impact (low Shapley value) one class while offering little benefit on its own (Class 1). 
Important to identify for data removal;
(2) \textbf{Good Outliers} (Top-Right): Outlier points that, despite being in less dense region, improve their class performance (high Shapley value).
(3) \textbf{Good Inliers} (Bottom-Left): Inlier points contributing to loss distributions skewed toward zero, crucial for data addition;
(4) \textbf{Bad Inlier} (Bottom-Right): Not outlier but harm Class 2 if added to dataset. }
    \label{fig:synthetic-exp}
    
    \begin{tikzpicture}[baseline=(current bounding box.north)]
        
        \node[anchor=north west, inner sep=0, outer sep=0] at (0, -3.2) {\includegraphics[width=5.2cm, keepaspectratio=true]{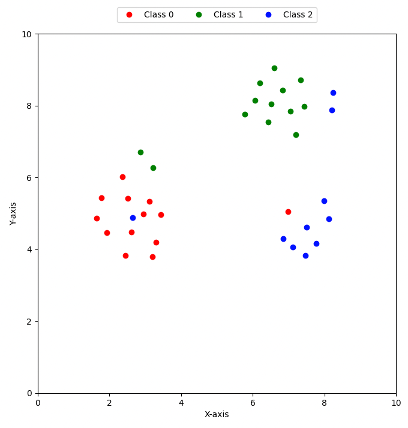}};
        \node[anchor=north] at (2.5, -8.6) {Training Set};
        
        \node[anchor=north west, inner sep=0, outer sep=0] at (6, 0) {\includegraphics[width=5.2cm, keepaspectratio=true]{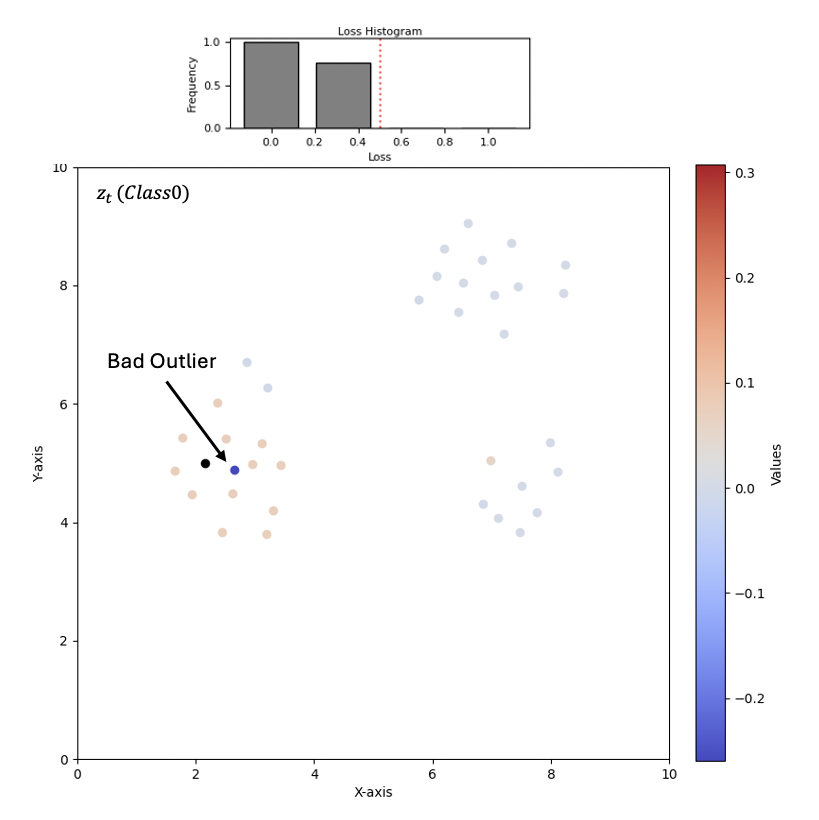}};
        \node[anchor=north] at (8.5, -5.3) {(1) Bad Outlier};
        
        \node[anchor=north west, inner sep=0, outer sep=0] at (11.5, 0) {\includegraphics[width=5.2cm, keepaspectratio=true]{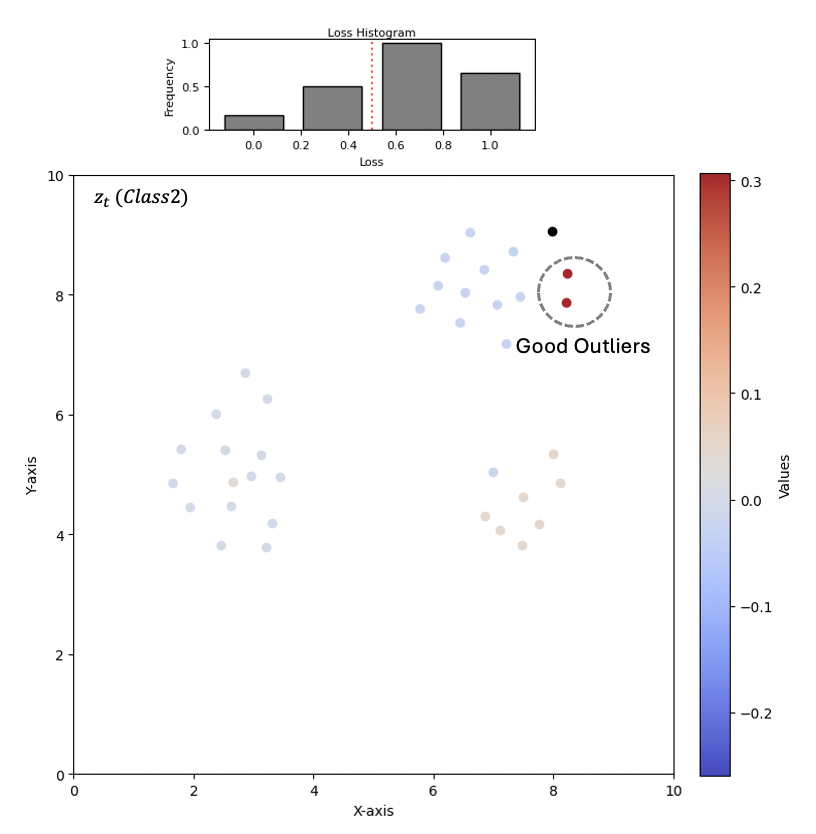}};
        \node[anchor=north] at (14, -5.3) {(2) Good Outliers};
        
        \node[anchor=north west, inner sep=0, outer sep=0] at (6, -6) {\includegraphics[width=5.2cm, keepaspectratio=true]{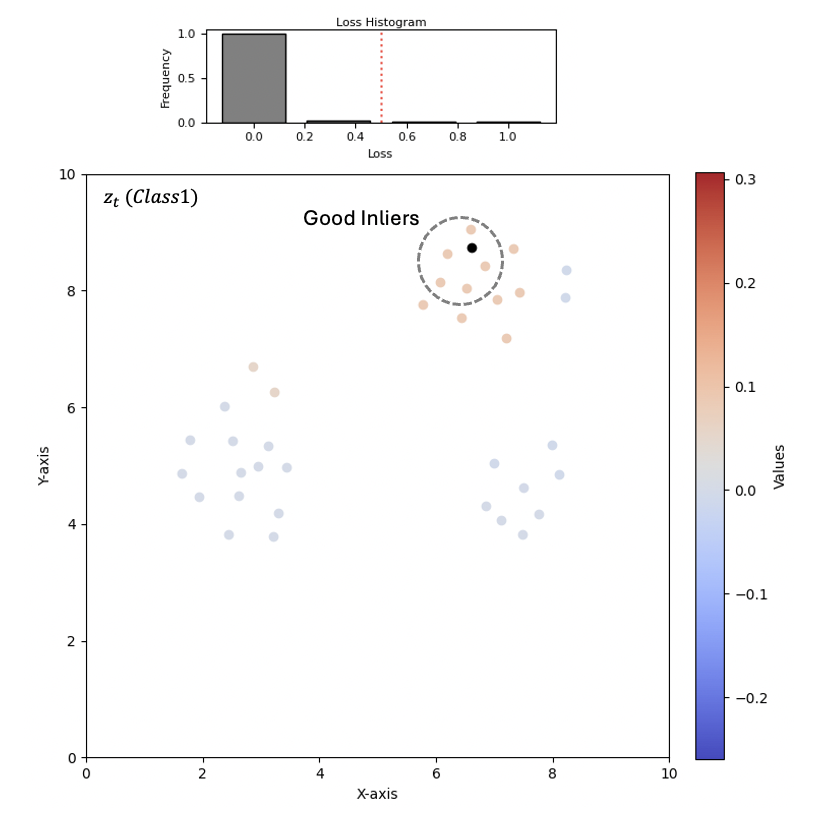}};
        \node[anchor=north] at (8.5, -11.5) {(3) Good Inliers};
        
        \node[anchor=north west, inner sep=0, outer sep=0] at (11.5, -6) {\includegraphics[width=5.2cm, keepaspectratio=true]{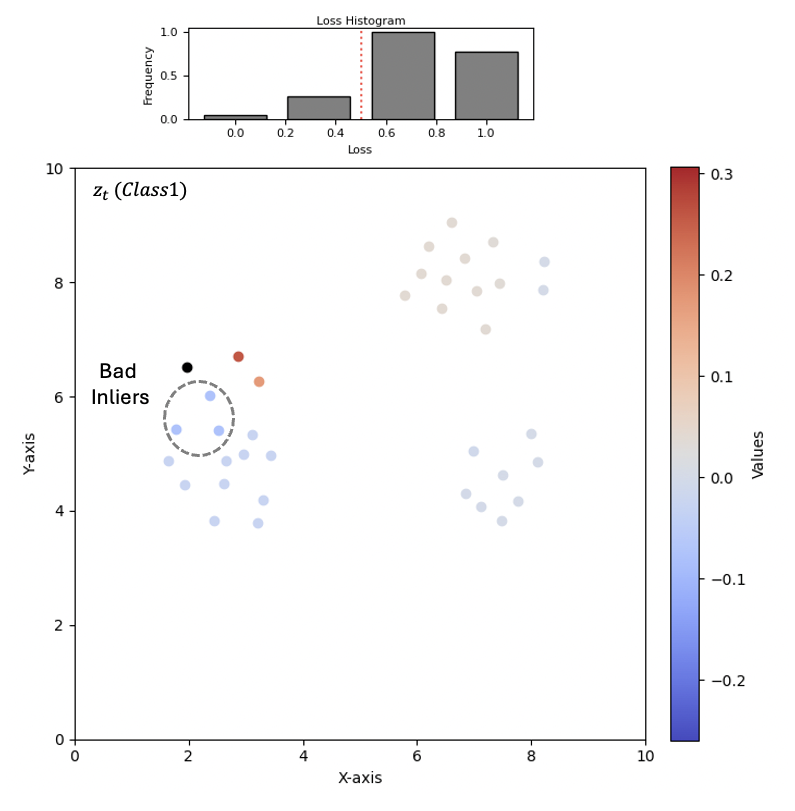}};
        \node[anchor=north] at (14, -11.5) {(4) Bad Inlier};
        
    \end{tikzpicture}
    
\end{figure*}

\textbf{WaKA for data removal and addition.}
To address these data valuation tasks, we propose two formulations of WaKA: one for data removal (\(\text{WaKA}_{\text{rem}}\)) and one for data addition (\(\text{WaKA}_{\text{add}}\)). Both formulations rely on a fixed decision threshold \(\tau \in [0, 1]\), which divides the domain of loss values into two regions.

For data removal, \(\text{WaKA}_{\text{rem}}\) identifies outliers that negatively affect one class while contributing little to other classes. It also accounts positively for points that, despite being outliers, improve their class. The formulation is:

\begin{align*}
\text{WaKA}_{\text{rem}}(z_i)
= & \;\mathbf{1}\bigl(y_{\alpha_i} = y_t\bigr)
\sum_{l \,>\, 1-\tau}
\left|
F_{(1-\ell)\#Q}\bigl(l\bigr)
- F_{(1-\ell)\#Q_{-z_i}}\bigl(l\bigr)
\right|
\\
& \;-\;
\mathbf{1}\bigl(y_{\alpha_i} \neq y_t\bigr)
\sum_{l}
\left|
F_{\ell_\#Q}(l)
- F_{\ell_\#Q_{-z_i}}(l)
\right|.
\end{align*}

For data addition, \(\text{WaKA}_{\text{add}}\) prioritizes inliers but it also penalizes points that are not outliers but negatively affect other classes. The formulation is:

\begin{align*}
\text{WaKA}_{\text{add}}(z_i)
= & \;\mathbf{1}\bigl(y_{\alpha_i} = y_t\bigr)
\sum_{l}
\left|
F_{(1-\ell)\#Q}\bigl(l\bigr)
- F_{(1-\ell)\#Q_{-z_i}}\bigl(l\bigr)
\right|
\\
& \;-\;
\mathbf{1}\bigl(y_{\alpha_i} \neq y_t\bigr)
\sum_{l \,\le\, \tau}
\left|
F_{\ell_\#Q}(l)
- F_{\ell_\#Q_{-z_i}}(l)
\right|.
\end{align*}

These formulations are straightforward applications of WaKA tailored for data removal and data addition tasks.
While other formulations are possible, these provide a simple yet effective approach to prioritizing specific points based on their contributions to the loss distribution. 
In the context of data valuation, \(z_t\) typically refers to a test point; however, when \(z_t = z_i\), we enter what we call self-attribution, in which the focus shifts to understanding the contribution of \(z_i\) to its own loss distribution.
In the following section, we adapt WaKA for membership inference, not only by looking at the loss distribution of \(z_i\) but also by incorporating the loss of a specific model.

\subsection{Adapting WaKA for Membership Inference}

In line with the ``security game'' framework employed in LiRA for evaluating MIAs, the adversary's objective is to ascertain whether a specific point $i$ is part of the training dataset. 
This framework involves an interaction between a challenger and an adversary.
First, the challenger samples a training dataset and trains a model over it. 
Then, depending on the outcome of a private random bit, the challenger sends either a fresh challenge point to the adversary or a point from the training set. 
The adversary, having query access to both the distribution (\emph{i.e.}, in the sense that it can have access to samples for this distribution) and the trained model, must then decide if the given point was part of the training set. 
This structured game can be used to assess the adversary's ability to correctly infer membership in terms of his advantage compared to a random guess, thereby evaluating the robustness of the model against such attacks.

Since we are dealing with $k$-NN models, adding the point \(z_i\) to the training set can only improve the model's performance, meaning that the loss will decrease. 
As such, the distribution \(\mathbb{L}_{-z_i}\) is shifted towards lower values when transitioning to \(\mathbb{L}\). 
However, in the context of the security game, we are given a specific loss value \(\ell(z_i)^*\) for the true model. 
As a result, models with loss greater than \(\ell(z_i)^*\) are incentivized to indicate that \(z_i\) is part of the training set, as the distribution is shifting towards \(\ell(z_i)^*\), while models with loss less than \(\ell(z_i)^*\) suggest the opposite, since adding \(z_i\) to the training set would further decrease the loss.
Therefore, we can refine the model spaces \(\mathcal{F}\) and \(\mathcal{F}_{-z_i}\) into two partitions:

\begin{equation}
\begin{split}
\mathcal{F}^+ &= \{ f \in \mathcal{F} : \ell(z_i; f) \geq \ell(z_i)^* \}, \\
\mathcal{F}^- &= \{ f \in \mathcal{F} : \ell(z_i; f) < \ell(z_i)^* \}, \\
\mathcal{F}_{-z_i}^+ &= \{ f \in \mathcal{F}_{-z_i} : \ell(z_i; f) \geq \ell(z_i)^* \}, \\
\mathcal{F}_{-z_i}^- &= \{ f \in \mathcal{F}_{-z_i} : \ell(z_i; f) < \ell(z_i)^* \}.
\end{split}
\end{equation}

\begin{definition}[t-WaKA]
The target-WaKA attribution score for point \(z_i\), denoted as \(\tWaKA(z_i)\), is defined as:
\begin{equation}
\begin{split}
\tWaKA(z_i) = W_1 \left( \mathbb{L} \mid \mathcal{F}^+, \mathbb{L}_{-z_i} \mid \mathcal{F}_{-z_i}^+ \right) \\
  - W_1 \left( \mathbb{L} \mid \mathcal{F}^-, \mathbb{L}_{-z_i} \mid \mathcal{F}_{-z_i}^- \right).
\end{split}
\end{equation}
\end{definition}

We can use the previous formula to obtain the following simplification:
\begin{multline}
\tWaKA(z_i) =  \frac{1}{k} \sum_{l \geq \ell(z_i)^*} \left| F_{\mathbb{\ell_\#Q}}(l) - F_{\mathbb{\ell_\#Q}_{-z_i}}(l) \right|  \\ -  \frac{1}{k} \sum_{l < \ell(z_i)^*} \left| F_{\mathbb{\ell_\#Q}}(l) - F_{\mathbb{\ell_\#Q}_{-z_i}}(l) \right|.
\end{multline}

\subsection{Attribution Algorithm}

{\small
\begin{algorithm*}[!t]
\caption{Counting the marginal contributions of a point of interest $z_i$ with respect to a test point $z_t$}
\label{alg:waka_membership_attack}
\begin{algorithmic}[1]
\State \textbf{Input:} Sorted training set $D$, point of interest $i$, test point $t$, number of neighbors $k$.
\State \textbf{Output:} Contributions for all losses of point $i$

\State Initialize $\text{CPV}[j] = \sum_{m \neq i}^{j} \mathbf{1}(y_{\alpha_m(D)} = y_t)$ for $j = 1, \ldots, N$
\Comment{Cumulative positive votes up to $z_j$ excluding $z_i$}
\State Initialize $\text{CNV}[j] = \sum_{m \neq i}^{j} \mathbf{1}(y_{\alpha_m(D)} \neq y_t)$ for $j = 1, \ldots, N$
\Comment{Cumulative negative votes up to $z_j$ excluding $z_i$}

\State $\text{Contributions} \gets \mathbf{0}$

\For{$j = i+1$ to $N$}
\If{$y_{\alpha_j(D)} \neq y_{\alpha_i(D)}$ and $j > k$}
\ForAll{$l$ in $\{0, 1/k, \ldots, 1\}$}
\State $\text{NV} \gets \text{round}(l \cdot k)$
\Comment{Number of negative votes for loss $l$}
\State $\text{PV} \gets k - \text{NV}$
\Comment{Number of positive votes for loss $l$}

\State $\delta_{\text{PV}, i} \gets \mathbf{1}(y_{\alpha_i(D)} = y_t)$
\State $\delta_{\text{NV}, i} \gets \mathbf{1}(y_{\alpha_i(D)} \neq y_t)$
\State $\delta_{\text{PV}, j} \gets \mathbf{1}(y_{\alpha_j(D)} = y_t)$
\State $\delta_{\text{NV}, j} \gets \mathbf{1}(y_{\alpha_j(D)} \neq y_t)$

\If{$\text{CPV}[j] \geq \text{PV} - \delta_{\text{PV}, j}$ and $\text{CNV}[j] \geq \text{NV} - \delta_{\text{NV}, j}$ and $\text{CPV}[j] \geq \text{PV} - \delta_{\text{PV}, i}$ and $\text{CNV}[j] \geq \text{NV} - \delta_{\text{NV}, i}$} 
\Comment{k-nearest neighbors combinations should be valid}

\State $\text{Count}_{-z_i} \gets \binom{\text{CPV}[j]}{\text{PV} - \delta_{\text{PV}, j}} \cdot \binom{\text{CNV}[j]}{\text{NV} - \delta_{\text{NV}, j}}$
\State $\text{Count} \gets \binom{\text{CPV}[j]}{\text{PV} - \delta_{\text{PV}, i}} \cdot \binom{\text{CNV}[j]}{\text{NV} - \delta_{\text{NV}, i}}$
\State $\text{Contributions}[l] \gets \text{Contributions}[l] + \frac{\text{Count} - \text{Count}_{-z_i}}{ 2^{j}}$
\EndIf
\EndFor
\EndIf
\EndFor

\State \textbf{return} $\text{Contributions}$
\end{algorithmic}
\end{algorithm*}
}

To compute the 1-Wasserstein distance, we assume access to approximations of Probability Mass Functions (PMFs) of $\mathbb{L}$ and $\mathbb{L}_{-i}$, which take the form of histograms. 
More precisely, for each possible loss value \(l\) in the set \(\{0, 1/k, \ldots, 1\}\), we want to count the associated number of existing $k$-NN models, including and excluding \(z_i\). 
In the context of $k$-NNs, there is no need to sample many subsets of the training set to compute these values as we can count exactly the contribution of a point $z_i$. 
A key insight is the observation that computing the difference in PMFs due to $z_i$ is proportional to calculating the marginal contribution of $z_i$ when $z_i$ is added to the training set $D$. 
A contribution occurs for any \( y_i \) if \( y_i \neq y_j \) for all \( j > i \), with the points sorted relative to the test point \( z_t \). 
To realize this, we have designed Algorithm~\ref{alg:waka_membership_attack} to provide an efficient method for computing the marginal contributions. 

In a nutshell, the algorithm starts by ordering the training set with respect to the test point $z_t$ before iterating through the sorted labels to store the cumulative positive and negative votes, \(\text{CPV[j]}\) and \(\text{CNV[j]}\), at each index $j$. 
Note that if $z_t = z_i$, we are in the context of self-attribution, or in the context of an MIA with t-WaKA.
Afterwards, the algorithm loops over the training set to find labels that differ from $y_i$ (this step could be parallelized), the label of the point of interest, which could be precomputed in the previous step. 
For each differing label with an index greater than $k$, a pass is done over the possible loss values to compute the marginal contribution of the point of interest $z_i$. 
This process consists of two steps: first, we count the combinations of $k$ nearest neighbors (of $z_t$) a particular loss, in which $z_i$ is included are counted. 
Similarly, we count the combinations in which $z_j$ is excluded for the same loss. 
Then, this count is normalized using the term $2^j$, which corresponds to all supersets of the $k$ nearest neighbors. 
Finally, the marginal contribution is either stored or aggregated.
Once the marginal contributions are calculated, the computation of the 1-Wasserstein distance is straightforward using histograms of the losses (a proof and more details are provided in Appendix~\ref{sec:appendix-algorithm-details}.


\subsection{Computational analysis} 
The worst-case time complexity of the algorithm is \(O(N \log N + kN)\). 
More precisely, the \(O(N \log N)\) term comes from sorting the \(N\) training points with respect to their distance to the test point \(z_t\). 
Initializing the cumulative vote functions for any $j$, \(\text{CPV[j]}\) and \(\text{CNV[j]}\), requires \(O(N)\). 
The main loop, which iterates over the training set and processes each point's contributions to each loss, runs in \(O(k \cdot N)\). 
This complexity is reduced by using an approximation leveraging the fact that contributions decrease exponentially due to the \(2^j\) factor and that $k$ becomes much smaller than $j$. 
This enables to focus on a fixed-size neighborhood around the target point rather than considering all training points.

In practice in our experiments, we used a neighborhood of 100 points to compute these values. 
By restricting the computation to this fixed neighborhood, \(\text{CPV[j]}\) and  \(\text{CNV[j]}\) can also be measured with a constant complexity. 
To efficiently identify the neighborhood, we can use an optimized data structure such as a kd-tree, which allows for identifying the nearest neighbors in \(O(\log N)\) time.
With this approximation, only a single reusable \(k\)-NN model needs to be trained, which reduces significantly the memory usage compared to using multiple $k$-NN shadow models as done in LiRA. 
More precisely, the complexity of computing marginal contributions for any point of interest in this approximation becomes \(O(\log N + K)\), in which \(K\) is the fixed size of the neighborhood. 
This approach is particularly efficient for large datasets.
In the following experiments, we have employed this approximation  and observed a minimal impact on the estimated value.


\section{Experimental Evaluation}
\label{sec:exp}

In this section, we present our experimental evaluation of WaKA as a general attribution method and t-WaKA as a membership inference attack (MIA) on six widely-used public datasets (see Table~\ref{tab:datasets} in Appendix~\ref{sec:datasets}).
Our main objective with the following experiments is to explore the dual use of WaKA: first, as a tool for understanding the contribution of individual data points to the utility and privacy of $k$-NN models and second, as an efficient approach for conducting MIAs on specific data points.

\subsection{Experimental Setting}

To apply WaKA to standard image and text datasets in machine learning, we prioritized using pre-trained neural network embeddings to minimize dependencies to specific datasets. 
For CIFAR-10, we used a custom pre-trained embedding based on ImageNet, extracting feature representations from its last layer to obtain a transferable encoding. We evaluated both ImageNet embeddings and custom embeddings trained on a reserved portion of CIFAR-10. While the latter showed a slight performance gain, the overall improvement was negligible. Therefore, we chose to use ImageNet embeddings, as this avoids partitioning the training set and reduces the risk of data leakage.
For the CelebA dataset, we employed a pre-trained Vision Transformer (ViT) embedding~\cite{dosovitskiy2020image}. 
For the IMDB and Yelp reviews textual datasets, we used a pre-trained version of the Sentence-BERT (SBERT) model~\cite{reimers2019sentence}, ensuring that it did not include the two datasets during training, thereby aligning with our goal of reduced data dependencies. 
In a nutshell, SBERT is a modification of the BERT architecture that produces sentence embeddings optimized for semantic similarity tasks and downstream classification. 
Finally, for tabular datasets, we employed a straightforward encoding approach, which includes one-hot encoding for categorical features and normalization for numerical features. 
Note that the CelebA and Yelp datasets were added and only used for data valuation experiments.

Our experiments have been designed around two key scenarios for evaluating attribution methods. 
The first scenario, which we call ``test-attribution'', represents the classical data valuation setting, in which the validation set is used to compute aggregated attribution scores and measure their impact on utility through performance on a separate test set. 
The second scenario, which we name ``self-attribution'', focuses on computing the attribution value of a point  $z_i$  based on how well the model predicts the corresponding label  $y_i$. 
This situation quantifies the direct contribution of each point to its own label prediction and we conjecture that this self-attribution is more correlated to privacy scores than utility. 
In a nutshell, to compute self-attribution, we only need to use the values attributed to each point to predict itself while for test-attribution, we must decide on how to aggregate the values calculated for each test point.
More precisely, for test-based attribution, the Shapley value, denoted as \( \DSV_{\text{test}}(z_i) \), computes the average contribution of each training point across the test set, formulated as:
\[
\DSV_{\text{test}}(z_i) = \frac{1}{|D_{\text{test}}|} \sum_{z_t \in D_{\text{test}}} v_{\text{shap}}(z_i; z_t),
\]
in which \( v_{\text{shap}}(z_i; z_t) \) is the Shapley value of training point \( z_i \) with respect to test point \( z_t \). 
This method benefits from Shapley's linearity axiom, which ensures that contributions from individual data points can sum to the total utility. 
Note that a test point can assign both positive and negative valuations to a training point, depending on its effect on the model's performance. Similarly, for WaKA\(_{\text{rem}}\) and WaKA\(_{\text{add}}\), we compute the average across the test set by evaluating each test point \(z_t \in D_{\text{test}}\).

In our experiments, we selected $k=1$ as the least privacy-preserving parameter value, since the $k$-NN model's predictions are directly influenced by individual points. 
We also chose $k=5$ because it is a commonly used parameter in the literature, offering more label anonymization through generalization, as predictions are influenced by a larger set of neighboring points.

\subsection{WaKA for Utility-driven Data Minimization}

\begin{figure*}[h!]

    \centering
\caption{Attribution using $WaKA_{\text{rem}}$ for data removal and $WaKA_{\text{add}}$ for data addition, compared with two test-attribution methods: Data Shapley and Leave-One-Out (LOO). \textbf{Data addition} starts with an empty dataset (0\%) and progressively adds points, either randomly (black line) or in descending order of importance as ranked by an attribution method (e.g., Shapley, $WaKA_{\text{add}}$, LOO). \textbf{Data removal} begins with the full dataset (100\%) and iteratively removes the least valuable points according to each method. The x-axis represents the percentage of data added or removed, while the y-axis tracks the model’s accuracy on the test set}

    \label{tab:data-minimization-utility}
    \begin{tikzpicture}[baseline=(current bounding box.north)]
        \node at (2.6, 7.5) {CIFAR10};
        \node at (7, 7.5) {Adult};
        \node at (11.5, 7.5) {Yelp};
        \node at (15.5, 7.5) {IMDB};
        
        \node[rotate=90] at (-0.5, 5.0) {Data Removal};
        \node[rotate=90] at (-0.5, 1.3) {Data Addition};

        \node[anchor=north west, inner sep=0, outer sep=0] (image1) at (0, 7.2) {\includegraphics[width=0.26\linewidth]{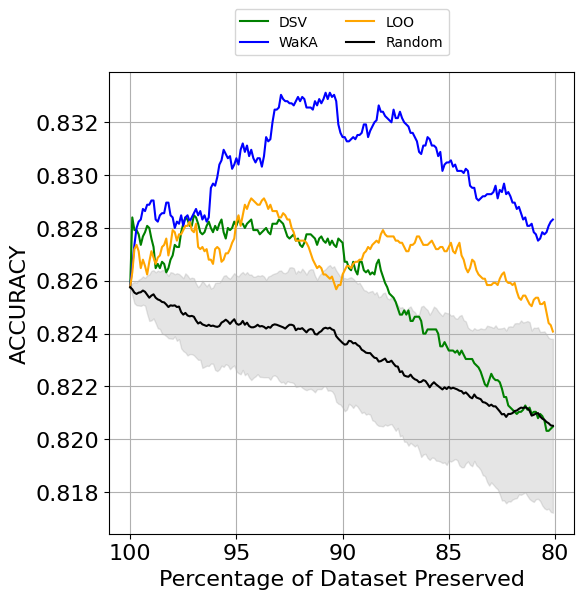}};
        \node[anchor=north west, inner sep=0, outer sep=0] (image2) at (4.5, 7.2) {\includegraphics[width=0.26\linewidth]{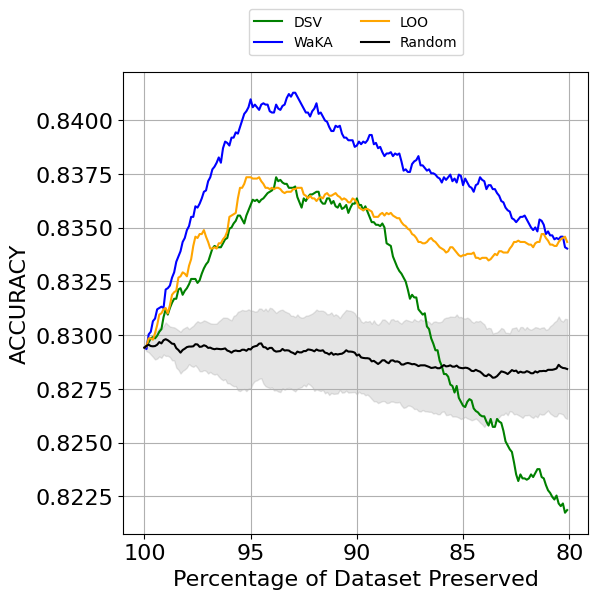}};
        \node[anchor=north west, inner sep=0, outer sep=0] (image3) at (9.0, 7.2) {\includegraphics[width=0.26\linewidth]{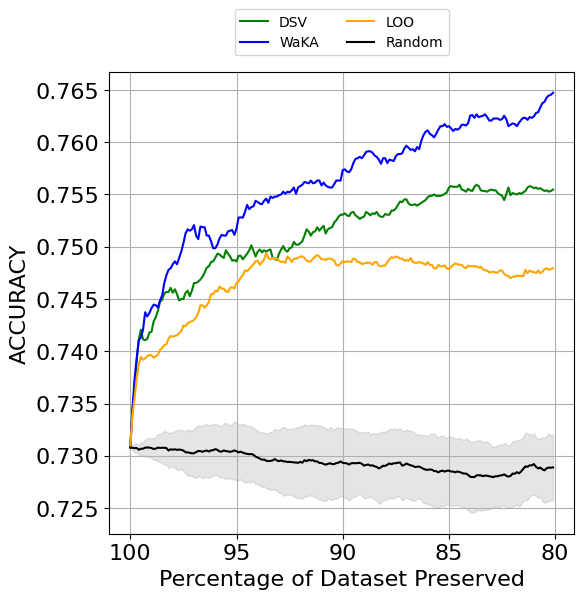}};
        \node[anchor=north west, inner sep=0, outer sep=0] (image4) at (13.4, 7.2) {\includegraphics[width=0.26\linewidth]{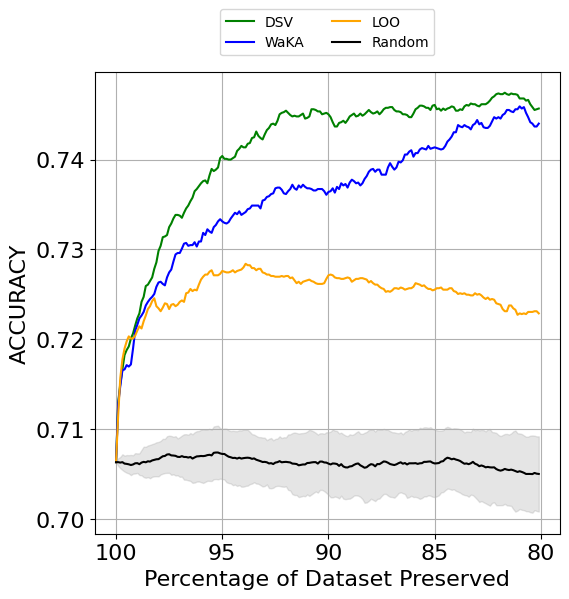}};

        \node[anchor=north west, inner sep=0, outer sep=0] (image5) at (0, 3.4) {\includegraphics[width=0.26\linewidth]{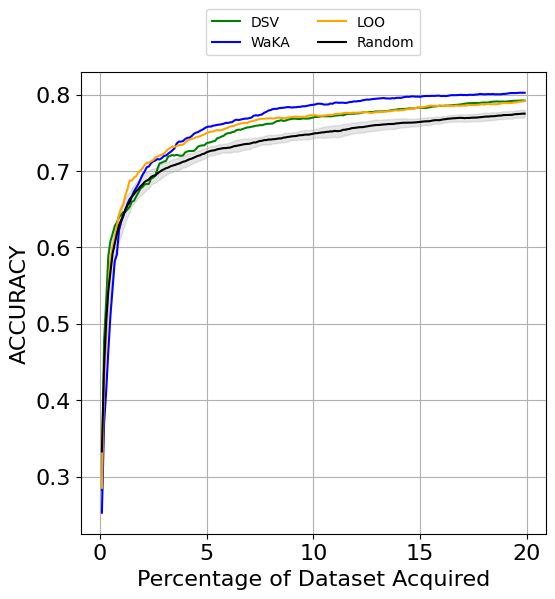}};
        \node[anchor=north west, inner sep=0, outer sep=0] (image6) at (4.5, 3.4) {\includegraphics[width=0.26\linewidth]{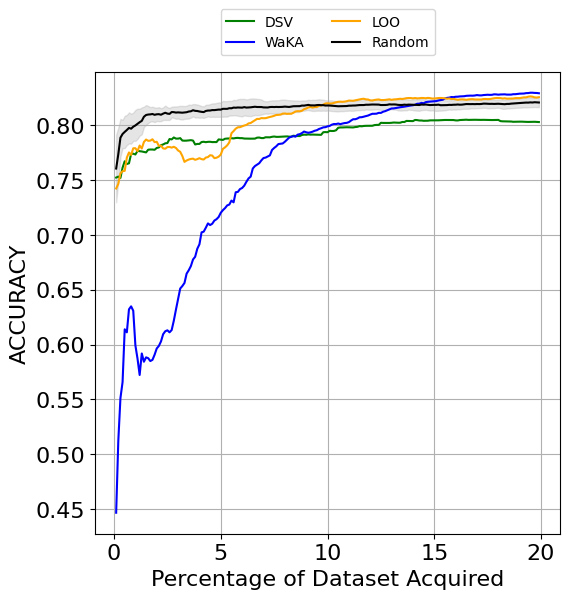}};
        \node[anchor=north west, inner sep=0, outer sep=0] (image7) at (9.0, 3.4) {\includegraphics[width=0.26\linewidth]{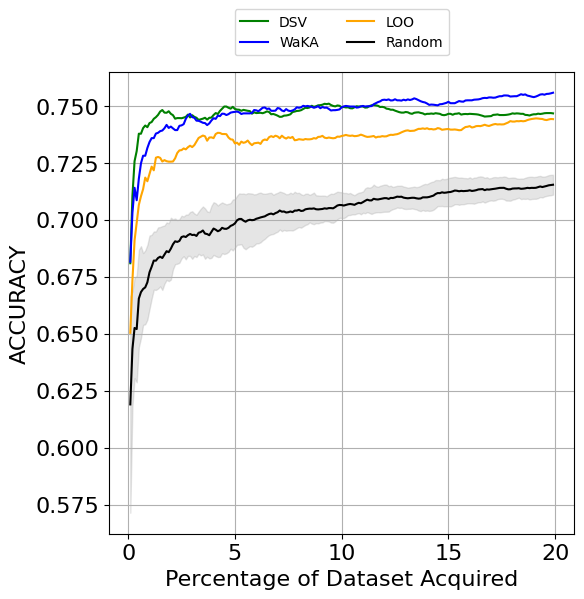}};
        \node[anchor=north west, inner sep=0, outer sep=0] (image8) at (13.4, 3.4) {\includegraphics[width=0.26\linewidth]{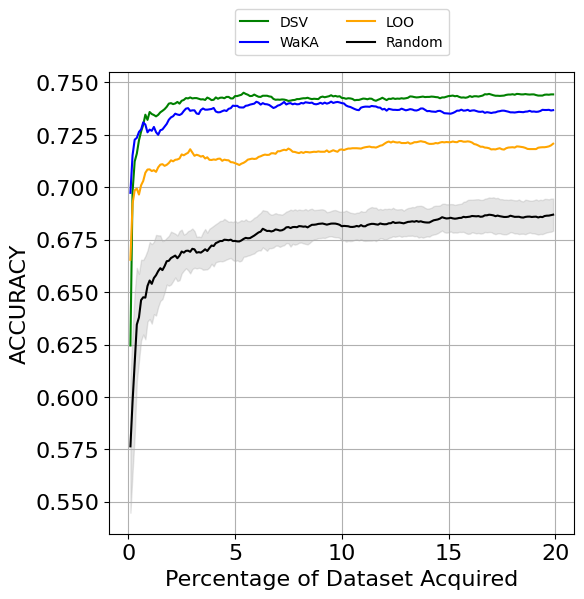}};
    \end{tikzpicture}

    \label{fig:data-minimization-experiments}

\end{figure*}

The state-of-the-art approach for utility-driven data minimization is Data Shapley Value (DSV)~\cite{jia2019efficient}. 
In addition to WaKA and DSV, the basic Leave-One-Out (LOO) method was also included, serving as a benchmark due to its simplicity in utility-driven data removal and addition tasks. In data addition,  models are trained incrementally by adding a certain percent of the dataset in descending order of importance based on an attribution method. 
Conversely, in data removal, models start with the full dataset and are retrained after progressively removing the least valuable points. This setup allows us to assess the impact of each method on model performance as data is added or removed. Looking at the results (Figure~\ref{fig:data-minimization-experiments}), it is clear that the LOO method generally underperforms as a data valuation technique, consistent with the findings of~\cite{ghorbani2019data}.

This reinforces the limitations of LOO in effectively identifying the most influential points for utility-driven data minimization.

\begin{figure}[h]
    \centering
    \includegraphics[width=0.6\linewidth]{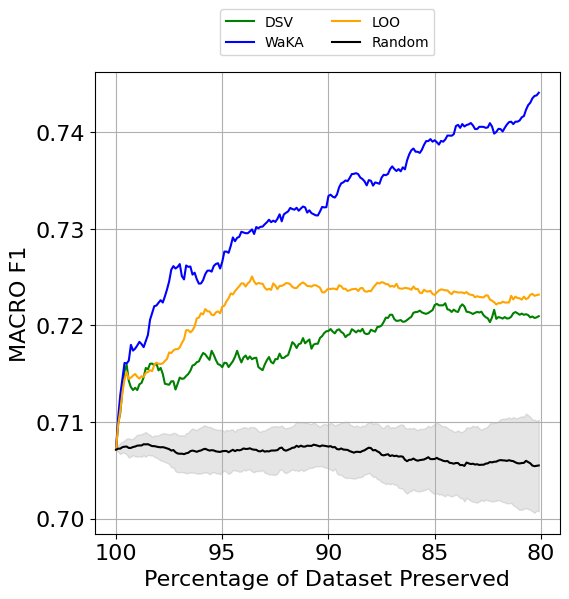}
    \caption{F1 score on the Yelp dataset for data removal. WaKA\(_{\text{rem}}\) maintains significantly better performance compared to DSV and LOO, highlighting its robustness on imbalanced datasets.}
    \label{fig:yelp-f1}
\end{figure}

\begin{figure}[h]
    \centering
    \includegraphics[width=0.6\linewidth]{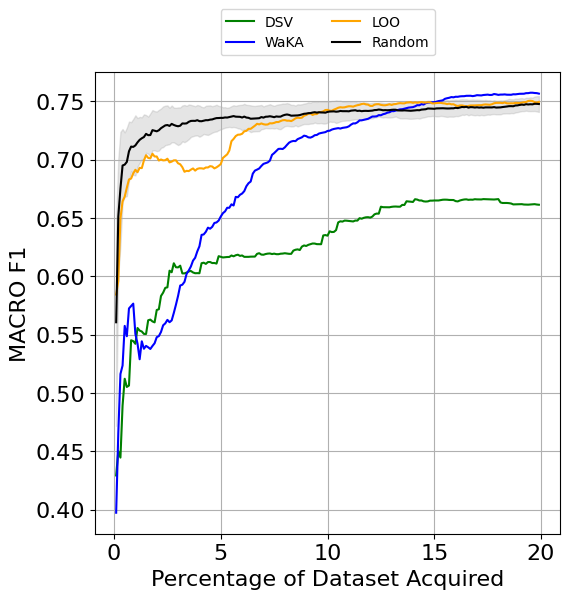}
    \caption{F1 score on the Adult dataset for data addition. DSV predominantly favors adding majority class samples, leading to performance worse than random addition. WaKA\(_{\text{add}}\), by contrast, effectively balances the dataset and achieves better results.}
    \label{fig:adult-f1}
\end{figure}

In contrast, WaKA\(_{\text{rem}}\) and WaKA\(_{\text{add}}\) consistently match or outperform DSV, particularly in imbalanced datasets. 
For instance, Yelp, with a minority class of ~\(0.4\), and Adult, with a minority class of ~\(0.24\), are noticeably more imbalanced compared to CIFAR and IMDB, which are perfectly balanced through preprocessing. 
As shown in Figure~\ref{fig:data-minimization-experiments}, WaKA demonstrates greater robustness in these cases. 
On the Yelp dataset, we observe that while DSV's macro F1 score is comparable to LOO for data removal, WaKA\(_{\text{rem}}\) maintains significantly better performance (Figure~\ref{fig:yelp-f1}). Similarly, on the Adult dataset, Figure~\ref{fig:adult-f1} illustrates that during data addition, DSV predominantly favors adding majority class samples, leading to performance worse than random addition, whereas WaKA\(_{\text{add}}\) effectively balances the dataset, yielding superior results.

An important aspect of WaKA's formulation is its sensitivity to the parameter $\tau$, which controls the weighting of outliers and inliers during data removal and data addition. The role of $\tau$ is to introduce flexibility in how probability mass is moved within the loss distribution, influencing whether points are prioritized based on their effect on model utility. Specifically, when $\tau = 1.0$ for addition and $\tau = 0.0$ for removal, the formulations become identical. In data addition, the goal is to emphasize inliers—points that contribute significantly to reducing loss—while penalizing those that negatively impact other points depending on their loss distribution. In data removal, we focus on outliers, particularly those that increase loss, while still accounting for outliers that provide positive contributions. Empirical evaluations (see Appendix ~\ref{sec:appendix-utility-results}) indicate that $\tau = 0.5$ leads to stable results across datasets, though higher variance is observed in imbalanced datasets such as Bank and CelebA. This suggests that further exploration of $\tau$ could be valuable in contexts where dataset imbalance affects data attribution.

\begin{figure}[h]
    \centering
    \includegraphics[width=0.6\linewidth]{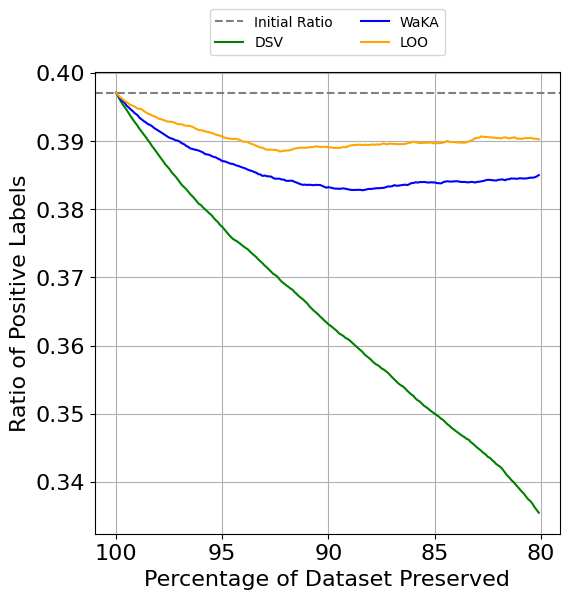}
    \caption{Effect of data removal on class balance for the Yelp dataset.
    WaKA\(_{\text{rem}}\) preserves class proportions, while DSV disproportionately removes points from the minority class, exacerbating imbalance.}
    \label{fig:class-balance}
\end{figure}

Finally, to further explore the impact of data valuation methods, we examined their influence on class balance during data removal. Figure~\ref{fig:class-balance} highlights that DSV tends to disproportionately remove points from the minority class, exacerbating class imbalance. 
In contrast, WaKA\(_{\text{rem}}\) demonstrates a more balanced removal strategy, effectively preserving class proportions. We include all results, such as CelebA and Bank datasets, in Appendix ~\ref{sec:appendix-utility-results}. 
These additional experiments confirm the bias of Shapley values towards the majority class and further underscore the robustness of WaKA in handling imbalanced data scenarios.

\subsection{WaKA for Privacy Evaluation and Auditing}

\textbf{Privacy scores.} In our privacy evaluation, we followed a similar approach to~\cite{carlini2022privacy} by computing the average ASR on all training points. 
In practice, this was done by simulating security games as described previously, in which multiple random partitions of the training set were created, and the LiRA method was applied to evaluate the ASR. 
Once the ASR values were obtained, the attribution scores were computed from each method and sorted to analyze their correlation with the ASR. 
This allows to identify how well each attribution method aligns with the likelihood of a successful MIA.

\begin{figure*}[h!]
\centering
\caption{Correlation between ASR and self-attribution values across different datasets for $k$-NN values of $k=1$ and $k=5$. 
\textbf{Self-attribution} measures the extent to which a data point contributes to its own prediction, while ASR (Attack Success Rate) represents the likelihood of a successful membership inference attack, serving as a measure of privacy risk per point. 
The ASR increases monotonically with self-attribution values in most datasets, but the behavior varies across datasets. 
For example, the Bank dataset exhibits a steep increase in ASR around the 70th percentile of self-WaKA values, while other datasets, such as IMDB, show a more gradual augmentation. 
Additionally, \textbf{test-attribution} is less correlated with ASR compared to self-attribution, indicating that self-attribution is a stronger predictor of privacy risk. 
The darker curves represent self-attribution methods, including self-Shapley (red) and self-WaKA (blue).}

    \begin{tikzpicture}[baseline=(current bounding box.north)]
        \node at (2.6, 7.5) {CIFAR10};
        \node at (7, 7.5) {Bank};
        \node at (11.5, 7.5) {Adult};
        \node at (15.5, 7.5) {IMDB};
        
        \node[rotate=90] at (-0.5, 5.0) {K=1};
        \node[rotate=90] at (-0.5, 1.3) {K=5};
        
        \node[anchor=north west, inner sep=0, outer sep=0] (image1) at (0, 7.2) {\includegraphics[width=0.27\linewidth]{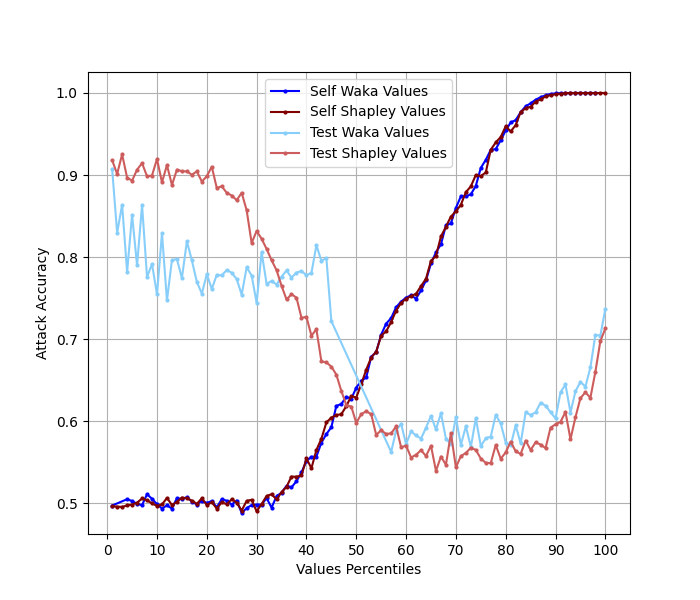}};
        \node[anchor=north west, inner sep=0, outer sep=0] (image2) at (4.5, 7.2) {\includegraphics[width=0.27\linewidth]{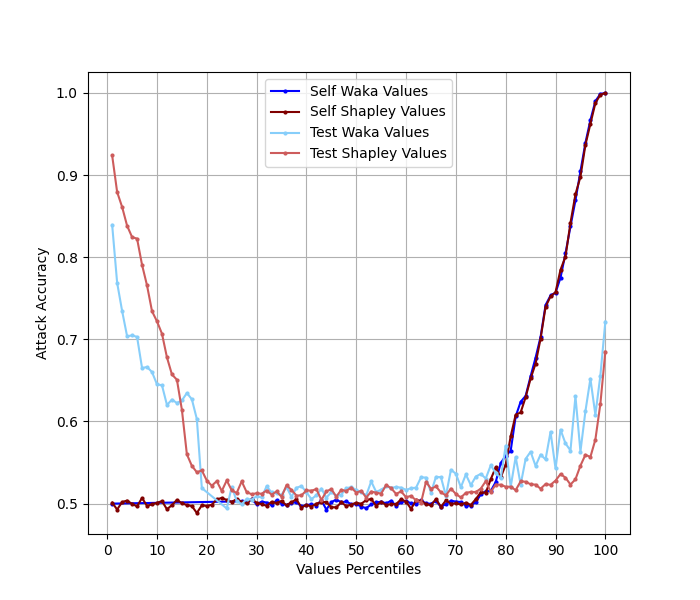}};
        \node[anchor=north west, inner sep=0, outer sep=0] (image3) at (9.0, 7.2) {\includegraphics[width=0.27\linewidth]{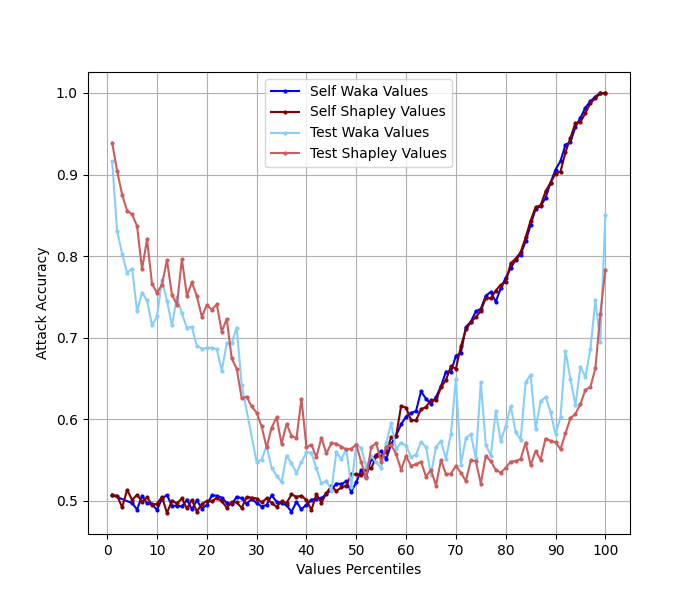}};
        \node[anchor=north west, inner sep=0, outer sep=0] (image4) at (13.4, 7.2) {\includegraphics[width=0.27\linewidth]{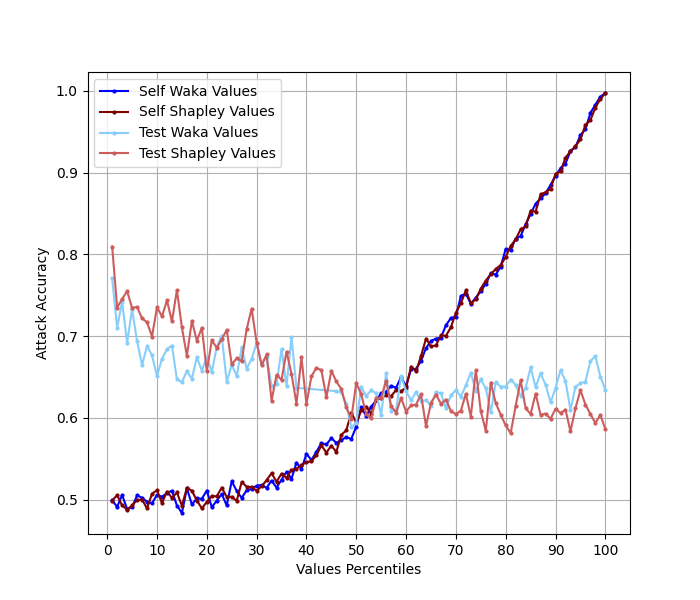}};
        
        \node[anchor=north west, inner sep=0, outer sep=0] (image5) at (0, 3.4) {\includegraphics[width=0.27\linewidth]{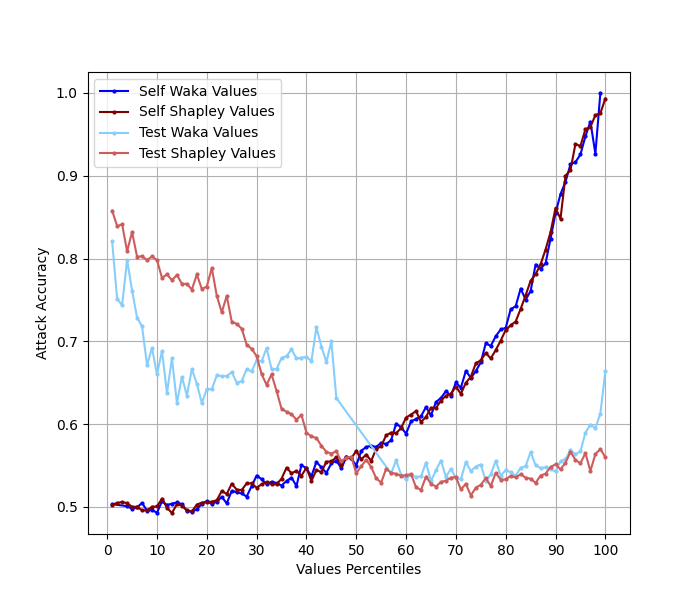}};
        \node[anchor=north west, inner sep=0, outer sep=0] (image6) at (4.5, 3.4) {\includegraphics[width=0.27\linewidth]{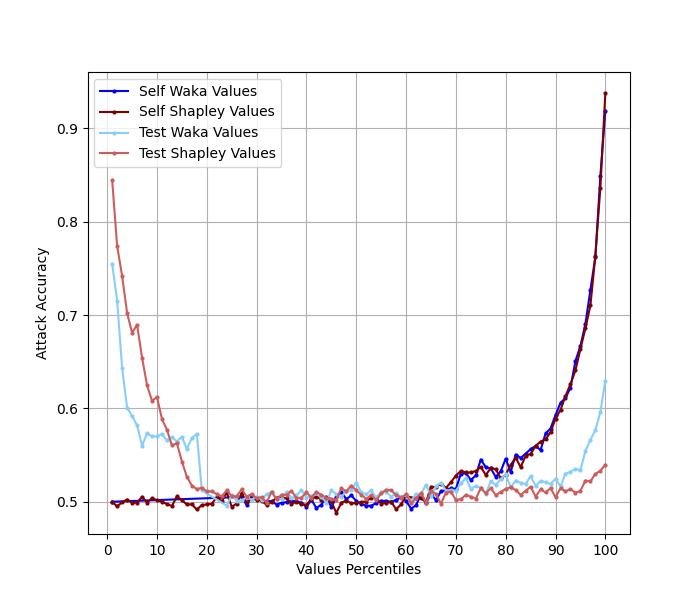}};
        \node[anchor=north west, inner sep=0, outer sep=0] (image7) at (9.0, 3.4) {\includegraphics[width=0.27\linewidth]{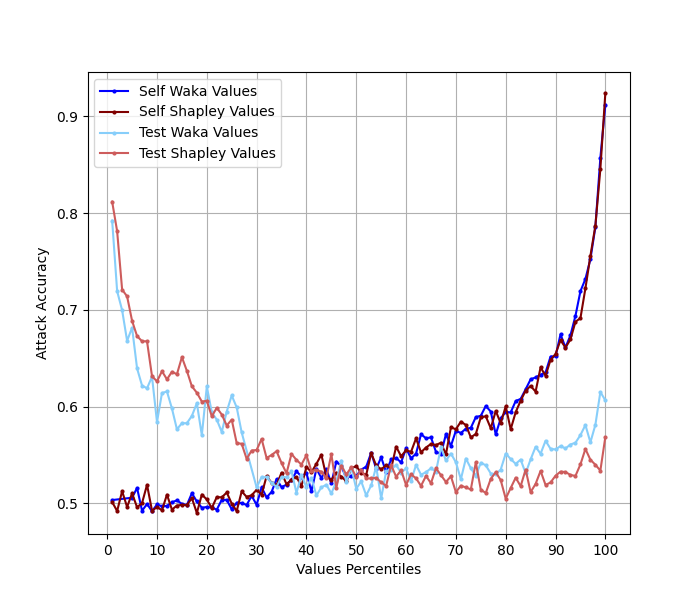}};
        \node[anchor=north west, inner sep=0, outer sep=0] (image8) at (13.4, 3.4) {\includegraphics[width=0.27\linewidth]{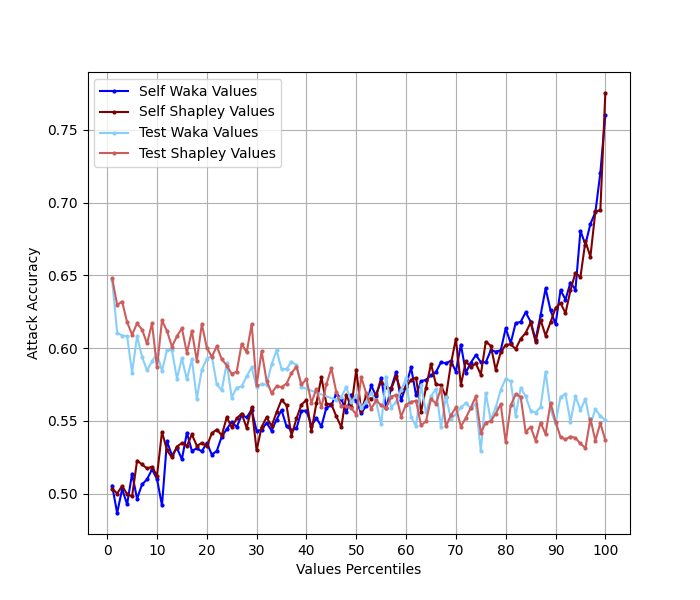}};
    \end{tikzpicture}

    \label{fig:asr_correlation}
\end{figure*}

Both self-Shapley and self-WaKA have shown a monotonic increase in self-attribution values as the ASR increases (see Figure~\ref{fig:asr_correlation}), for both values of the  $k$-NN parameter (\emph{i.e.}, $k = 1$ and $k = 5$). 
However, it is important to note that the correlation between ASR and self-attribution values does not behave uniformly across all datasets.
For instance, in the Bank dataset, the attack accuracy remains close to 0.5 (\emph{i.e.}, random guessing) until about the 70-th percentile of self-WaKA values, whereas other datasets, such as IMDB, exhibit a more gradual increase in ASR across percentiles. 
This indicates that different datasets exhibit varying levels of correlation between ASR and self-attribution. 
Furthermore, test-attribution methods showed different correlations, with notably a higher average ASR towards extremes. 
This means that very low or very high attribution scores tend to correlate more strongly with higher ASR values. 
More precisely, data points that are highly detrimental to utility were found to have consistently high ASR values across all datasets.
This observation challenges the common belief that only high-value points pose significant privacy risks as lower-value points can also be privacy vulnerable.

Additionally, increasing the parameter value $k$  reduces the average ASR (Table \ref{tab:asr_stats_k1_k5} in Appendix~\ref{sec:appendix-privacy-results}) while shifting higher ASR values towards the upper percentiles of the attribution scores. 
We further tested the Spearman rank correlation between ASR and attribution scores, finding that they are highly correlated across datasets (see Table \ref{tab:table_waka_shap_spearman} in Appendix~\ref{sec:appendix-privacy-results}). 
This suggests that self-attribution methods, such as in particular self-WaKA, can more reliably indicate privacy risks in $k$-NN models than test attribution.
This can be explained by the fact that self-Shapley can be understood through a game-theoretic lens, in which each data point $z_i$ is a ``player'' contributing to the prediction. 
In this setting, a high self-Shapley value means that $z_i$ plays a dominant role in predicting itself, indicating that whatever the coalition, its neighbors contribute little. 
Similarly, a high self-WaKA value means that removing $z_i$ significantly changes the loss distribution of predicting itself across all possible subsets of the training set,
which aligns more closely with LiRA.

Finally, we have conducted additional experiments to determine whether self-WaKA values could provide privacy insights for models beyond $k$-NN classifiers. 
To investigate, LiRA and trained logistic regression shadow models we first looked at the logits of the models confidence, following the approach outlined by Carlini and co-authors.
More precisely, the LiRA scores were compared with the self-WaKA values for $k=1$ to see if there was a correlation with the ASR. 
Our observations revealed that, across all datasets, the average ASR for the logistic regression model was lower than that of $k$-NN with $k=5$ (see Table \ref{tab:asr_stats_k1_k5}). 
Although the correlation between value percentiles and ASR was less pronounced than in the $k$-NN case, a significant relationship can still be observed (see Figure \ref{fig:privacy-eval-regression} in Appendix~\ref{sec:appendix-privacy-results}). 
This suggests that self-attribution on $k$-NN models may offer insights into the data that can be extrapolated to other types of models, although we leave as future works the detailed investigation of such research avenes.

\textbf{Privacy influences.} The ``onion effect'' as explored in~\cite{carlini2022privacy} throught the introduction the concept of privacy influence (\( \text{PrivInf} \)) as a method to quantify the influence that removing a specific training example \( z \) has on the privacy risk of a target example \( z' \). 
Specifically, it is defined as the change in the membership inference accuracy on \( z' \) after removing \( z \), averaged over models trained without \( z \):
\[
\text{PrivInf}(\text{Remove}(z) \Rightarrow z') := \mathbb{E}_{f \in \mathcal{F}, S \subseteq D} \left[ \mathbf{1}(z' \in S) \mid z \notin S \right]
\]
This definition can be used to analyze how the removal of certain training samples (\emph{e.g.}, inliers or outliers) affects the ASR of other data points.

Following the same experimental setting as in~\cite{carlini2022privacy}, we started by removing 10\% of the training set with the highest self-WaKA values before re-evaluating the ASR on the remaining points. 
The success of MIAs across all data points is compared by plotting the AUC curve on a log scale, following the authors’ recommendation (see Figure \ref{fig:auc-post-removal} in Appendix~\ref{sec:appendix-privacy-results}). 
As observed by Carlini and collaborators, while the overall risk is reduced after removing certain training points, it remains higher than the expected decrease.
Similar patterns were observed when removing points with the highest ASR or self-Shapley values, whereas random removal or other attribution methods resulted in no significant change in ASR reduction (see Tables~\ref{tab:stats_before_after_removal1} and~\ref{tab:stats_before_after_removal2} in Appendix~\ref{sec:appendix-privacy-results}). 
This confirms that $k$-NN models exhibit similar privacy layers, for which removing vulnerable data does not proportionally reduce the vulnerability to privacy attacks, thus confirming the ``onion effect''. 
The distribution of ASR was also analyzed across all points. 
For CIFAR10, many points still exhibit high ASR values close to 1.0 after removal. 
For the Bank, Adult and IMDB datasets, a general shift occurs towards lower ASR values after data removal, though some points with high ASR remain, particularly near the upper percentiles. 
This highlights the varying impact of data removal on privacy risks across datasets.

Computing privacy influence (\( \text{PrivInf} \)) is computationally expensive, as it requires removing a point \( z \) or a subset of points \( Z \) and re-evaluating privacy scores for all other points \( z' \) in the training set to determine which points have the most influence on their privacy score. 
Instead of directly computing \( \text{PrivInf} \), we investigated whether changes in the ASR of the remaining points after removing 10\% of the training set could be explained by their self-attribution values. 
To achieve this, an approach to compute point-wise influence on self-attribution values is needed.
In particular for Shapley, this task is non-trivial, as exact Shapley values are computed recursively, starting from the last sorted data point (\emph{i.e.}, complexity of \( O(n) \).
Efficiently re-computing Shapley values after removing a point would require adapting the Shapley $k$-NN algorithm to support faster recalculations.

For WaKA, the situation is different as the contributions of each point to the WaKA value for a point \( z_i \) are independent and can be stored, allowing for re-computation of the self-WaKA value in \( O(1) \) time. 
This provides a strong advantage for self-WaKA over self-Shapley in terms of computational efficiency. 
To compute the influence of removing a subset \( Z \subseteq D \) on the self-WaKA values—denoted as \( \text{WaKAInf}(\text{Remove}(Z) \Rightarrow z) \) —the contributions of each removed point \( z_j \in Z \)on the self-WaKA values are summed on the remaining points. The total WaKA influence is defined as:
\begin{equation}
\begin{split}
\text{WaKAInf}(\text{Remove}(Z) \Rightarrow z) = \\
\sum_{z_j \in \mathcal{N}(z) \cap Z}  \text{WaKA}_{self}(z; D_{-z_j}) - \text{WaKA}_{self}(z; D) ,
\end{split}
\end{equation}
in which \( \mathcal{N}(z) \) represents the fixed neighborhood around \( z \), and \( z_j \) corresponds to a point that was part of the subset \( Z \) removed. 
If no influencing points are included, the total influence is zero. 

\begin{figure*}

    \centering
\caption{The relationship between ASR change and total self-WaKA influence (WaKAInf) for $k=1$. Three regimes can be seen: negative WaKAInf values lead to negative ASR changes (reduced vulnerability), medium WaKAInf values show minimal ASR change while and higher WaKAInf values correlate with positive ASR changes (increased privacy risks). ASR histograms before and after removal for \( k=1 \).}

    \begin{tikzpicture}[baseline=(current bounding box.north)]
        \node at (2.6, 7.5) {CIFAR10};
        \node at (7, 7.5) {Bank};
        \node at (11.5, 7.5) {Adult};
        \node at (15.5, 7.5) {IMDB};
        
        \node[rotate=90] at (-0.5, 5.0) {WaKAInf};
        
        \node[anchor=north west, inner sep=0, outer sep=0] (image1) at (0, 7.2) {\includegraphics[width=0.27\linewidth]{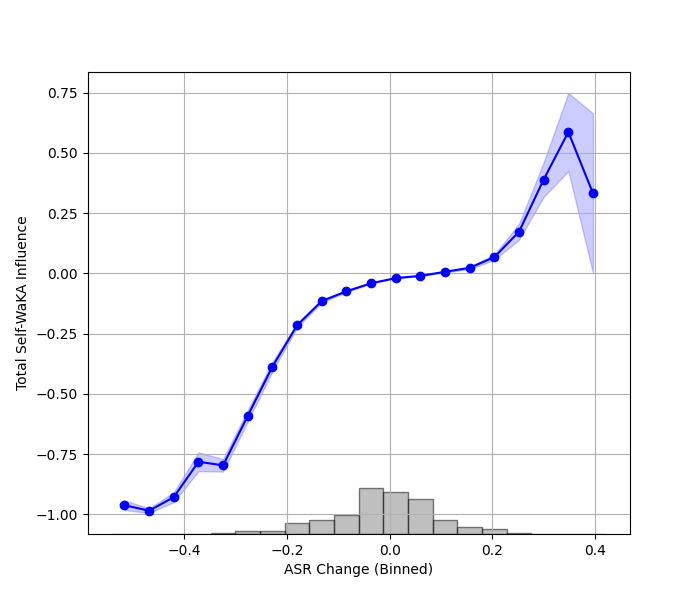}};
        \node[anchor=north west, inner sep=0, outer sep=0] (image2) at (4.5, 7.2) {\includegraphics[width=0.27\linewidth]{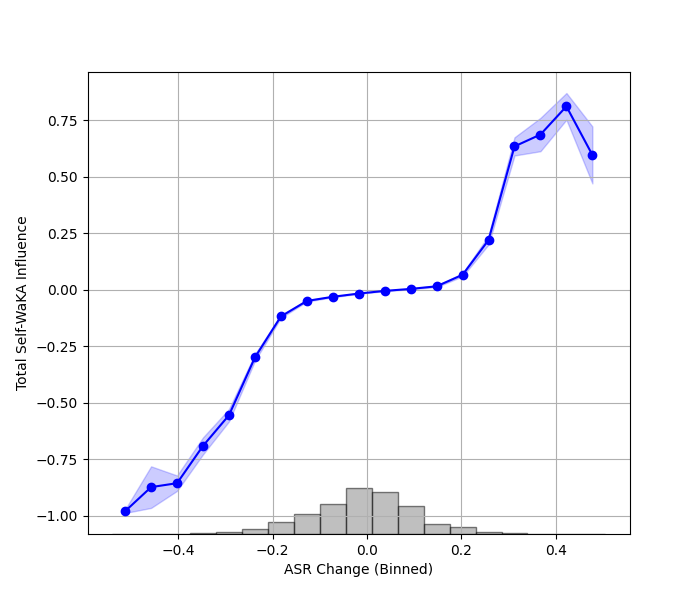}};
        \node[anchor=north west, inner sep=0, outer sep=0] (image3) at (9.0, 7.2) {\includegraphics[width=0.27\linewidth]{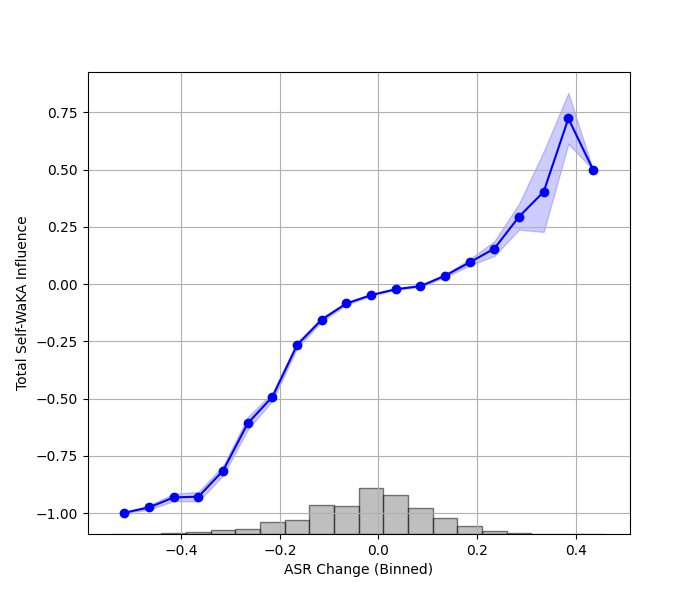}};
        \node[anchor=north west, inner sep=0, outer sep=0] (image4) at (13.4, 7.2) {\includegraphics[width=0.27\linewidth]{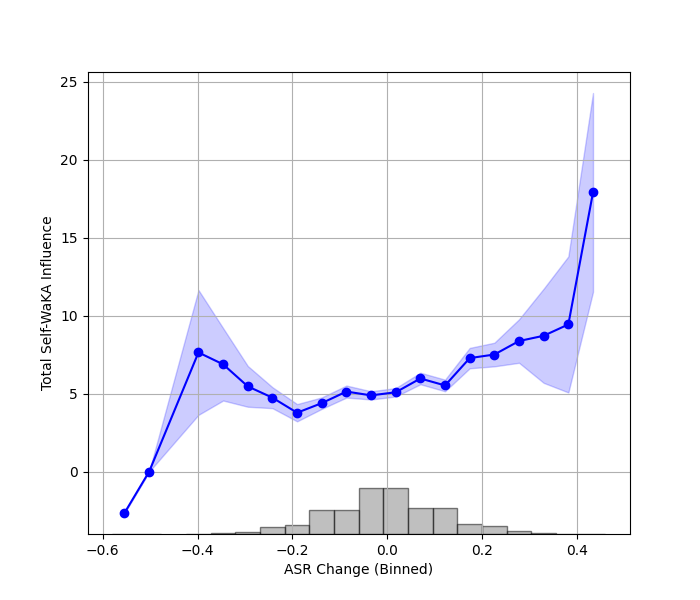}};
        
    \end{tikzpicture}

    \label{fig:privacy_inf_waka_influences}

\end{figure*}

Figure \ref{fig:privacy_inf_waka_influences} shows the correlation between ASR change and WaKAInf, using k=1, the value of the $k$ parameter with the highest privacy risks. 
Three distinct regimes across all datasets can be observed. 
In the leftmost region, points with negative WaKAInf values correspond to negative ASR changes, indicating reduced vulnerability after removal. 
In the middle, in which WaKAInf values are near zero, the ASR changes are close to 0, suggesting little impact on privacy. 
Finally, in the rightmost region, higher WaKAInf values align with positive ASR changes, indicating increased vulnerability. 

\begin{figure*}
\caption{Comparison of data points ASR histograms at \( k=1 \). The blue bars represent the 100\% dataset scenario, while the red bars show the 90\% dataset scenario after removing the 10\% of points with the highest self-WaKA values. 
While the overall ASR distributions appear similar, the removal of these high-risk points significantly reduces the probability of having data points with ASR close to 100\%, indicating a mitigation of extreme privacy vulnerabilities.}

    \centering

    \begin{tikzpicture}[baseline=(current bounding box.north)]
        \node at (2.6, 7.5) {CIFAR10};
        \node at (7, 7.5) {Bank};
        \node at (11.5, 7.5) {Adult};
        \node at (15.5, 7.5) {IMDB};
        
        \node[rotate=90] at (-0.5, 5.0) {ASR};
        
        \node[anchor=north west, inner sep=0, outer sep=0] (image1) at (0, 7.2) {\includegraphics[width=0.27\linewidth]{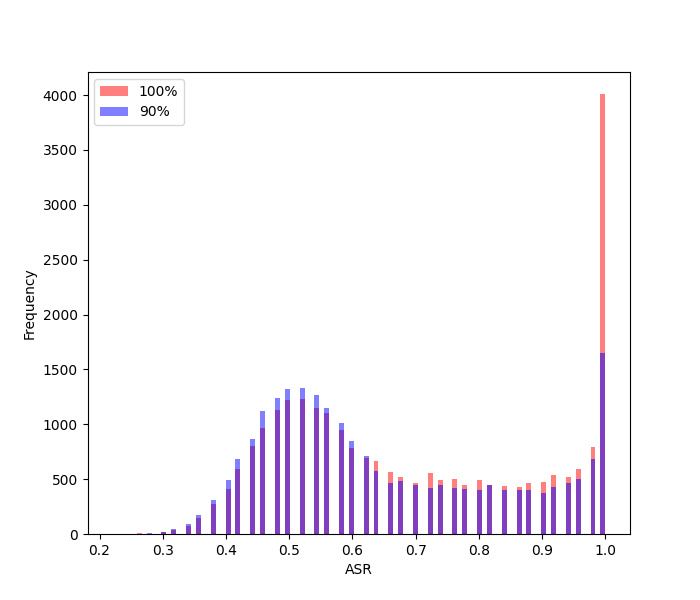}};
        \node[anchor=north west, inner sep=0, outer sep=0] (image2) at (4.5, 7.2) {\includegraphics[width=0.27\linewidth]{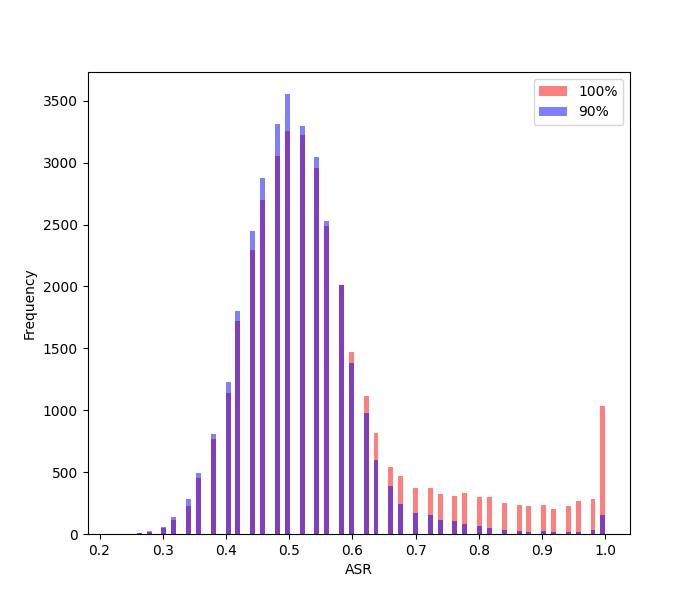}};
        \node[anchor=north west, inner sep=0, outer sep=0] (image3) at (9.0, 7.2) {\includegraphics[width=0.27\linewidth]{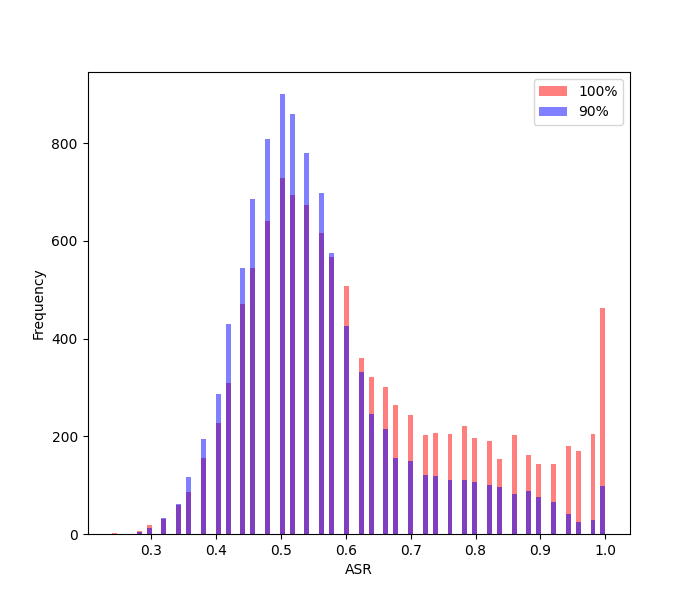}};
        \node[anchor=north west, inner sep=0, outer sep=0] (image4) at (13.4, 7.2) {\includegraphics[width=0.27\linewidth]{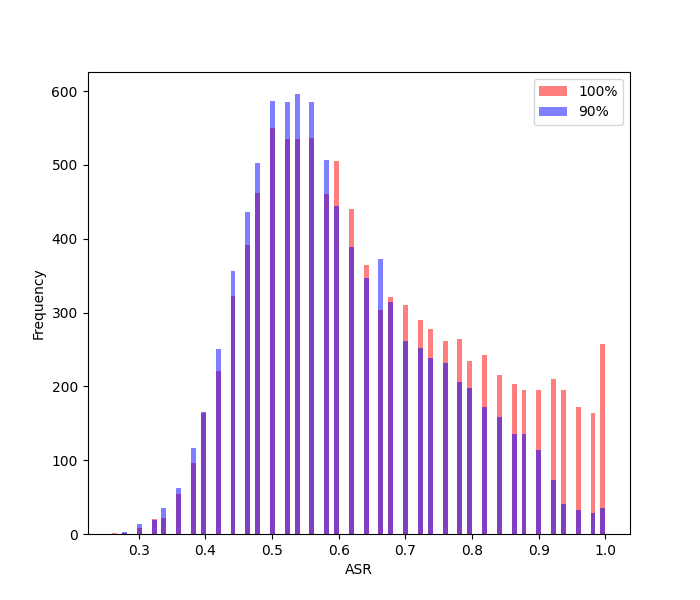}};
        
    \end{tikzpicture}

    \label{fig:privacy_inf_histograms}

\end{figure*}

\textbf{Membership inference attack through target-WaKA (t-WaKA).}
We have conducted extensive experiments using target-WaKA with multiple parameters ($k= 1$ to 5) across all six datasets. The results of the attacks, specifically TPR at a low FPR, are shown in Table~\ref{tab:tpr_comparison}. Additional results, including AUC curves (in log scale) and their corresponding values, are provided in Appendix~\ref{sec:appendix-privacy-results}. Inspired by the confidence-based attacks described in \cite{ye2022enhanced}, we introduce two new attacks on $k$-NN: a "confidence" attack (Conf) and calibrated confidence attack (Conf-calib). Both leverage the target model’s confidence on each target point and compare it to that of a $k$-NN model built on a small neighborhood (i.e. 100 points). For Conf, this is done with a single $k$-NN, whereas Conf-calib samples multiple $k$-NNs (akin to using shadow models) to calibrate the confidence score against variations in the local neighborhood. Since t-WaKA and LiRA share the same principles of leveraging a model’s response for membership inference, LiRA will serve as our main point of comparison.

To compute the mean AUC and TPRs (True Positive Rates) for all FPRs (False Positive Rates) used in ROC curves, a bootstrap approach was employed in which 48 security games are ran in parallel, using a predefined list of seeds. 
In each game, the training set was split in half, with a $k$-NN model trained on one half. 
From these two halves, 100 points were drawn at random for evaluation. 
The LiRA method was implemented with 16 shadow models and although we have tested with an increase in the number of shadow models, an early convergence was observed across all datasets. 
Training a $k$-NN model essentially involves storing the entire training set and optionally building an optimized structure, such as a $kd$-tree, to facilitate quick neighborhood searches during inference. 
In our experiment, we did not use any such structures but rather, we ordered all points with respect to the target or test point.

These experiments differ from the previous privacy evaluations, as 100 randomly sampled points from the dataset are targeted for a specific model, repeating the process 48 times. 
While one might expect that using attribution values would lead to effective attacks on that repeated security game, we found the opposite. 
More precisely, all attribution methods produced an average AUC close to 0.5, indicating random performance. 
We believe that this can be explained by the fact that attribution methods evaluate data points with respect to all possible models, rather than focusing on a specific model’s loss. 
In contrast, both LiRA and t-WaKA can incorporate the loss of a particular model, making them more effective for this type of targeted attack.

\begin{table}[htbp]

\centering
\caption{Comparison of TPR at FPR=0.05 among LiRA, $\tWaKA$, Conf, and Conf-calib for $K=5$}
  \label{tab:tpr_comparison}
\begin{tabular}{lcccc}
\toprule
Dataset & LiRA & $\tWaKA$ & Conf & Conf-calib \\
\midrule
Adult & 0.13 $\pm$ 0.17 & 0.13 $\pm$ 0.15 & 0.06 $\pm$ 0.11 & 0.14 $\pm$ 0.19 \\
Bank & 0.10 $\pm$ 0.16 & 0.10 $\pm$ 0.19 & 0.05 $\pm$ 0.07 & 0.12 $\pm$ 0.17 \\
CelebA & 0.07 $\pm$ 0.09 & 0.08 $\pm$ 0.12 & 0.06 $\pm$ 0.07 & 0.08 $\pm$ 0.11 \\
CIFAR10 & 0.16 $\pm$ 0.24 & 0.15 $\pm$ 0.16 & 0.05 $\pm$ 0.10 & 0.14 $\pm$ 0.17 \\
IMDB & 0.10 $\pm$ 0.16 & 0.14 $\pm$ 0.18 & 0.06 $\pm$ 0.11 & 0.12 $\pm$ 0.19 \\
Yelp & 0.11 $\pm$ 0.17 & 0.13 $\pm$ 0.17 & 0.07 $\pm$ 0.09 & 0.10 $\pm$ 0.16 \\
\bottomrule
\end{tabular}
\end{table}

For almost all values of $k$, the AUCs of LiRA, t-WaKA, and Conf-calib are remarkably comparable.
Following the recommendation of~\cite{carlini2022membership} to assess the success of membership inference attacks, we focus on TPR at a low FPR (5
As the value of the parameter $k$ increases, the success rate of the attacks decreases similarly for all three methods.
We observed that t-WaKA performs slightly worse and sometimes a little better than LiRA, but overall, the results are very close (see Table~\ref{tab:tpr_comparison}).
The Conf attack performs worse than the other methods, but Conf-calib performs just as well as LiRA and t-WaKA, sometimes even surpassing them.
Interestingly, t-WaKA remains the strongest attack on textual datasets (IMDB and Yelp).
In some scenarios  
Additionally, the results confirmed that t-WaKA shows significant correlation in rankings using the Spearman test, indicating consistent performance across different settings (Table \ref{tab:spearman_results} in Appendix~\ref{sec:appendix-privacy-results}). 
Note that as $k$ increase there is less and less rank correlation with LiRA. 
This outcome is likely due to the same number of shadow models being used for all $k$, but further investigation is needed to confirm this.

In terms of execution time, t-WaKA demonstrates significant efficiency compared to LiRA. 
We report the execution times for attacking the IMDB dataset across all values of $k$ from 1 to 5.
IMDB is the largest dataset we experimented with, featuring s-BERT embeddings of size 383, in contrast to the custom CIFAR10 embeddings, which are of size 191. 
For this dataset, t-WaKA completed the experiment in approximately 140.07 seconds, significantly faster than LiRA, which took 1708.83 seconds. 
Both methods were run on an Apple M1 Pro (16 GB, 8 cores), utilizing all available cores through a single Python process managed by scikit-learn.
LiRA employed 16 shadow $k$-NN models using kd-trees, while t-WaKA reused the same $k$-NN model across all security games. 
For fairness that, re LiRA’s implementation is model-agnostic and not optimized specifically for $k$-NN classifiers.

To better understand the distinction between high self-attribution points and high test-attribution points, we conducted an experiment involving data minimization on synthetic data (see Figure~\ref{fig:synthetic-exp} in Appendix~\ref{sec:appendix-privacy-results}) using a dataset generated with the \texttt{scikit-learn} library.
Both test-Shapley and self-Shapley values were computed for a $k$-NN model with $k=5$. 
Each iteration consists in removing the top 20\% of highest value points, ranked by their Shapley values. 
The results highlights a clear distinction between the two types of Shapley values. 
More precisely, test-Shapley values tend to concentrate between the decision boundary and the outer edges of the data distribution while self-Shapley values consistently highlight points that lie directly on the decision boundary. 
This underscores the effectiveness of self-Shapley values in identifying critical points near the decision boundary, which seem to be more vulnerable to membership inference attacks in $k$-NNs.

\section{Discussion and Conclusion}
\label{sec:discussion}

In this work, we have introduced WaKA, a novel attribution method specifically designed for $k$-NN classifiers that also functions as a MIA. 
WaKA leverages the 1-Wasserstein distance to efficiently assess the contribution of individual data points to the model's loss distribution. 
Our motivation for this paper stemmed from the realization that while membership inference and data attribution both focus on point-wise contributions, they are related to very different concepts, namely privacy and value. 
Indeed, membership inference aims to measure information leakage, whereas data valuation seeks to extract intrinsic insights about data utility.
Recognizing this distinction, we have design an attribution method inspired by membership inference principles, one that could serve multiple purpose. 
The insights from this study emphasize the need for further exploration of the relationship between model parameters, data attribution and privacy.

\textbf{Data Valuation.} WaKA proves to be highly effective for utility-driven data minimization by identifying both low- and high-value points.
This targeted approach allows for optimized dataset refinement. Our experiments further demonstrate WaKA’s robustness over Data Shapley Value (DSV) in the context of imbalanced datasets. 
This makes WaKA a versatile and reliable tool for data valuation.

\textbf{Self-attribution methods.} Due of its relationship with MIAs, WaKA is particularly interesting as a self-attribution method. 
We also introduced the self-Shapley term as the DSV of a point with respect to itself, a concept that has not been discussed in the literature to date. 
From an utilitarian point of view, self-Shapley value could be particularly appealing to individuals who prioritize their own utility over the collective utility captured by average DSV calculated using a test set, especially when their main interest is whether contributing their data directly improves their own prediction rather than the model’s overall performance. 
For example, in the context of a financial institution, an individual may only be willing to provide their data if it directly improves the accuracy of predictions that affect them personally, such as determining their creditworthiness. 
Self-Shapley offers individuals a way to know if their data would be beneficial for them only, making it a valuable tool in cases where personal utility is prioritized over collective fairness (\emph{i.e.}, favoring a greedy strategy). 
Self-WaKA correlates with self-Shapley, and thus it can be used for similar purposes. 
However, additionally Self-WaKA also brings a more direct rationale concerning membership privacy, due to its relation with LiRA. 
This demonstrates that privacy is not always directly correlated with value, as it largely depends on how value is defined. 
In the case of $k$-NN, we showed it is more closely tied to ``self-utility'' rather than overall contribution to the model's generalization performance.

\textbf{Privacy insights and the onion effect.} In our evaluation, we confirmed the ``onion effect'' previously described by Carlini and co-authors, in which removing certain data points only partially mitigates the risk of MIAs, leaving the remaining data still vulnerable. 
WaKA, not only corroborates this phenomenon but also provides an efficient approach to measuring privacy risk in $k$-NN models. 
This aligns with recent work by \cite{ye2022enhanced}, which underscores that membership inference risk is influenced not only by individual data points but also by their neighborhood. 
If $k$-NN is used as part of a Machine Learning pipeline—such as combining embeddings with $k$-NN for downstream tasks—WaKA can offer a way to understand membership privacy risks.
We leave as future work the exploration to whether self-WaKA values and WaKA influences can generalize to other types of models, such as interpreting privacy risks in the last layer of a large language model (LLM) or other deep learning architectures.

\textbf{Ethical Considerations.}
The development of WaKA raises ethical concerns. Adversaries could potentially exploit WaKA to selectively identify and target particularly vulnerable data points, either to enhance inference attacks or to manipulate datasets to maximize privacy leakage. However, WaKA also provides defensive capabilities, enabling practitioners to identify and remove high-risk data points or design privacy-aware data-selection strategies. It is essential that attribution insights derived from methods like WaKA be leveraged responsibly, aiming to strengthen privacy and avoid introducing new vulnerabilities.

\textbf{Protection against membership inference attack.} Our experiments reveal that increasing the  parameter $k$ in $k$-NN models has a significant effect on reducing the success rate of MIAs. 
We believe that this issue is under-explored in the current literature and warrants further investigation. 
In particular, the hybrid models employed in our study, such as the combination of s-BERT and $k$-NN used on the IMDB and Yelp datasets, suggest that the parameter $k$ could be strategically use as a defense mechanism against membership inference.

In addition, we believe that WaKA values could provide valuable insights for data removal task for other types of models, such as neural networks. 
While our current research focuses on $k$-NN classifiers, we will also investigate the extension of this approach to neural networks as future work. 
In particular, testing WaKA's applicability to these contexts could significantly enhance the understanding of data attribution and its impact on model robustness and privacy.

\begin{acks}
We thank the anonymous reviewers for their valuable feedback and suggestions. We also want to acknowledge that ChatGPT was used solely for rephrasing sentences as a grammatical tool.
\end{acks}

\bibliographystyle{ACM-Reference-Format}
\bibliography{main}

\appendix
\onecolumn

\section{Datasets}
\label{sec:datasets}

\begin{table*}[h!]
\centering
\caption{Description of the datasets used in the experiments}
\renewcommand{\arraystretch}{1.5} 
\begin{tabular}{|p{2cm}|p{1.5cm}|p{3cm}|p{1.5cm}|p{1.5cm}|p{2.5cm}|p{2cm}|}
\hline
\textbf{Dataset} & \textbf{Type} & \textbf{Features} & \textbf{Training Size} & \textbf{Test Size} & \textbf{Encoding} & \textbf{Reference} \\ \hline
CIFAR-10         & Image         & 32x32 images in 10 classes        & 25,000                & 5,000                & Custom pre-trained            & \cite{krizhevsky2009learning}     \\ \hline
Bank             & Tabular       & 17 features                       & 36,168                & 9,043                & One-hot and normalized & \cite{moro2014data}               \\ \hline
Adult            & Tabular       & 14 features                       & 32,561                & 16,281               & One-hot and normalized & \cite{kohavi1996scaling}          \\ \hline
IMDB             & Text          & Avg. 230-240 words per document   & 25,000                & 5,000                & SBERT pre-trained      & \cite{maas-EtAl:2011:ACL-HLT2011}           \\ \hline
Yelp             & Text          & Avg. 150 words per review         & 25,000               & 5,000                & SBERT pre-trained      & \cite{zhang2015character}         \\ \hline
CelebA           & Image         & 178x218 images with 40 attributes & 25,000                & 5,000                & ViT pre-trained            & \cite{liu2018large}      \\ \hline
\end{tabular}
\label{tab:datasets}
\end{table*}

\section{Notations}
\begin{table}[h!]
\centering
\begin{tabular}{|c|p{12cm}|}
\hline
\textbf{Symbol} & \textbf{Description} \\ \hline
$W_1(\mu, \nu)$ & 1-Wasserstein distance  between two probability distributions $\mu$ and $\nu$. \\ \hline
$\mathbb{L}$ & Distribution of loss values for $k$-NN models trained on all possible subsets of the dataset $D$. \\ \hline
$\mathbb{L}_{-z_i}$ & Distribution of loss values for $k$-NN models trained on subsets of $D$ excluding point $z_i$. \\ \hline
$\mathcal{L}$ & Finite set of possible discrete loss values for a $k$-NN classifier. \\ \hline
$\mathcal{F}$ & Space of all $k$-NN models trained on subsets of $D$. \\ \hline
$\mathcal{F}_{-z_i}$ & Space of all $k$-NN models trained on subsets of $D$ excluding point $z_i$. \\ \hline
$\mathbb{Q}$ & Uniform distribution over models in $\mathcal{F}$ trained on subsets $S$ such that $z_i \in S$. \\ \hline
$\mathbb{Q}_{-z_i}$ & Uniform distribution over models in $\mathcal{F}$ trained on subsets $S$ such that $z_i \notin S$. \\ \hline
$\ell(f)$ & Loss function mapping a $k$-NN model $f$ to a real-valued loss. \\ \hline
$\mathbb{\ell_\#Q}$ & Pushforward distribution of $\mathbb{Q}$ through the loss function $\ell$. \\ \hline
$\mathbb{\ell_\#Q}_{-z_i}$ & Pushforward distribution of $\mathbb{Q}_{-z_i}$ through the loss function $\ell$. \\ \hline

$F_{\mathbb{\ell_\#Q}}(l)$ & Cumulative distribution function (CDF) of the loss distribution $\mathbb{L}$, evaluated at the loss value $l$. \\ \hline
$F_{\mathbb{\ell_\#Q}_{-z_i}}(l)$ & Cumulative distribution function (CDF) of the loss distribution $\mathbb{L}_{-z_i}$, evaluated at the loss value $l$. \\ \hline
$\mathcal{F}^+, \mathcal{F}^-$ & Partition of $\mathcal{F}$ into subsets with losses greater than or equal to, or less than target loss $\ell(z_i)^*$, respectively. \\ \hline
$\mathcal{F}_{-z_i}^+, \mathcal{F}_{-z_i}^-$ & Partition of $\mathcal{F}_{-z_i}$ into subsets with losses greater than or equal to, or less than target loss $\ell(z_i)^*$, respectively. \\ \hline
$\WaKA(z_i)$ & Wasserstein $k$-NN Attribution (WaKA) score for point $z_i$, measuring the 1-Wasserstein distance between $\mathbb{L}$ and $\mathbb{L}_{-z_i}$. \\ \hline
$\tWaKA(z_i)$ & Target-WaKA score for point $z_i$, measuring the difference in 1-Wasserstein distances for the partitions $\mathcal{F}^+$ and $\mathcal{F}^-$. \\ \hline
\end{tabular}
\end{table}

\section{Algorithm \ref{alg:waka_membership_attack} Details and Proof}
\label{sec:appendix-algorithm-details}

To compute the differences between the loss distributions \(\mathbb{L}\) and \(\mathbb{L}_{-z_i}\), we first need to calculate the number of possible subsets \(S\) of the training set \(D\) that lead to a particular loss value \(l\). The loss is computed using the \(k\)-nearest neighbors of a test point \(z_t\). 

For each subset \(S\), we are interested in how the inclusion of the point \(z_i\) affects the loss, which happens when \(z_i\) pushes out a point \(z_j\) from the \(k\)-th position to the \((k+1)\)-th position, i.e. $\alpha_{j}(D) > \alpha_i(D)$. 

To formalize this, we define the count \(C_{-z_i}(l, z_j)\) for subsets that exclude \(z_i\) and yield a loss \(l\) as:

\[
C_{-z_i}(l, z_j) = \#\left\{ S \subseteq D \mid z_i \notin S, \alpha_k(S) = \alpha_j(D), \ell(z_t; S, k) = l \right\} \cdot 2^{N-1-j}.
\]

where:
\begin{itemize}
    \item \(\alpha_k(S) = \alpha_j(D)\) means that the point \(z_j\) is both the j-th sorted point w.r.t $z_t$ and the \(k\)-th nearest neighbor in the subset \(S\),
    \item \(\ell(z_t; S, k)\) is the loss function parameterized by the subset \(S\) and the number of neighbors \(k\),
    \item The factor \(2^{N-1-j}\) accounts for the number of possible supersets of \(S\), considering the positions of other points in the training set.
\end{itemize}

Similarly, \(C(l, z_j)\) counts the subsets where \(z_i\) is included, with \(z_j\) now at \((k+1)\)-th position, and the loss is \(l\).

\[
C(l, z_j) = \#\left\{ S \subseteq D \mid z_i \in S, \alpha_{k+1}(S) = \alpha_j(D), \ell(z_t; S, k) = l \right\} \cdot 2^{N-j},
\]

Note that we are considering the same subsets as before but with the inclusion of $z_i$. The normalized contribution to the frequency of the loss value \(l\), denoted as \(\delta(l, z_j)\) and \(\delta_{-z_i}(l, z_j)\), is obtained by dividing \(C(l, z_j)\) and \(C_{-z_i}(l, z_j)\) by the total number of possible subsets. Specifically:

\[
\delta(l, z_j) = \frac{C(l, z_j)}{2^{N}}, \quad \delta_{-z_i}(l, z_j) = \frac{C_{-z_i}(l, z_j)}{2^{N-1}}.
\]

To compare the differences between the loss distributions \(\mathbb{L}\) and \(\mathbb{L}_{-z_i}\), we sum the differences in the normalized contributions of the loss values in both cases:

\[
\sum_{l \in \mathcal{L}} \sum_{\substack{z_j}} {\delta(l, z_j) - \delta_{-z_i}(l, z_j)} 
\]

Here, \(\mathcal{L}\) is the set of possible loss values \(\left\{0, \frac{1}{k}, \frac{2}{k}, \dots, 1 \right\}\), and the summation over \(z_j\) accounts for all points that can occupy the \(k\)-th nearest neighbor position in the subsets.

We simplify the term inside the summation as follows:

\begin{align}
\frac{C \cdot 2^{N-j}}{2^{N}} - \frac{C_{-z_i} \cdot 2^{N-1-j}}{2^{N-1}} 
&= \frac{1}{2^{N-1}} \cdot \left( \frac{C}{2} \cdot 2^{N-j} - C_{-z_i} \cdot 2^{N-1-j} \right) \\
&= \frac{1}{2^{N-1}} \left( C \cdot 2^{N-1-j} - C_{-z_i} \cdot 2^{N-1-j} \right) \\
&= \frac{2^{N-1-j}}{2^{N-1}} \left( C - C_{-z_i} \right) \\
&= \frac{C - C_{-z_i}}{2^j}.
\end{align}

Thus, the difference between the normalized contributions simplifies to \(\frac{C - C_{-z_i}}{2^j}\), where \(C\) and \(C_{-z_i}\) are shorthand notations for the counts of subsets for a given loss value \(l\) with and without \(z_i\), and a given point \(z_j\). Notice that when \(z_j\) and \(z_i\) have the same label, i.e. $y_j = y_i$, the $k$-nearest neighbors are identical, thus $C-C_{-z_i} = 0$.

By applying this simplification to the previous expression, we have:

\[
\sum_{l \in \mathcal{L}} \sum_{\substack{z_j}} {\delta(l, z_j) - \delta_{-z_i}(l, z_j)}  = \sum_{l \in \mathcal{L}} \sum_{\substack{z_j \\ y_i \neq y_j}} \frac{C(l, z_j) - C_{-z_i}(l, z_j)}{2^j}.
\]

The 1-Wasserstein distance between the loss distributions \(\mathbb{L}\) and \(\mathbb{L}_{-z_i}\) is defined as:

\[
W_1(\mathbb{L}, \mathbb{L}_{-z_i}) = \sum_{l \in \mathcal{L}} \left| F_{\mathbb{L}}(l) - F_{\mathbb{L}_{-z_i}}(l) \right| \cdot \Delta l,
\]

where \(F_{\mathbb{L}}(l)\) and \(F_{\mathbb{L}_{-z_i}}(l)\) represent the cumulative distribution functions for \(\mathbb{L}\) and \(\mathbb{L}_{-z_i}\) evaluated at the loss value \(l\), and \(\Delta l = \frac{1}{k}\) is the difference between successive loss values.

We know that the cumulative distribution functions \(F_{\mathbb{L}}(l)\) and \(F_{\mathbb{L}_{-z_i}}(l)\) can be described as the sum of their respective normalized contributions up to the loss value \(l\). However, since we are ultimately interested in the differences between the normalized contributions \(\delta(l, z_j)\) and \(\delta_{-z_i}(l, z_j)\) (rather than computing the cumulative distributions explicitly), we can substitute the expression for the difference of these terms directly into the Wasserstein distance formula.

Therefore, instead of expressing the 1-Wasserstein distance in terms of cumulative distributions, we can express it directly as the sum of the absolute differences in the normalized contributions:

\[
W_1(\mathbb{L}, \mathbb{L}_{-z_i}) = \frac{1}{k} \sum_{l \in \mathcal{L}} \left| \sum_{z_j} \left( \delta(l, z_j) - \delta_{-z_i}(l, z_j) \right) \right|.
\]

Substituting the earlier expression for \(\delta(l, z_j) - \delta_{-z_i}(l, z_j)\), we get:

\[
W_1(\mathbb{L}, \mathbb{L}_{-z_i}) = \frac{1}{k} \sum_{l \in \mathcal{L}} \left| \sum_{\substack{z_j \\ y_i \neq y_j}} \frac{C(l, z_j) - C_{-z_i}(l, z_j)}{2^j} \right|.
\]

\section{Utility-driven Data Minimization Results and Analysis}
\label{sec:appendix-utility-results}

\begin{figure*}[h!]
    \centering
    \caption{Results for the CIFAR10 dataset across various metrics and tasks. Rows show data addition and removal tasks, while columns display Accuracy, Macro-F1, and Label Ratio Evolution (for data removal only).}
    \label{fig:cifar10-results}
    \begin{tikzpicture}[baseline=(current bounding box.north)]
        \node at (3, 8) {Accuracy};
        \node at (8, 8) {Macro-F1};
        \node at (13, 8) {Label Ratio Evolution};
        
        \node[rotate=90] at (-0.5, 5) {Data Addition};
        \node[rotate=90] at (-0.5, -0.5) {Data Removal};
        
        \node[anchor=north west, inner sep=0, outer sep=0] (image1) at (0, 7.5) {\includegraphics[width=0.3\linewidth]{images/CIFAR/Utility/accuracy_ac.png}};
        \node[anchor=north west, inner sep=0, outer sep=0] (image2) at (5, 7.5) {\includegraphics[width=0.3\linewidth]{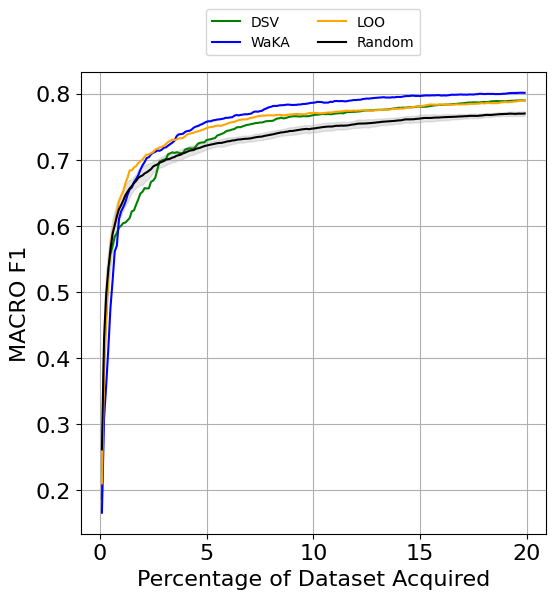}};
        
        \node[anchor=north west, inner sep=0, outer sep=0] (image4) at (0, 2.5) {\includegraphics[width=0.3\linewidth]{images/CIFAR/Utility/accuracy_re.png}};
        \node[anchor=north west, inner sep=0, outer sep=0] (image5) at (5, 2.5) {\includegraphics[width=0.3\linewidth]{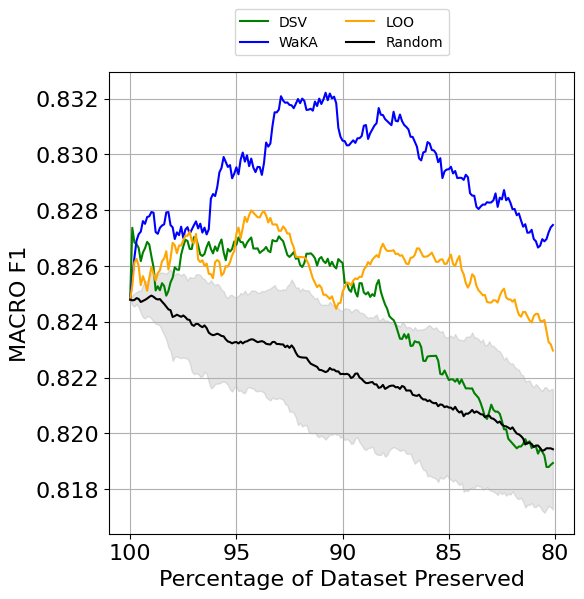}};
        \node[anchor=north west, inner sep=0, outer sep=0] (image6) at (10, 2.5) {\includegraphics[width=0.3\linewidth]{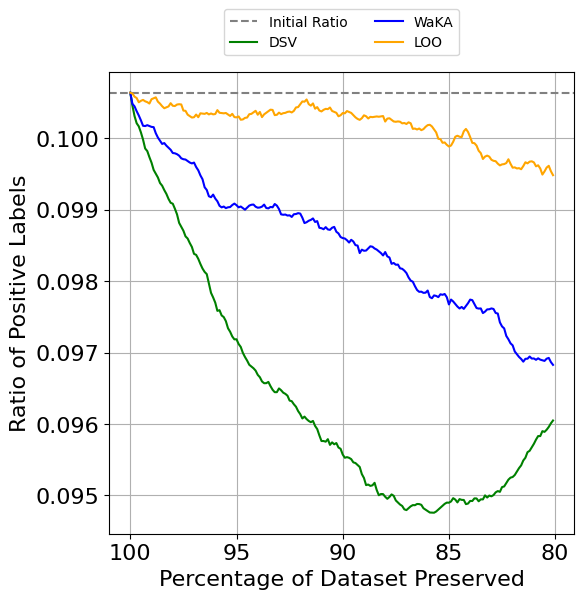}};
    \end{tikzpicture}
\end{figure*}

\begin{figure}[h!]
    \centering
    \caption{Analysis of $\tau$ values for the CIFAR10 dataset computed using the $\WaKAadd$ and $\WaKArem$ formulas. $\tau$ is varied from 0.0 to 1.0 in increments of 0.2.}
    \label{fig:cifar10-tau-analysis}
    \begin{tikzpicture}[node distance=1cm, baseline=(current bounding box.north)]
        \node (addImg) {\includegraphics[width=0.35\linewidth]{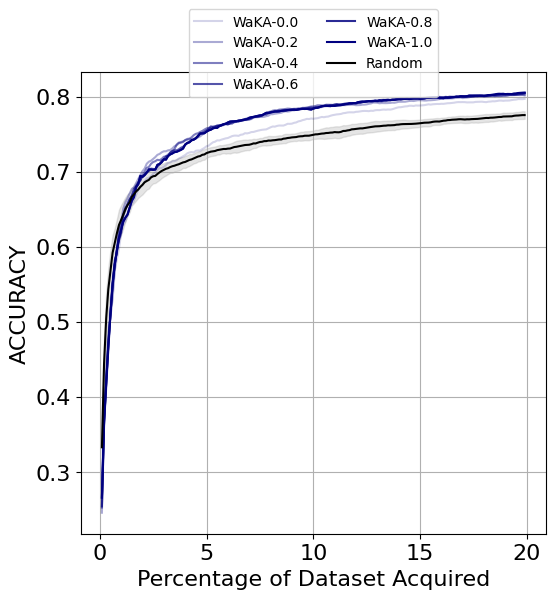}};
        \node[above=0.3cm of addImg] {Data Addition};
        
        \node (remImg) [right=2cm of addImg] {\includegraphics[width=0.37\linewidth]{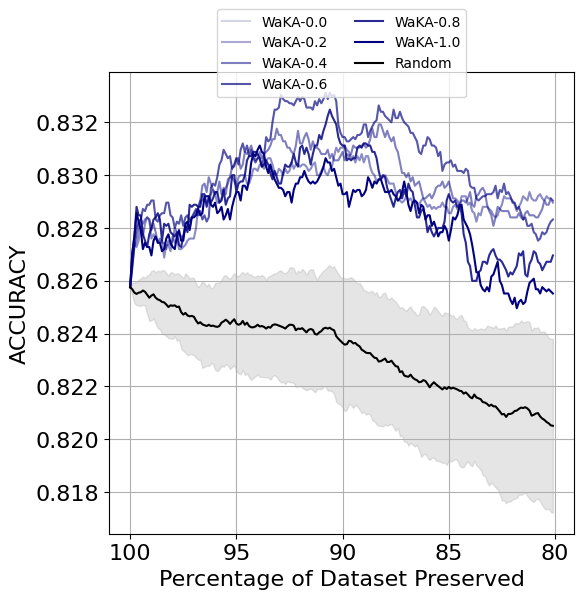}};
        \node[above=0.3cm of remImg] {Data Removal};
    \end{tikzpicture}
\end{figure}
\begin{figure*}[h!]
    \centering
    \caption{Results for the Adult dataset across various metrics and tasks. Rows show data addition and removal tasks, while columns display Accuracy, Macro-F1, and Label Ratio Evolution (for data removal only).}
    \label{fig:adult-results}
    \begin{tikzpicture}[baseline=(current bounding box.north)]
        \node at (3, 8) {Accuracy};
        \node at (8, 8) {Macro-F1};
        \node at (13, 8) {Label Ratio Evolution};
        
        \node[rotate=90] at (-0.5, 5) {Data Addition};
        \node[rotate=90] at (-0.5, -0.5) {Data Removal};
        
        \node[anchor=north west, inner sep=0, outer sep=0] (image1) at (0, 7.5) {\includegraphics[width=0.3\linewidth]{images/Adult/Utility/accuracy_ac.png}};
        \node[anchor=north west, inner sep=0, outer sep=0] (image2) at (5, 7.5) {\includegraphics[width=0.3\linewidth]{images/Adult/Utility/macro_f1_ac.png}};
        
        \node[anchor=north west, inner sep=0, outer sep=0] (image4) at (0, 2.5) {\includegraphics[width=0.3\linewidth]{images/Adult/Utility/accuracy_re.png}};
        \node[anchor=north west, inner sep=0, outer sep=0] (image5) at (5, 2.5) {\includegraphics[width=0.3\linewidth]{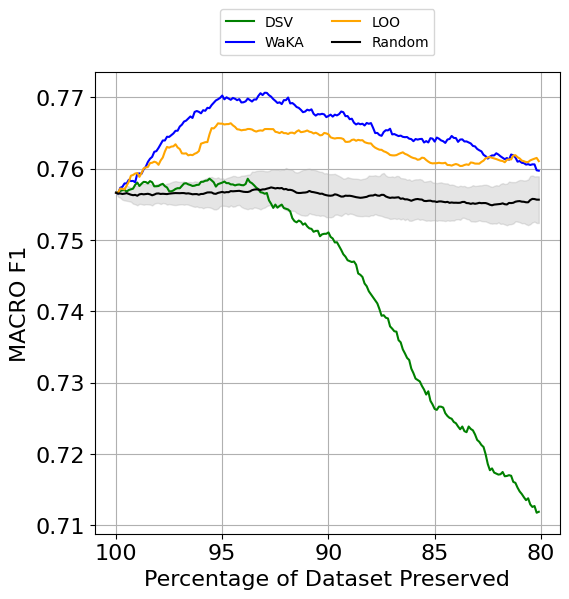}};
        \node[anchor=north west, inner sep=0, outer sep=0] (image6) at (10, 2.5) {\includegraphics[width=0.3\linewidth]{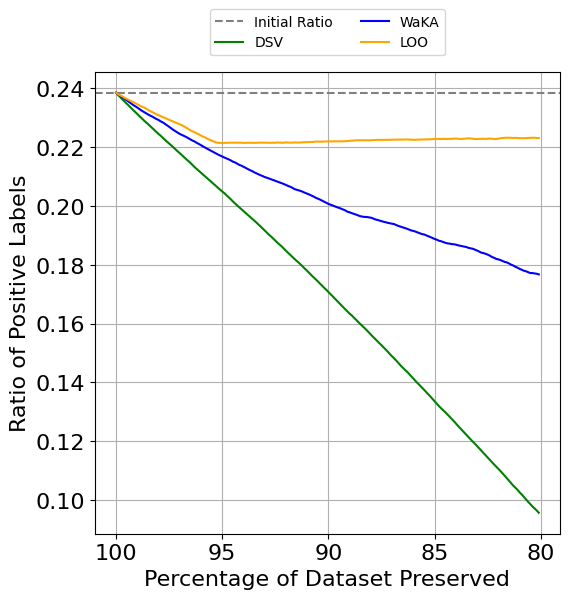}};
    \end{tikzpicture}
\end{figure*}

\begin{figure}[h!]
    \centering
    \caption{Analysis of $\tau$ values for the Adult dataset computed using the $\WaKAadd$ and $\WaKArem$ formulas. $\tau$ is varied from 0.0 to 1.0 in increments of 0.2.}
    \label{fig:cifar10-tau-analysis}
    \begin{tikzpicture}[node distance=1cm, baseline=(current bounding box.north)]
        \node (addImg) {\includegraphics[width=0.35\linewidth]{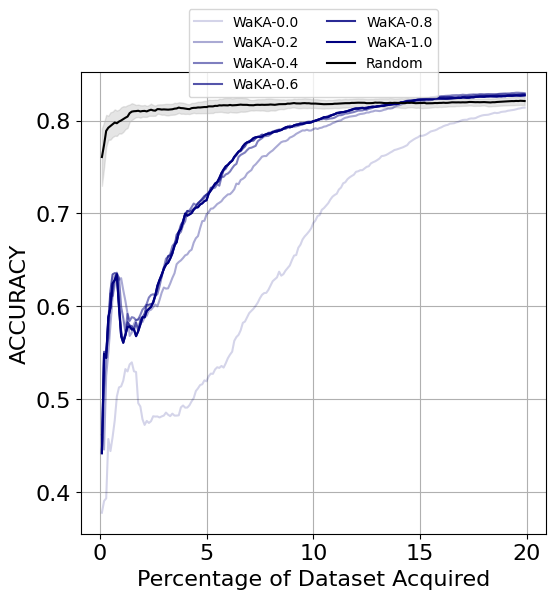}};
        \node[above=0.3cm of addImg] {Data Addition};
        
        \node (remImg) [right=2cm of addImg] {\includegraphics[width=0.37\linewidth]{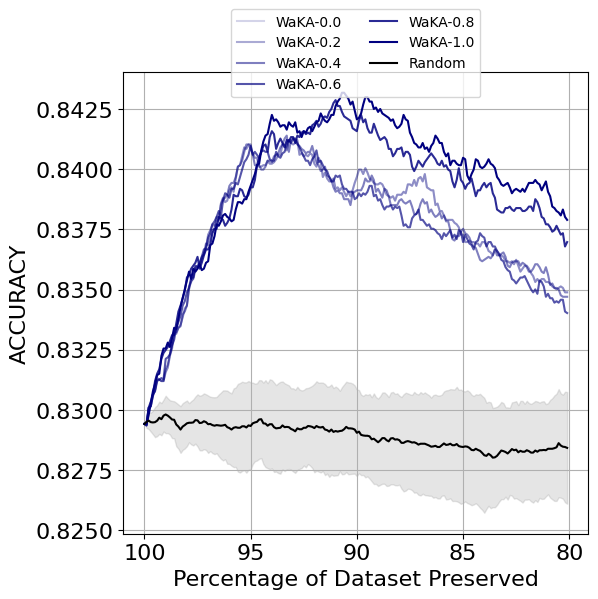}};
        \node[above=0.3cm of remImg] {Data Removal};
    \end{tikzpicture}
\end{figure}

\begin{figure*}[h!]
    \centering
    \caption{Results for the Yelp dataset across various metrics and tasks. Rows show data addition and removal tasks, while columns display Accuracy, Macro-F1, and Label Ratio Evolution (for data removal only).}
    \label{fig:yelp-results}
    \begin{tikzpicture}[baseline=(current bounding box.north)]
        \node at (3, 8) {Accuracy};
        \node at (8, 8) {Macro-F1};
        \node at (13, 8) {Label Ratio Evolution};
        
        \node[rotate=90] at (-0.5, 5) {Data Addition};
        \node[rotate=90] at (-0.5, -0.5) {Data Removal};
        
        \node[anchor=north west, inner sep=0, outer sep=0] (image1) at (0, 7.5) {\includegraphics[width=0.3\linewidth]{images/Yelp/Utility/accuracy_ac.png}};
        \node[anchor=north west, inner sep=0, outer sep=0] (image2) at (5, 7.5) {\includegraphics[width=0.3\linewidth]{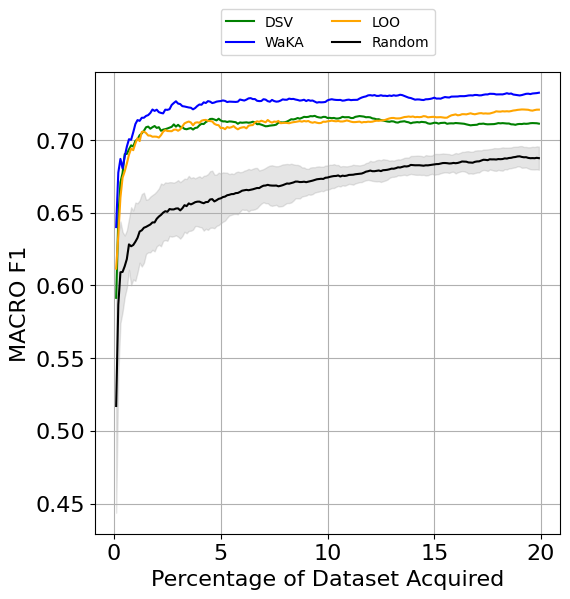}};
        
        \node[anchor=north west, inner sep=0, outer sep=0] (image4) at (0, 2.5) {\includegraphics[width=0.3\linewidth]{images/Yelp/Utility/accuracy_re.png}};
        \node[anchor=north west, inner sep=0, outer sep=0] (image5) at (5, 2.5) {\includegraphics[width=0.3\linewidth]{images/Yelp/Utility/macro_f1_re.png}};
        \node[anchor=north west, inner sep=0, outer sep=0] (image6) at (10, 2.5) {\includegraphics[width=0.3\linewidth]{images/Yelp/Utility/label_ratio_evolution.png}};
    \end{tikzpicture}
\end{figure*}

\begin{figure}[h!]
    \centering
    \caption{Analysis of $\tau$ values for the Yelp dataset computed using the $\WaKAadd$ and $\WaKArem$ formulas. $\tau$ is varied from 0.0 to 1.0 in increments of 0.2.}
    \label{fig:cifar10-tau-analysis}
    \begin{tikzpicture}[node distance=1cm, baseline=(current bounding box.north)]
        \node (addImg) {\includegraphics[width=0.35\linewidth]{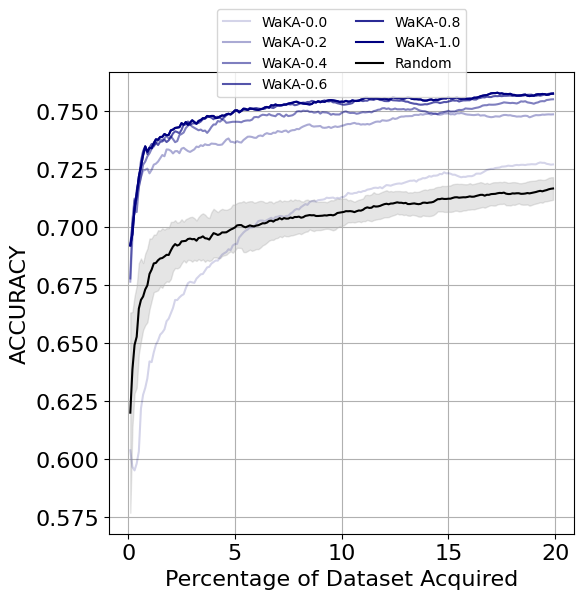}};
        \node[above=0.3cm of addImg] {Data Addition};
        
        \node (remImg) [right=2cm of addImg] {\includegraphics[width=0.37\linewidth]{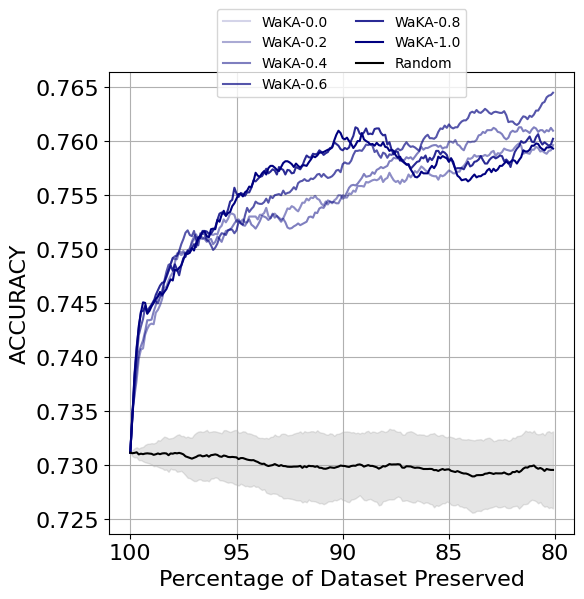}};
        \node[above=0.3cm of remImg] {Data Removal};
    \end{tikzpicture}
\end{figure}

\begin{figure*}[h!]
    \centering
    \caption{Results for the IMDB dataset across various metrics and tasks. Rows show data addition and removal tasks, while columns display Accuracy, Macro-F1, and Label Ratio Evolution (for data removal only).}
    \label{fig:imdb-results}
    \begin{tikzpicture}[baseline=(current bounding box.north)]
        \node at (3, 8) {Accuracy};
        \node at (8, 8) {Macro-F1};
        \node at (13, 8) {Label Ratio Evolution};
        
        \node[rotate=90] at (-0.5, 5) {Data Addition};
        \node[rotate=90] at (-0.5, -0.5) {Data Removal};
        
        \node[anchor=north west, inner sep=0, outer sep=0] (image1) at (0, 7.5) {\includegraphics[width=0.3\linewidth]{images/IMDB/Utility/accuracy_ac.png}};
        \node[anchor=north west, inner sep=0, outer sep=0] (image2) at (5, 7.5) {\includegraphics[width=0.3\linewidth]{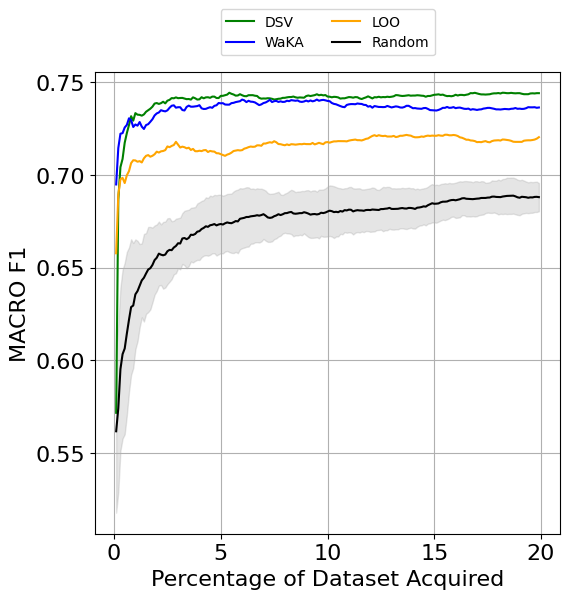}};
        
        \node[anchor=north west, inner sep=0, outer sep=0] (image4) at (0, 2.5) {\includegraphics[width=0.3\linewidth]{images/IMDB/Utility/accuracy_re.png}};
        \node[anchor=north west, inner sep=0, outer sep=0] (image5) at (5, 2.5) {\includegraphics[width=0.3\linewidth]{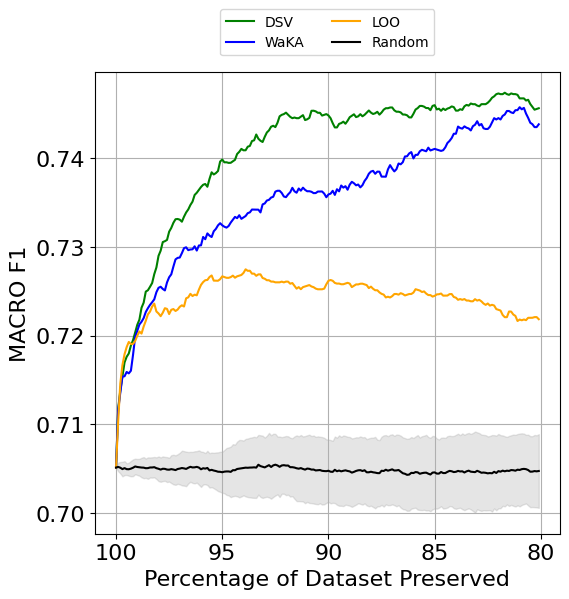}};
        \node[anchor=north west, inner sep=0, outer sep=0] (image6) at (10, 2.5) {\includegraphics[width=0.3\linewidth]{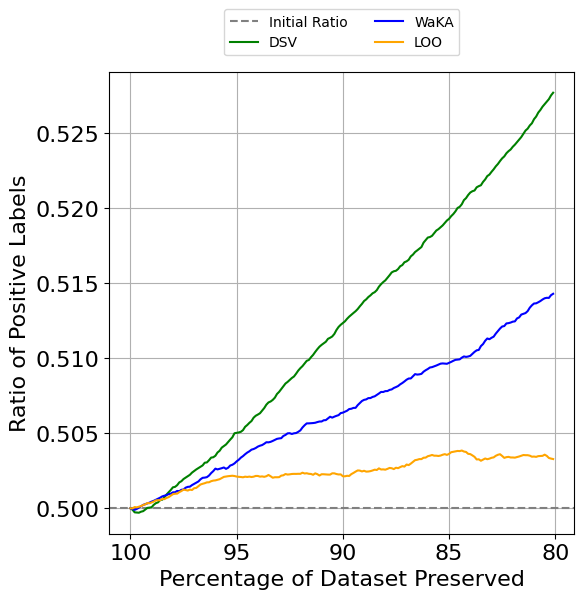}};
    \end{tikzpicture}
\end{figure*}

\begin{figure}[h!]
    \centering
    \caption{Analysis of $\tau$ values for the IMDB dataset computed using the $\WaKAadd$ and $\WaKArem$ formulas. $\tau$ is varied from 0.0 to 1.0 in increments of 0.2.}
    \label{fig:cifar10-tau-analysis}
    \begin{tikzpicture}[node distance=1cm, baseline=(current bounding box.north)]
        \node (addImg) {\includegraphics[width=0.35\linewidth]{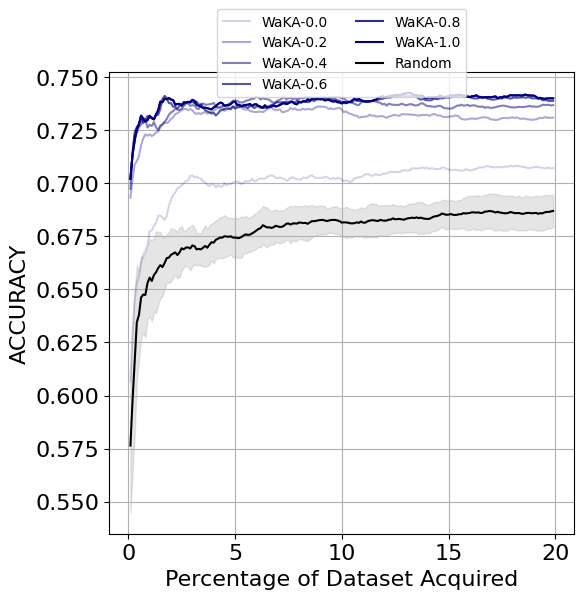}};
        \node[above=0.3cm of addImg] {Data Addition};
        
        \node (remImg) [right=2cm of addImg] {\includegraphics[width=0.37\linewidth]{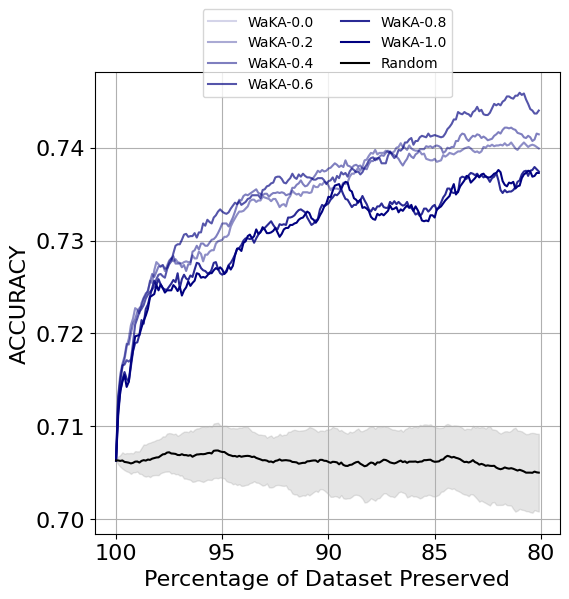}};
        \node[above=0.3cm of remImg] {Data Removal};
    \end{tikzpicture}
\end{figure}

\begin{figure*}[h!]
    \centering
    \caption{Results for the CelebA dataset across various metrics and tasks. Rows show data addition and removal tasks, while columns display Accuracy, Macro-F1, and Label Ratio Evolution (for data removal only).}
    \label{fig:celeba-results}
    \begin{tikzpicture}[baseline=(current bounding box.north)]
        \node at (3, 8) {Accuracy};
        \node at (8, 8) {Macro-F1};
        \node at (13, 8) {Label Ratio Evolution};
        
        \node[rotate=90] at (-0.5, 5) {Data Addition};
        \node[rotate=90] at (-0.5, -0.5) {Data Removal};
        
        \node[anchor=north west, inner sep=0, outer sep=0] (image1) at (0, 7.5) {\includegraphics[width=0.3\linewidth]{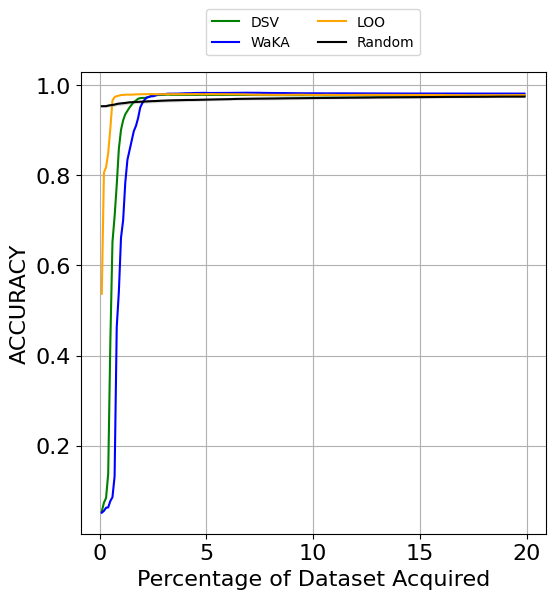}};
        \node[anchor=north west, inner sep=0, outer sep=0] (image2) at (5, 7.5) {\includegraphics[width=0.3\linewidth]{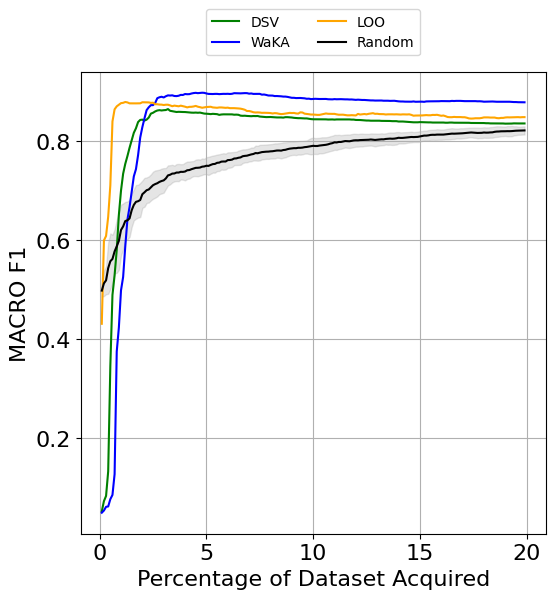}};
        
        \node[anchor=north west, inner sep=0, outer sep=0] (image4) at (0, 2.5) {\includegraphics[width=0.3\linewidth]{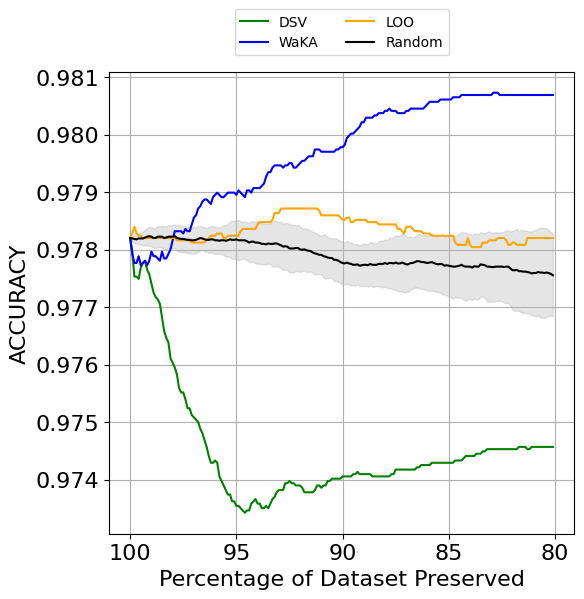}};
        \node[anchor=north west, inner sep=0, outer sep=0] (image5) at (5, 2.5) {\includegraphics[width=0.3\linewidth]{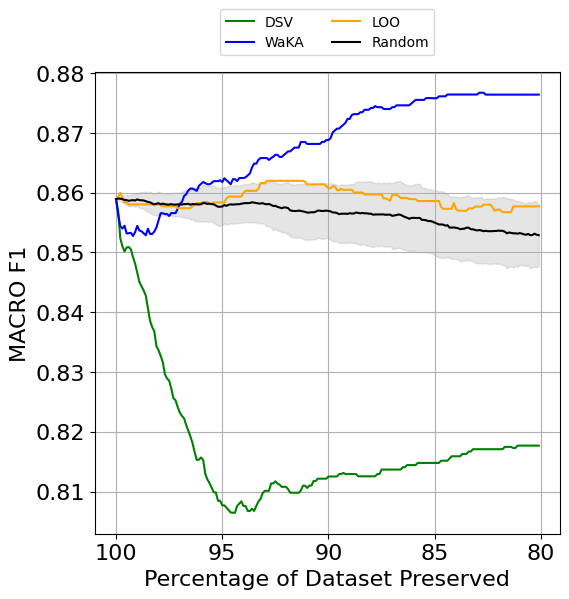}};
        \node[anchor=north west, inner sep=0, outer sep=0] (image6) at (10, 2.5) {\includegraphics[width=0.3\linewidth]{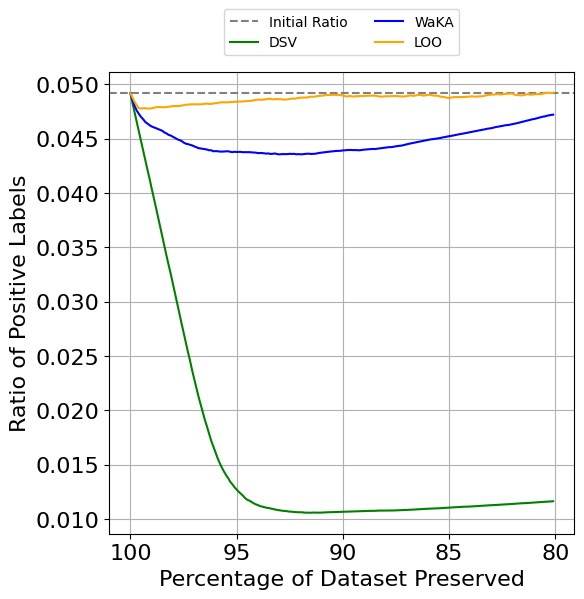}};
    \end{tikzpicture}
\end{figure*}

\begin{figure}[h!]
    \centering
    \caption{Analysis of $\tau$ values for the CelebA dataset computed using the $\WaKAadd$ and $\WaKArem$ formulas. $\tau$ is varied from 0.0 to 1.0 in increments of 0.2.}
    \label{fig:cifar10-tau-analysis}
    \begin{tikzpicture}[node distance=1cm, baseline=(current bounding box.north)]
        \node (addImg) {\includegraphics[width=0.35\linewidth]{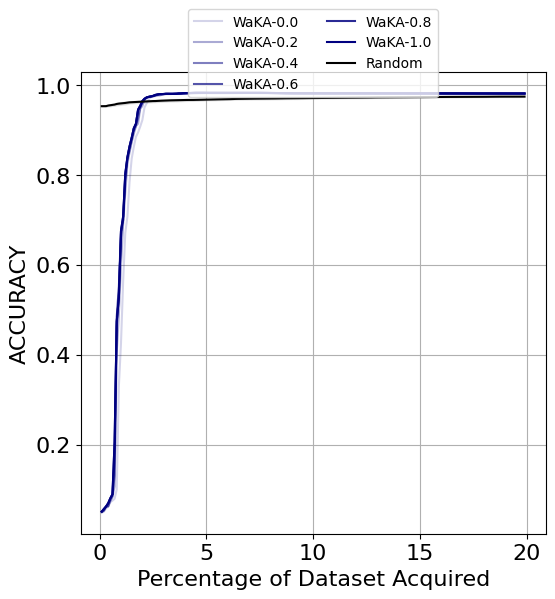}};
        \node[above=0.3cm of addImg] {Data Addition};
        
        \node (remImg) [right=2cm of addImg] {\includegraphics[width=0.37\linewidth]{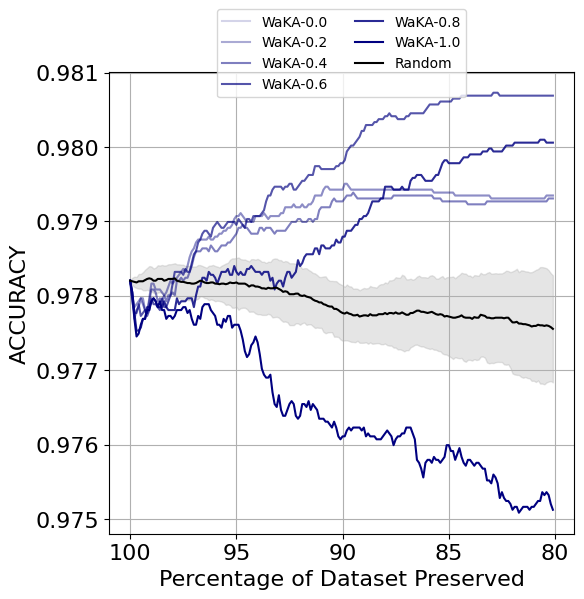}};
        \node[above=0.3cm of remImg] {Data Removal};
    \end{tikzpicture}
\end{figure}

\begin{figure*}[h!]
    \centering
    \caption{Results for the Bank dataset across various metrics and tasks. Rows show data addition and removal tasks, while columns display Accuracy, Macro-F1, and Label Ratio Evolution (for data removal only).}
    \label{fig:bank-results}
    \begin{tikzpicture}[baseline=(current bounding box.north)]
        \node at (3, 8) {Accuracy};
        \node at (8, 8) {Macro-F1};
        \node at (13, 8) {Label Ratio Evolution};
        
        \node[rotate=90] at (-0.5, 5) {Data Addition};
        \node[rotate=90] at (-0.5, -0.5) {Data Removal};
        
        \node[anchor=north west, inner sep=0, outer sep=0] (image1) at (0, 7.5) {\includegraphics[width=0.3\linewidth]{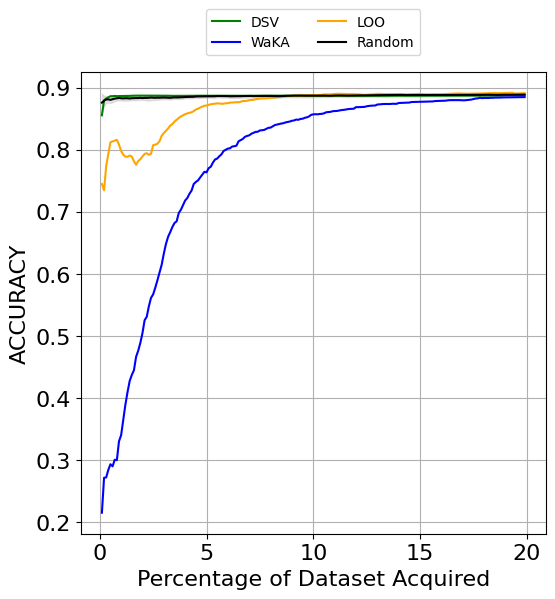}};
        \node[anchor=north west, inner sep=0, outer sep=0] (image2) at (5, 7.5) {\includegraphics[width=0.3\linewidth]{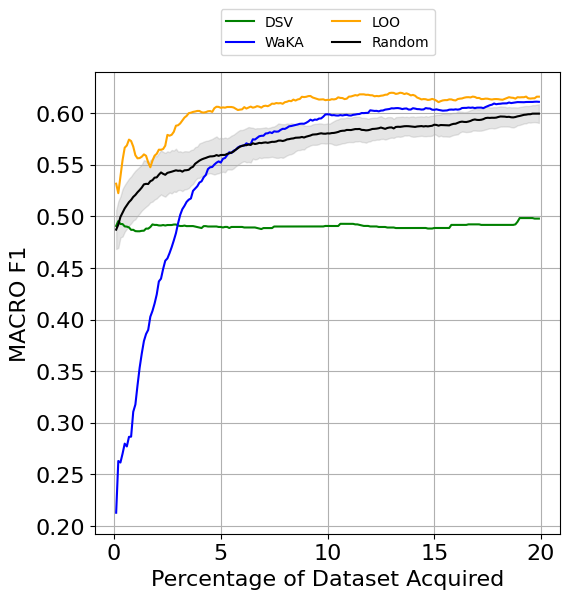}};
        
        \node[anchor=north west, inner sep=0, outer sep=0] (image4) at (0, 2.5) {\includegraphics[width=0.3\linewidth]{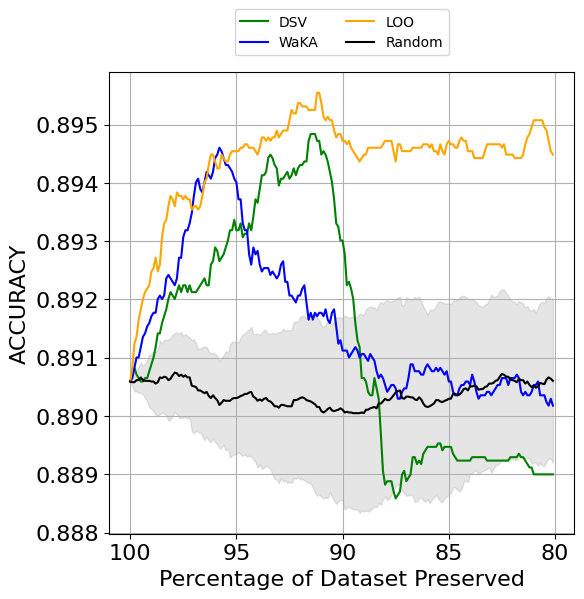}};
        \node[anchor=north west, inner sep=0, outer sep=0] (image5) at (5, 2.5) {\includegraphics[width=0.3\linewidth]{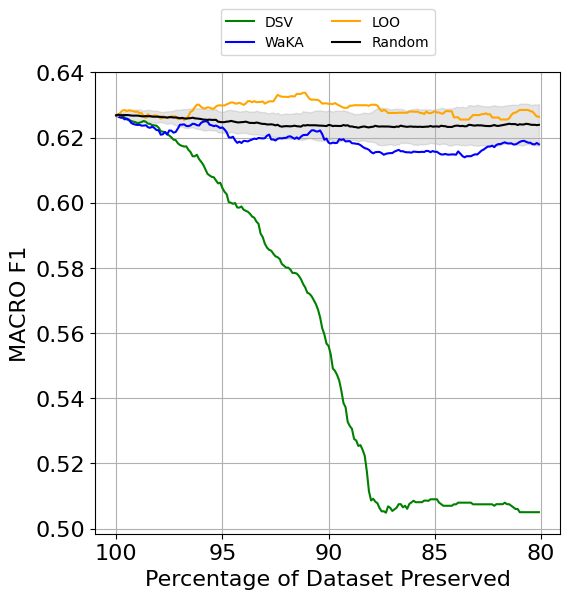}};
        \node[anchor=north west, inner sep=0, outer sep=0] (image6) at (10, 2.5) {\includegraphics[width=0.3\linewidth]{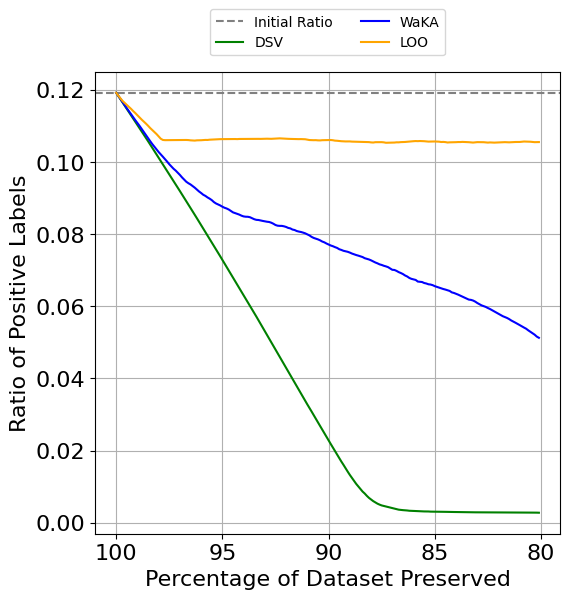}};
    \end{tikzpicture}
\end{figure*}
        
\begin{figure}[h!]
    \centering
    \caption{Analysis of $\tau$ values for the Bank dataset computed using the $\WaKAadd$ and $\WaKArem$ formulas. $\tau$ is varied from 0.0 to 1.0 in increments of 0.2.}
    \label{fig:cifar10-tau-analysis}
    \begin{tikzpicture}[node distance=1cm, baseline=(current bounding box.north)]
        \node (addImg) {\includegraphics[width=0.35\linewidth]{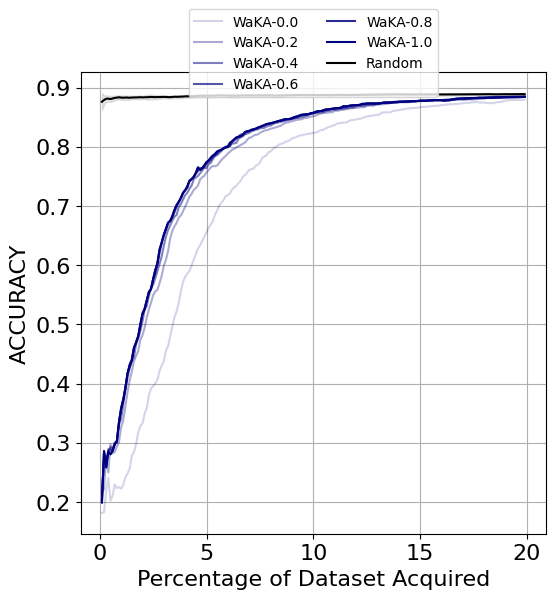}};
        \node[above=0.3cm of addImg] {Data Addition};
        
        \node (remImg) [right=2cm of addImg] {\includegraphics[width=0.37\linewidth]{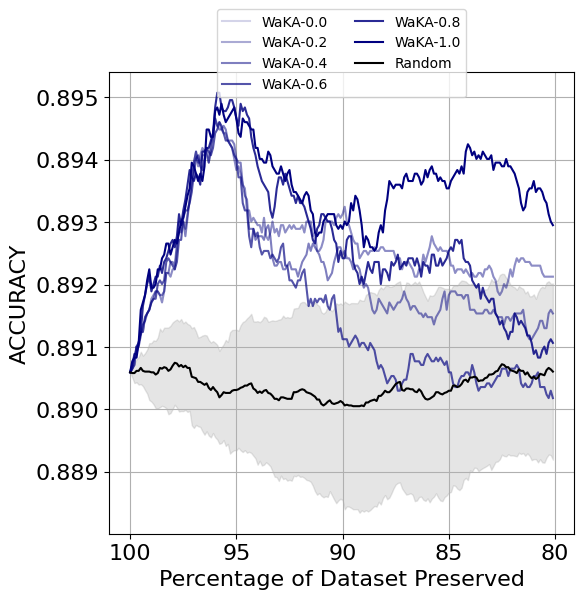}};
        \node[above=0.3cm of remImg] {Data Removal};
    \end{tikzpicture}
\end{figure}
\clearpage

\section{Privacy Evaluation and Auditing Results}
\label{sec:appendix-privacy-results}

\clearpage

   \begin{table}
    \caption{Spearman Correlation between self-WaKA and self-Shapley for K=1 and K=5}
    \begin{center}
    \begin{tabular}{|c|c|c|c|c|}
    \hline
    \textbf{Dataset} & \multicolumn{2}{|c|}{\textbf{K=1}} & \multicolumn{2}{|c|}{\textbf{K=5}} \\
    \cline{2-5} 
    & \textbf{Spearman} & \textbf{P-Value} & \textbf{Spearman} & \textbf{P-Value} \\
    \hline
    IMDB & 0.99 & 0.00 & 0.93 & 0.00 \\
\hline
CIFAR10 & 1.00 & 0.00 & 0.99 & 0.00 \\
\hline
adult & 0.99 & 0.00 & 0.98 & 0.00 \\
\hline
bank & 0.98 & 0.00 & 0.98 & 0.00 \\
\hline

    \end{tabular}
    \label{tab:table_waka_shap_spearman}
    \end{center}
    \end{table}

\begin{figure*}

    \centering
\caption{Correlation between self-WaKA values ($k=1$) and ASR for logistic regression models.
The results show a weaker correlation compared to $k$-NN models but a significant relationship still exists.}

    \begin{tikzpicture}[baseline=(current bounding box.north)]
        \node at (2.6, 7.5) {CIFAR10};
        \node at (7, 7.5) {Bank};
        \node at (11.5, 7.5) {Adult};
        \node at (15.5, 7.5) {IMDB};
        
        \node[anchor=north west, inner sep=0, outer sep=0] (image1) at (0, 7.2) {\includegraphics[width=0.27\linewidth]{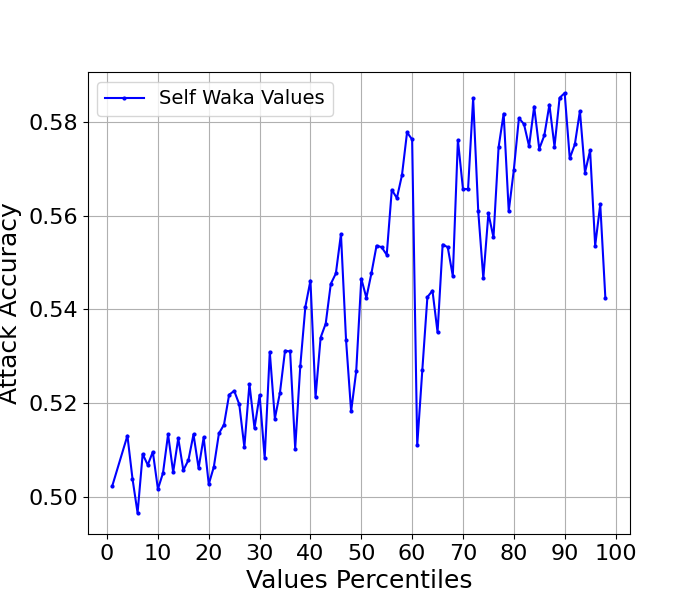}};
        \node[anchor=north west, inner sep=0, outer sep=0] (image2) at (4.5, 7.2) {\includegraphics[width=0.27\linewidth]{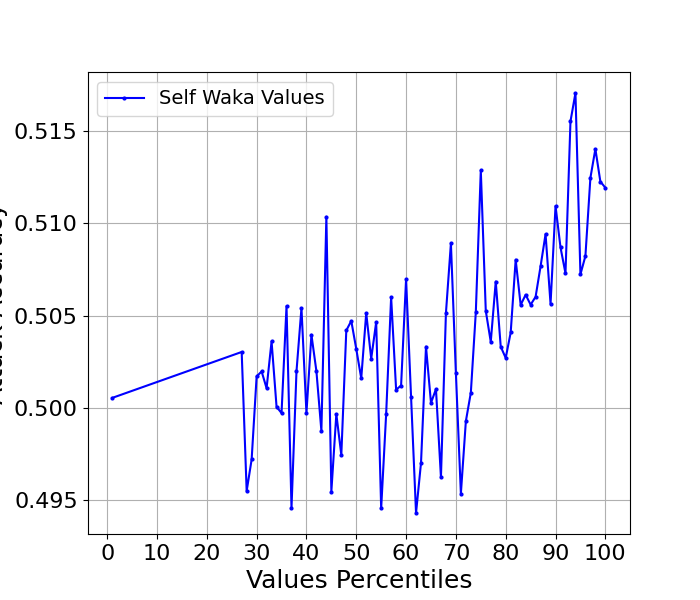}};
        \node[anchor=north west, inner sep=0, outer sep=0] (image3) at (9.0, 7.2) {\includegraphics[width=0.27\linewidth]{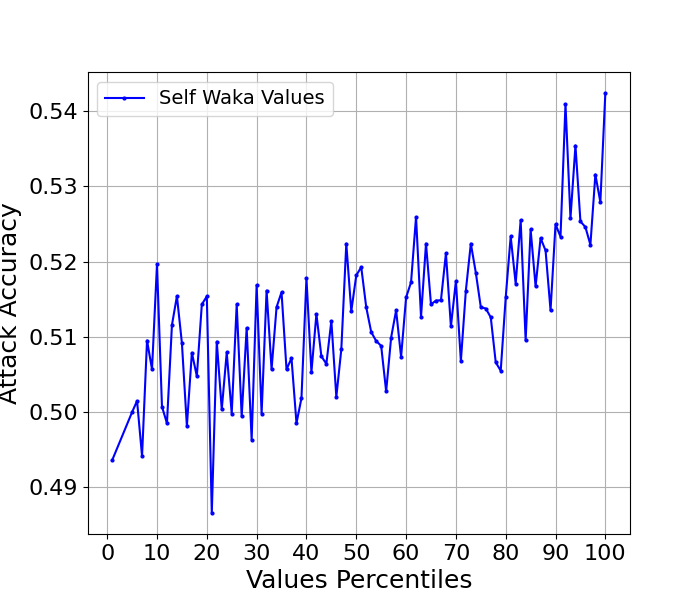}};
        \node[anchor=north west, inner sep=0, outer sep=0] (image4) at (13.4, 7.2) {\includegraphics[width=0.27\linewidth]{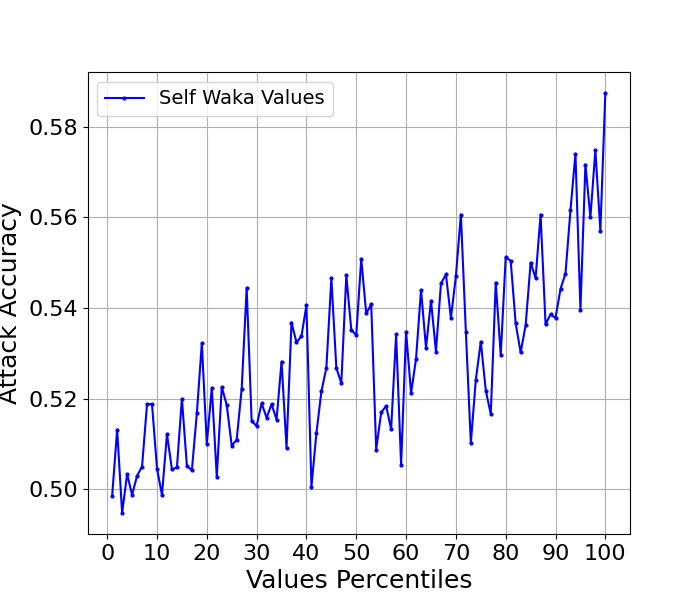}};
        
    \end{tikzpicture}

    \label{fig:privacy-eval-regression}

\end{figure*}

\begin{figure*}

    \centering
\caption{Comparison of AUC curves in log-log scale for privacy scores at \( k=1 \). 
The blue line represents the 100\% dataset scenario while the red line shows the 90\% dataset scenario.}

    \begin{tikzpicture}[baseline=(current bounding box.north)]
        \node at (2.6, 7.5) {CIFAR10};
        \node at (7, 7.5) {Bank};
        \node at (11.5, 7.5) {Adult};
        \node at (15.5, 7.5) {IMDB};
        
        \node[anchor=north west, inner sep=0, outer sep=0] (image1) at (0, 7.2) {\includegraphics[width=0.27\linewidth]{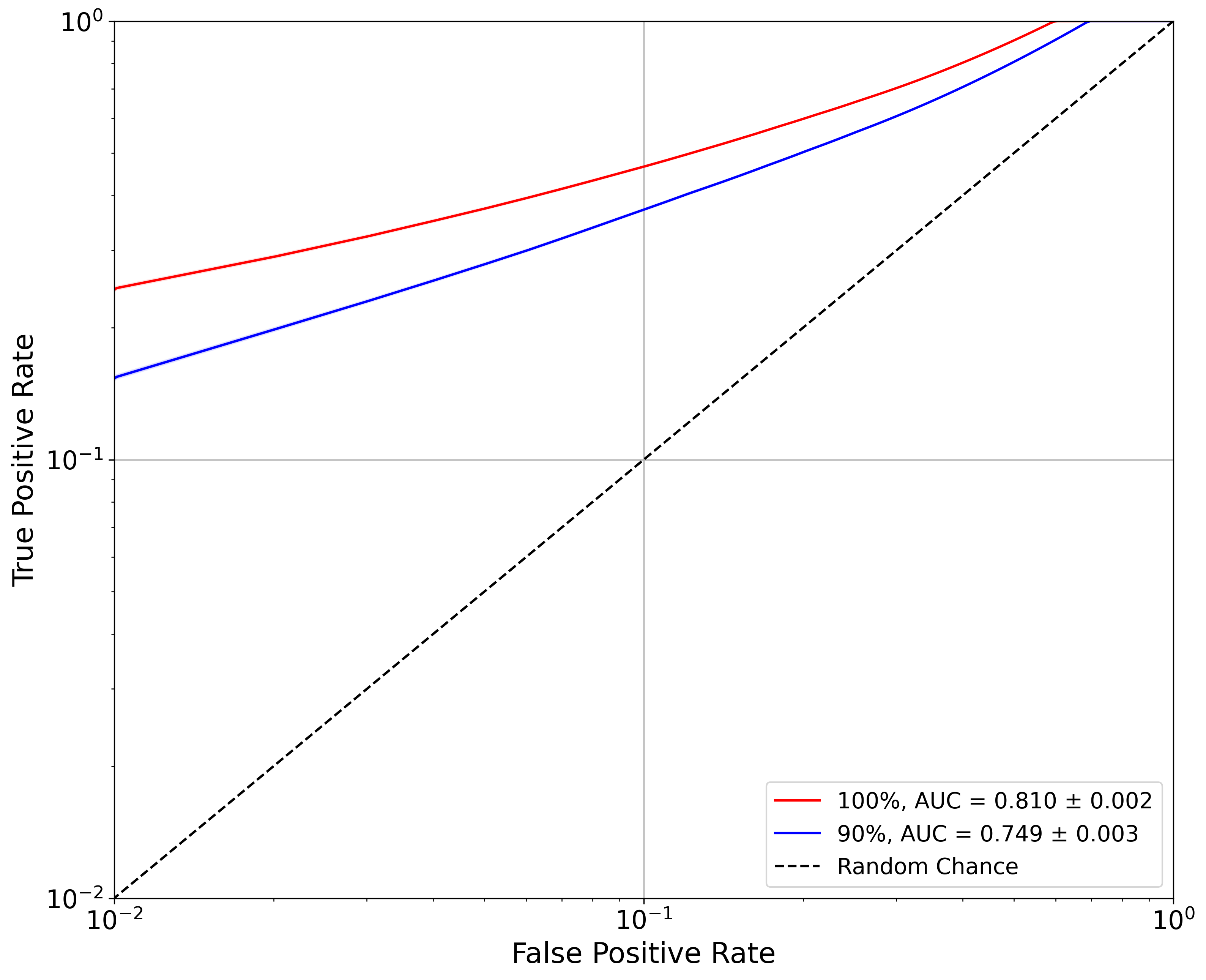}};
        \node[anchor=north west, inner sep=0, outer sep=0] (image2) at (4.5, 7.2) {\includegraphics[width=0.27\linewidth]{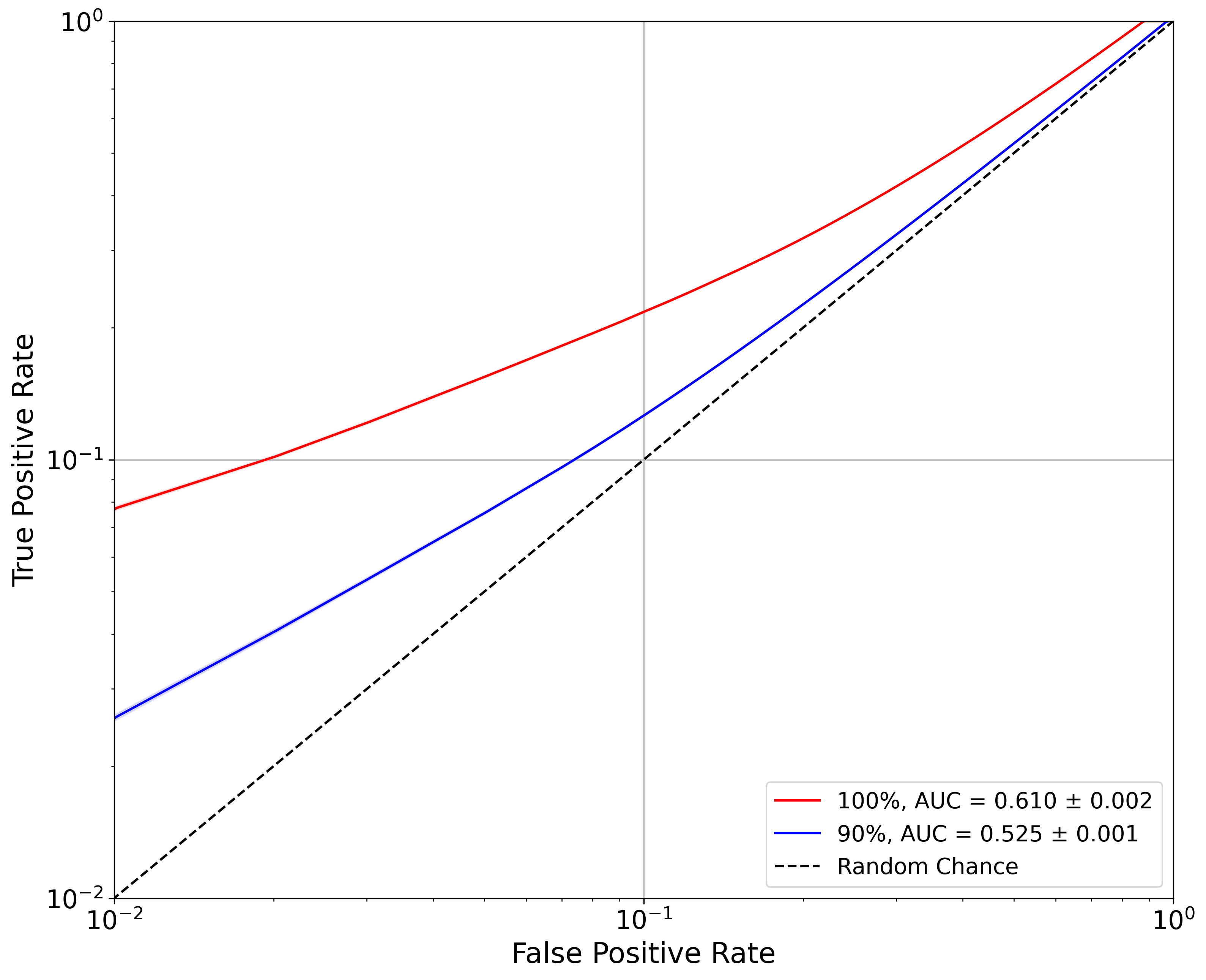}};
        \node[anchor=north west, inner sep=0, outer sep=0] (image3) at (9.0, 7.2) {\includegraphics[width=0.27\linewidth]{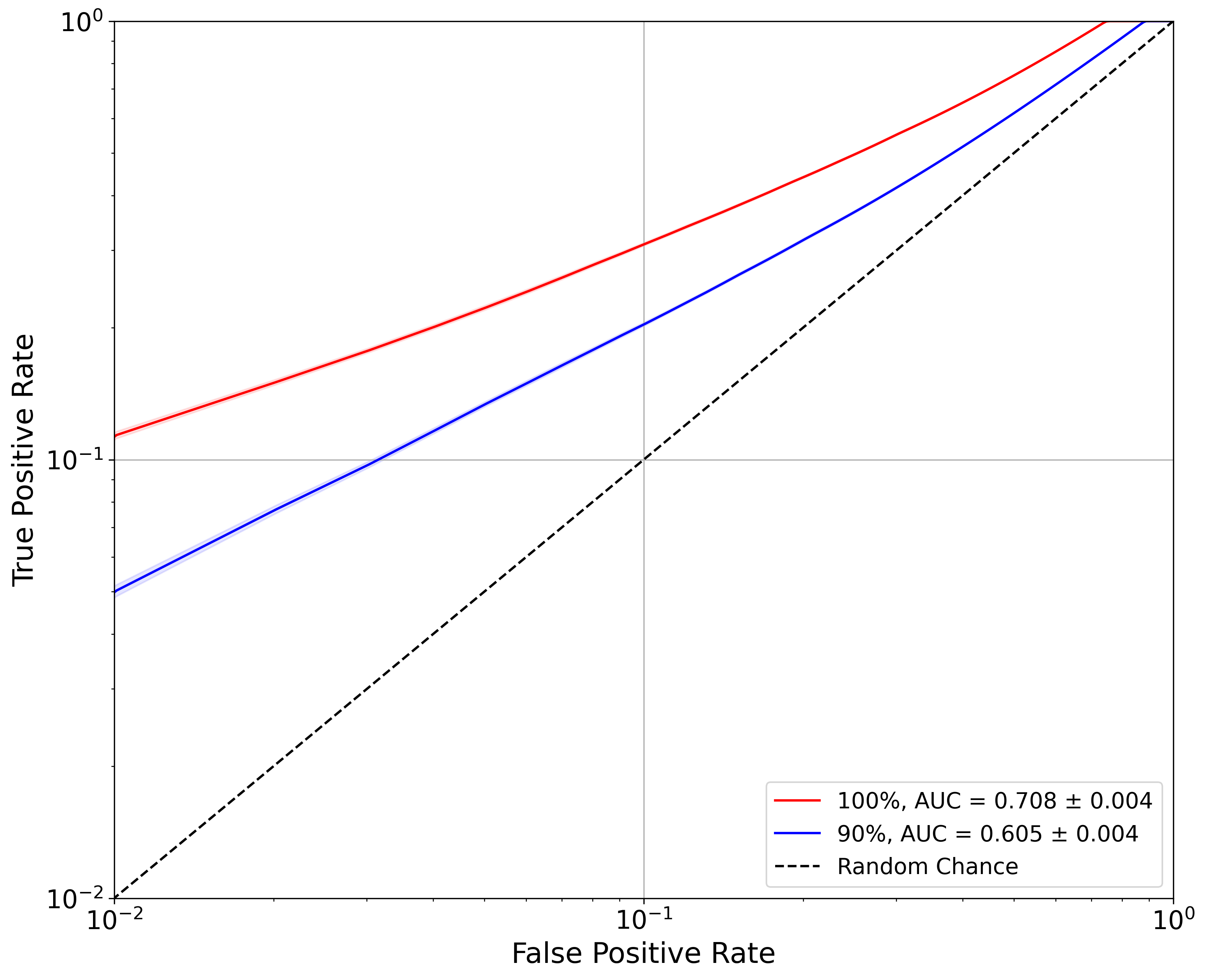}};
        \node[anchor=north west, inner sep=0, outer sep=0] (image4) at (13.4, 7.2) {\includegraphics[width=0.27\linewidth]{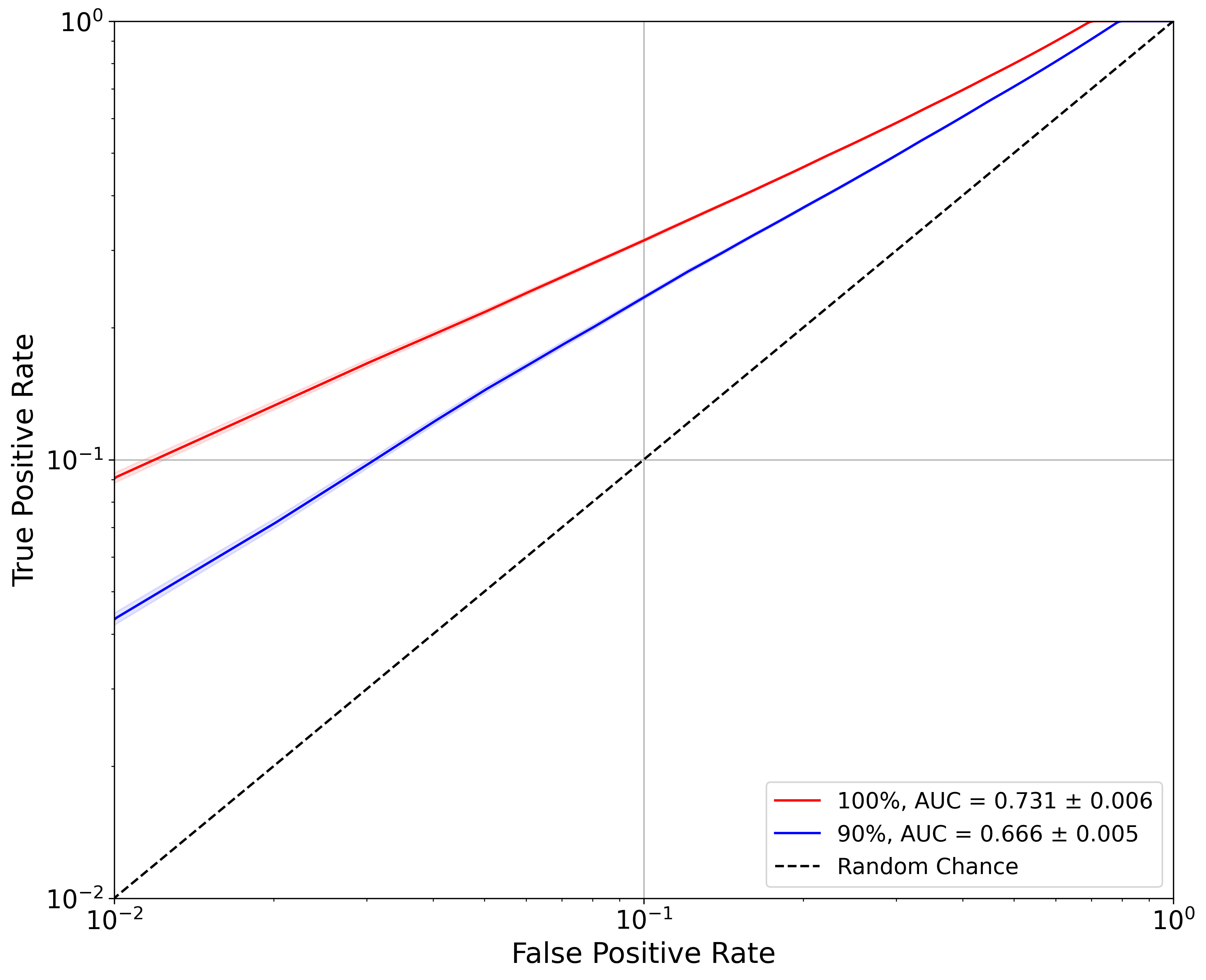}};
        
    \end{tikzpicture}

    \label{fig:auc-post-removal}

\end{figure*}

\begin{table*}
\caption{AUC of LiRA before and after removing 10\% of the dataset, using various attribution methods.}
\centering
\begin{tabular}{c c c c c c c c}
\hline
Dataset & K & AUC 100\% & AUC accuracy (90\%) & AUC self-waka (90\%) & AUC self-shapley (90\%) & AUC test-waka (90\%) & AUC test-shapley (90\%) \\
\hline
  adult & 1 & 0.731 & 0.610 & 0.605 & 0.604 & 0.696 & 0.704 \\
  adult & 5 & 0.604 & 0.566 & 0.555 & 0.555 & 0.594 & 0.598 \\
   bank & 1 & 0.810 & 0.527 & 0.525 & 0.525 & 0.597 & 0.608 \\
   bank & 5 & 0.691 & 0.524 & 0.512 & 0.510 & 0.552 & 0.558 \\
  CIFAR10 & 1 & 0.708 & 0.752 & 0.749 & 0.747 & 0.819 & 0.826 \\
  CIFAR10 & 5 & 0.596 & 0.626 & 0.631 & 0.628 & 0.696 & 0.699 \\
   IMDB & 1 & 0.610 & 0.671 & 0.666 & 0.666 & 0.740 & 0.754 \\
   IMDB & 5 & 0.555 & 0.592 & 0.580 & 0.582 & 0.611 & 0.617 \\
\hline
\label{tab:stats_before_after_removal1}
\end{tabular}
\end{table*}

\begin{table*}
\caption{TPR  at 5\% FPR of LiRA before and after removing 10\% of the dataset, using various attribution methods.}
\centering
\begin{tabular}{c c c c c c c c}
\hline
Dataset & K & TPR 100\% & TPR accuracy (90\%) & TPR self-waka (90\%) & TPR self-shapley (90\%) & TPR test-waka (90\%) & TPR test-shapley (90\%) \\
\hline
  adult & 1 & 0.217 & 0.137 & 0.133 & 0.134 & 0.213 & 0.216 \\
  adult & 5 & 0.080 & 0.076 & 0.069 & 0.072 & 0.074 & 0.082 \\
   bank & 1 & 0.373 & 0.077 & 0.075 & 0.075 & 0.144 & 0.152 \\
   bank & 5 & 0.189 & 0.070 & 0.062 & 0.060 & 0.079 & 0.078 \\
  CIFAR10 & 1 & 0.221 & 0.284 & 0.279 & 0.276 & 0.383 & 0.390 \\
  CIFAR10 & 5 & 0.081 & 0.123 & 0.125 & 0.123 & 0.197 & 0.199 \\
   IMDB & 1 & 0.154 & 0.150 & 0.144 & 0.142 & 0.219 & 0.228 \\
   IMDB & 5 & 0.080 & 0.066 & 0.060 & 0.059 & 0.090 & 0.096 \\
\hline
\label{tab:stats_before_after_removal2}
\end{tabular}
\end{table*}

\begin{figure*}
    \centering
    \caption{Experiment of data minimization on the synthetic data Moons generated using the scikit-learn library for a $k$-NN model with $k=5$. 
    The darker the points, the higher the corresponding Shapley values. 
    The results show that higher test-Shapley values tend to concentrate between the decision boundary and the external boundaries of the data, while higher self-Shapley values consistently identify points lying directly on the decision boundary.}
    \label{fig:synthetic-exp}
    
    \begin{tikzpicture}[baseline=(current bounding box.north)]
        
        \node at (1.8, 6.5) {100\%};
        \node at (5.4, 6.5) {80\%};
        \node at (9, 6.5) {60\%};
        \node at (12.6, 6.5) {40\%};
        \node at (16.2, 6.5) {20\%};
        
        \node[rotate=90] at (-0.5, 4) {test-Shapley};
        \node[rotate=90] at (-0.5, 1) {self-Shapley};
        
        
        \node[anchor=north west, inner sep=0, outer sep=0] at (0, 6) {\includegraphics[width=3.2cm, keepaspectratio=true]{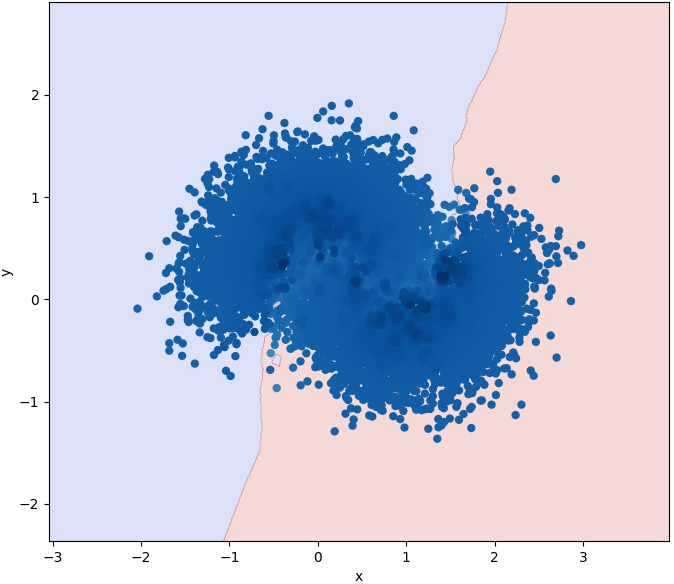}}; 
        \node[anchor=north west, inner sep=0, outer sep=0] at (3.6, 6) {\includegraphics[width=3.2cm, keepaspectratio=true]{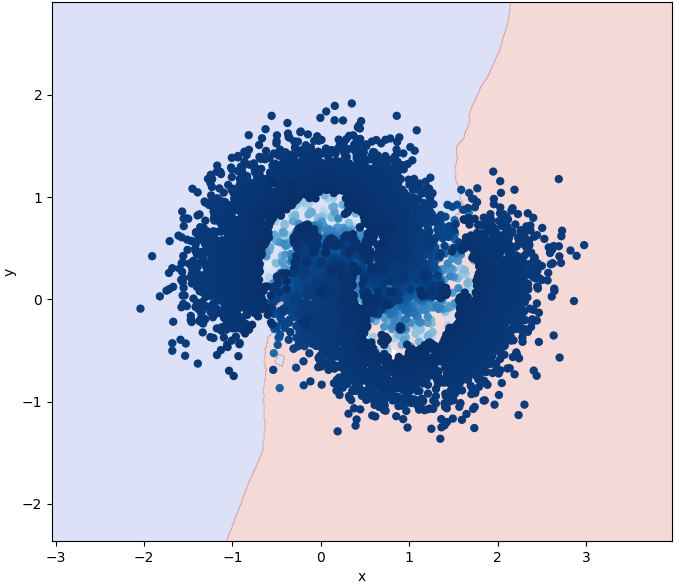}}; 
        \node[anchor=north west, inner sep=0, outer sep=0] at (7.2, 6) {\includegraphics[width=3.2cm, keepaspectratio=true]{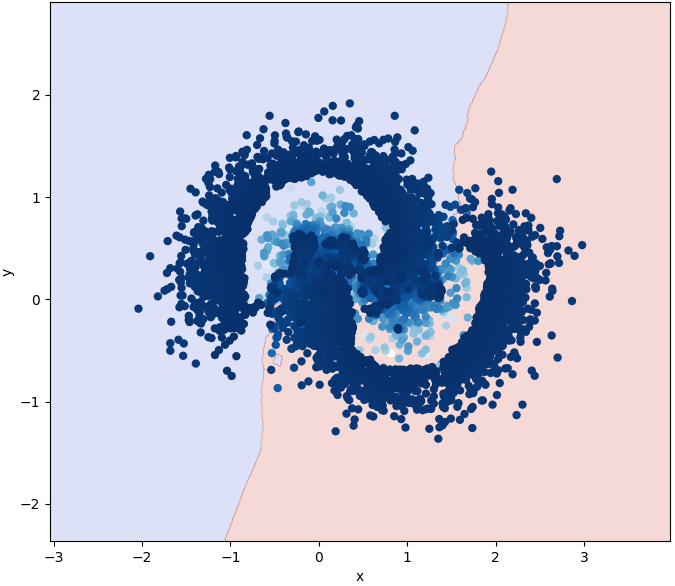}}; 
        \node[anchor=north west, inner sep=0, outer sep=0] at (10.8, 6) {\includegraphics[width=3.2cm, keepaspectratio=true]{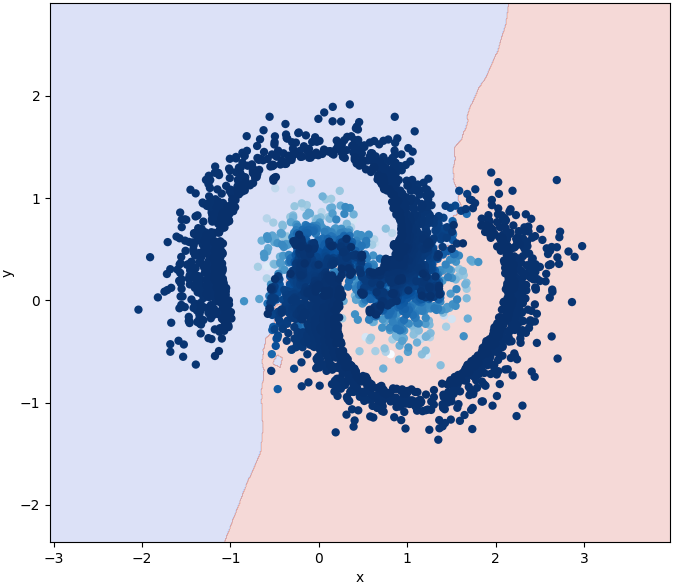}}; 
        \node[anchor=north west, inner sep=0, outer sep=0] at (14.4, 6) {\includegraphics[width=3.2cm, keepaspectratio=true]{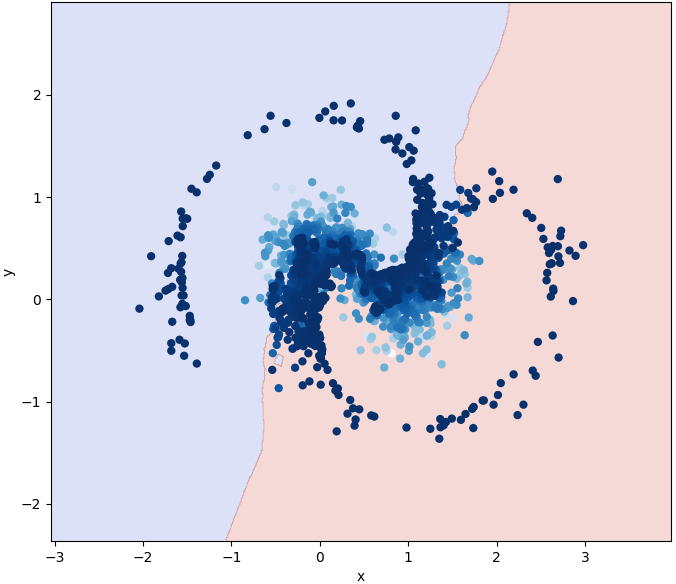}}; 
        
        \node[anchor=north west, inner sep=0, outer sep=0] at (0, 2.5) {\includegraphics[width=3.2cm, keepaspectratio=true]{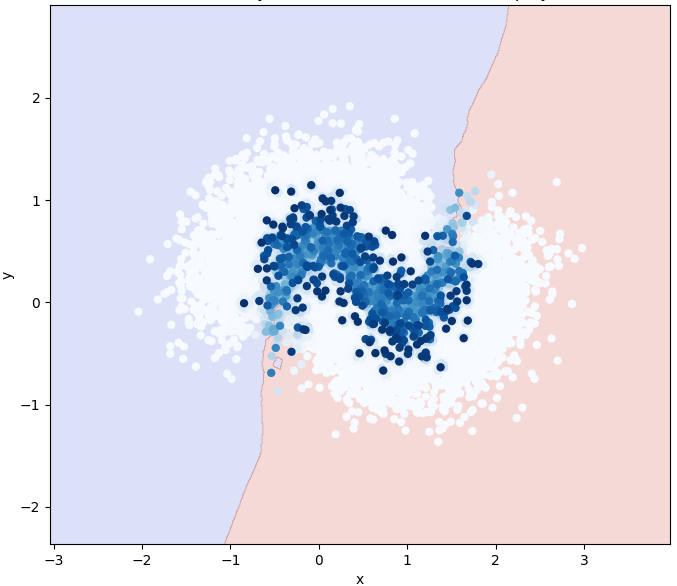}}; 
        \node[anchor=north west, inner sep=0, outer sep=0] at (3.6, 2.5) {\includegraphics[width=3.2cm, keepaspectratio=true]{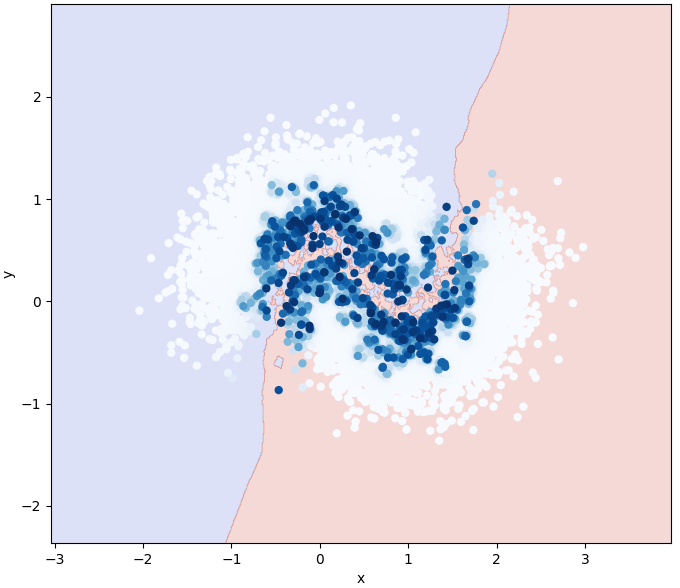}}; 
        \node[anchor=north west, inner sep=0, outer sep=0] at (7.2, 2.5) {\includegraphics[width=3.2cm, keepaspectratio=true]{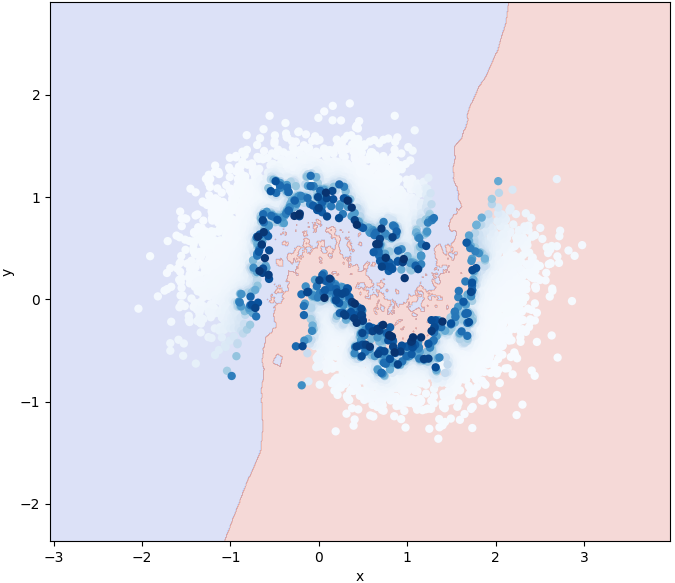}}; 
        \node[anchor=north west, inner sep=0, outer sep=0] at (10.8, 2.5) {\includegraphics[width=3.2cm, keepaspectratio=true]{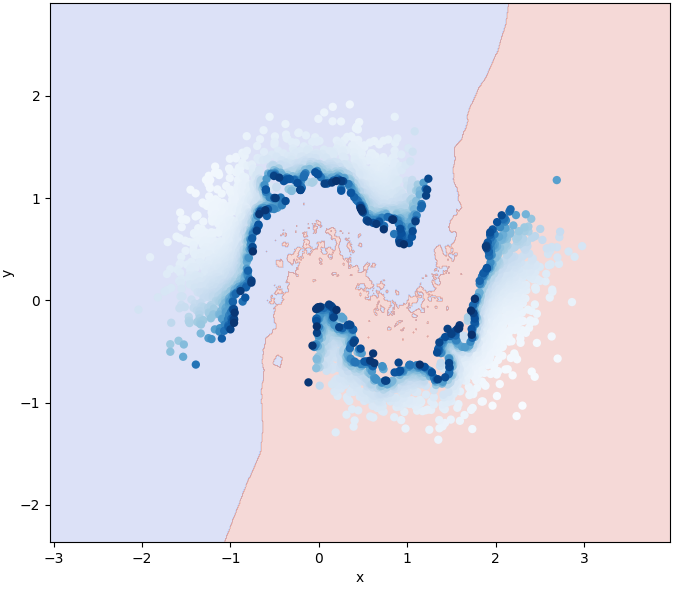}}; 
        \node[anchor=north west, inner sep=0, outer sep=0] at (14.4, 2.5) {\includegraphics[width=3.2cm, keepaspectratio=true]{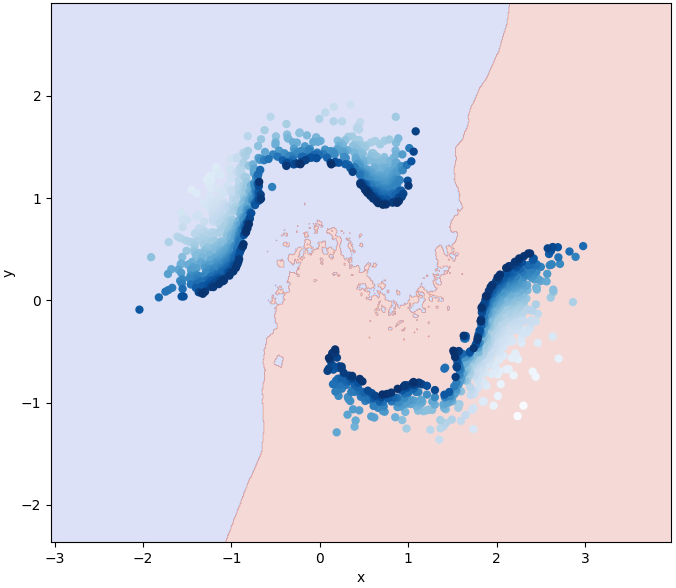}}; 
        
    \end{tikzpicture}
    
\end{figure*}

\begin{table*}
\centering
\caption{Spearman correlation results by dataset and parameter $k$.}
\begin{tabular}{lccccc}
\hline
Dataset & $K=1$ & $K=2$ & $K=3$ & $K=4$ & $K=5$ \\
\hline
IMDB   & 0.95 ($\pm$0.02) & 0.69 ($\pm$0.08) & 0.62 ($\pm$0.08) & 0.64 ($\pm$0.08) & 0.65 ($\pm$0.08) \\
CIFAR  & 0.97 ($\pm$0.01) & 0.84 ($\pm$0.06) & 0.70 ($\pm$0.12) & 0.66 ($\pm$0.10) & 0.65 ($\pm$0.12) \\
Adult  & 0.94 ($\pm$0.03) & 0.77 ($\pm$0.08) & 0.64 ($\pm$0.11) & 0.59 ($\pm$0.12) & 0.57 ($\pm$0.11) \\
Bank   & 0.75 ($\pm$0.07) & 0.64 ($\pm$0.14) & 0.56 ($\pm$0.18) & 0.49 ($\pm$0.15) & 0.45 ($\pm$0.15) \\
\hline
\end{tabular}
\label{tab:spearman_results}
\end{table*}

\begin{table}
\caption{Average ASR with Standard Error for $k=1$, $k=5$ and Logistic Regression (LogReg)}
\begin{center}
\begin{tabular}{|c|c|c|c|}
\hline
\textbf{Dataset} & \textbf{K=1} & \textbf{K=5} & \textbf{LogReg} \\
\hline
Adult  & 0.625 ± 0.002 & 0.561 ± 0.001 & 0.512 ± 0.001 \\
\hline
Bank   & 0.56  ± 0.001 & 0.531 ± 0.001 & 0.503 ± 0.0004 \\
\hline
CIFAR  & 0.702 ± 0.001 & 0.621 ± 0.001 & 0.541 ± 0.001 \\
\hline
IMDB   & 0.65  ± 0.002 & 0.573 ± 0.001 & 0.529 ± 0.001 \\
\hline
\end{tabular}
\label{tab:asr_stats_k1_k5}
\end{center}
\end{table}

\begin{figure*}[htbp]
    \centering
    \begin{tikzpicture}[node distance=0cm]
        \node at (2.2, 3.5) {LiRA};
        \node at (6.7, 3.5) {t-WaKA};
        \node at (11.2, 3.5) {Conf};
        \node at (15.5, 3.5) {Conf-calib};
        
        \node[anchor=north west] (img1) at (0,3) {\includegraphics[width=0.26\linewidth]{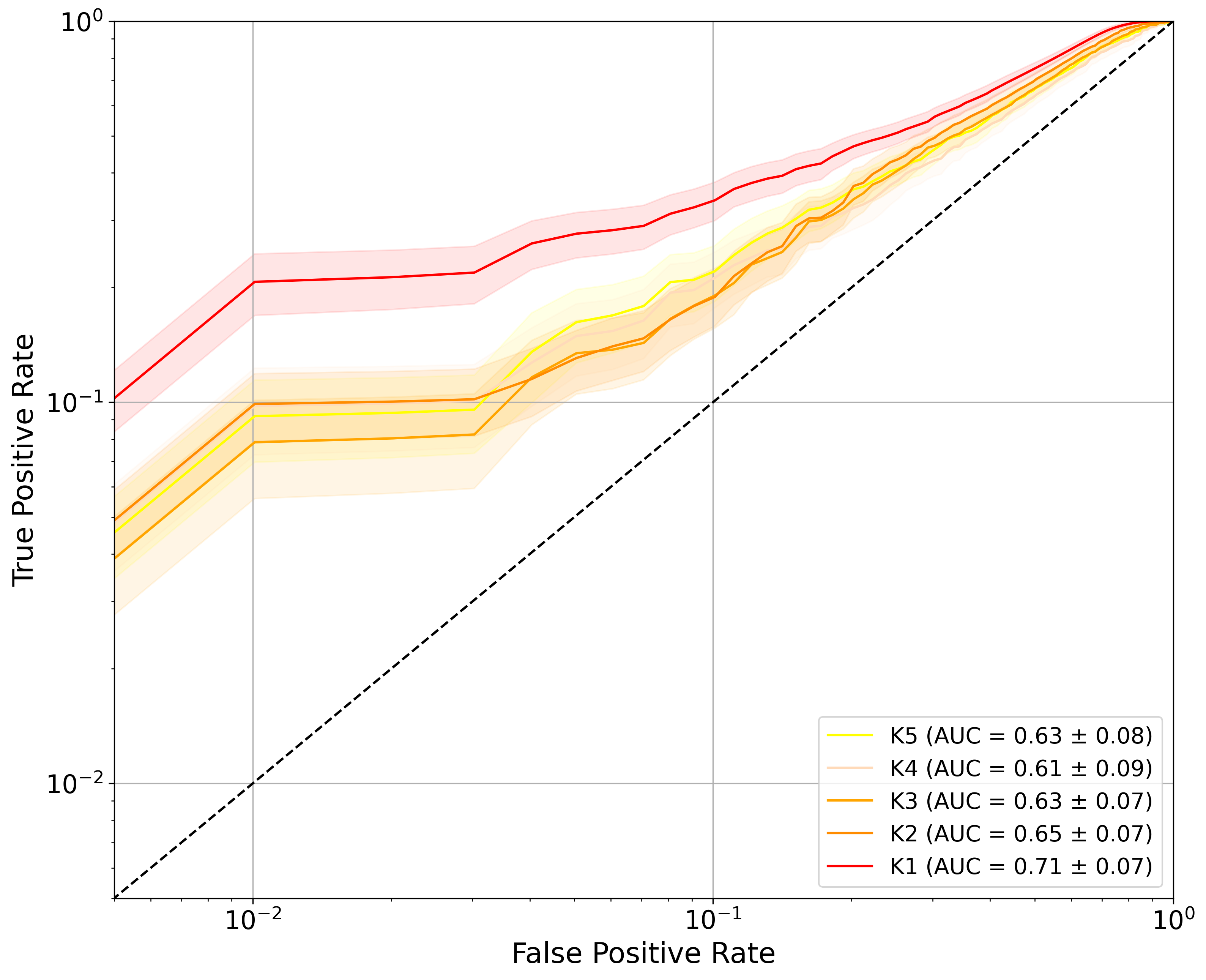}};
        \node[anchor=north west] (img2) at (4.4,3) {\includegraphics[width=0.26\linewidth]{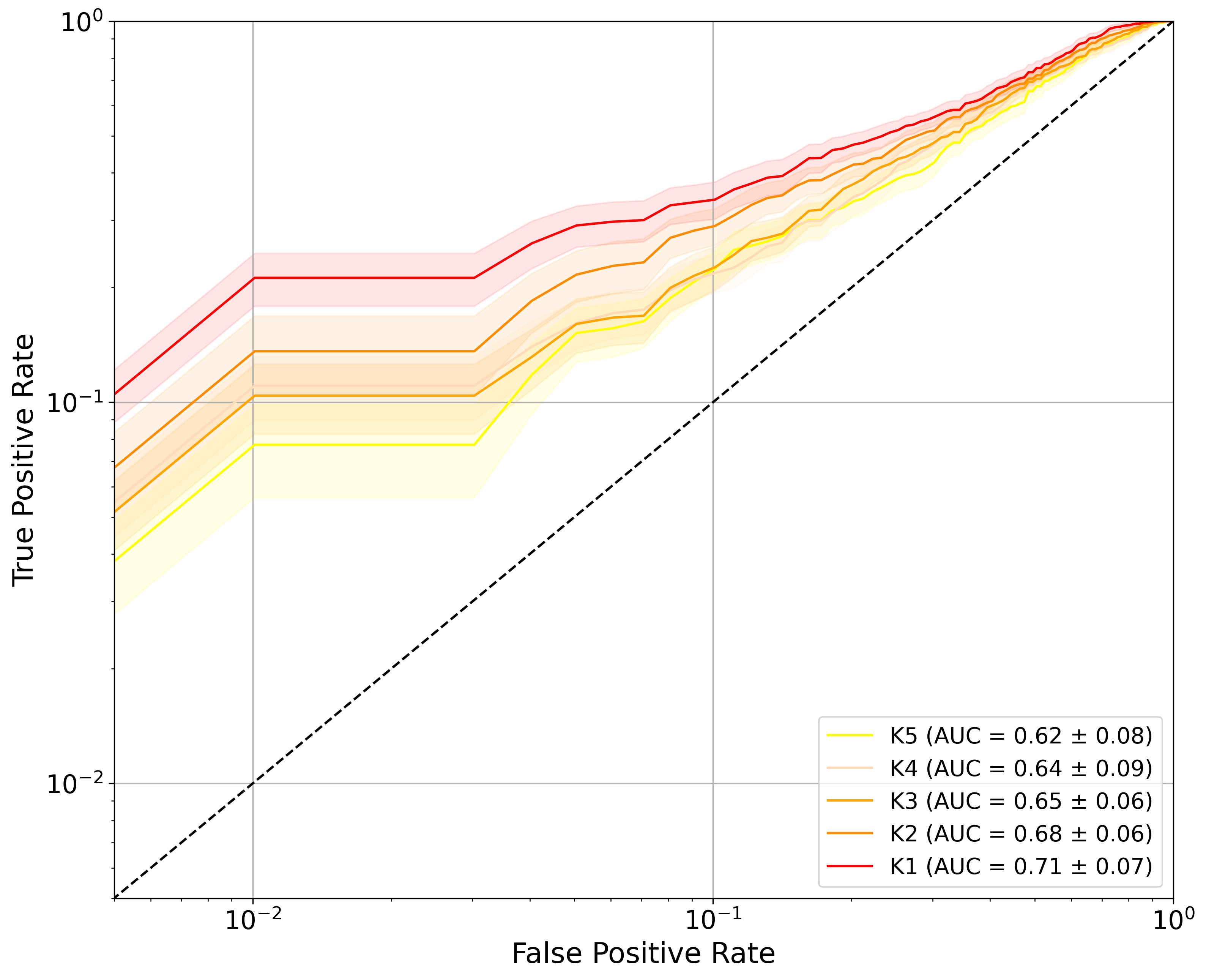}};
        \node[anchor=north west] (img3) at (8.8,3) {\includegraphics[width=0.26\linewidth]{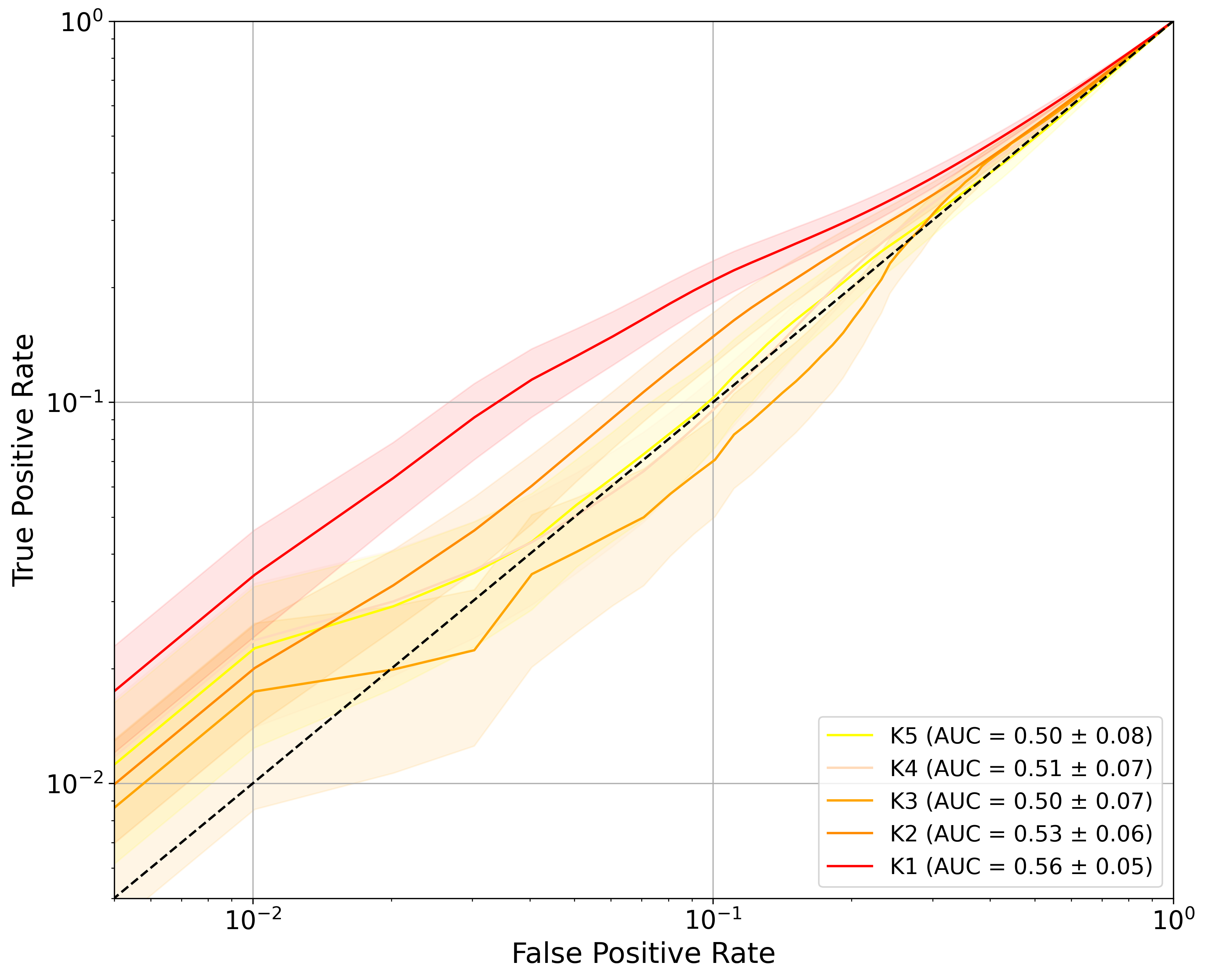}};
        \node[anchor=north west] (img4) at (13.2,3) {\includegraphics[width=0.26\linewidth]{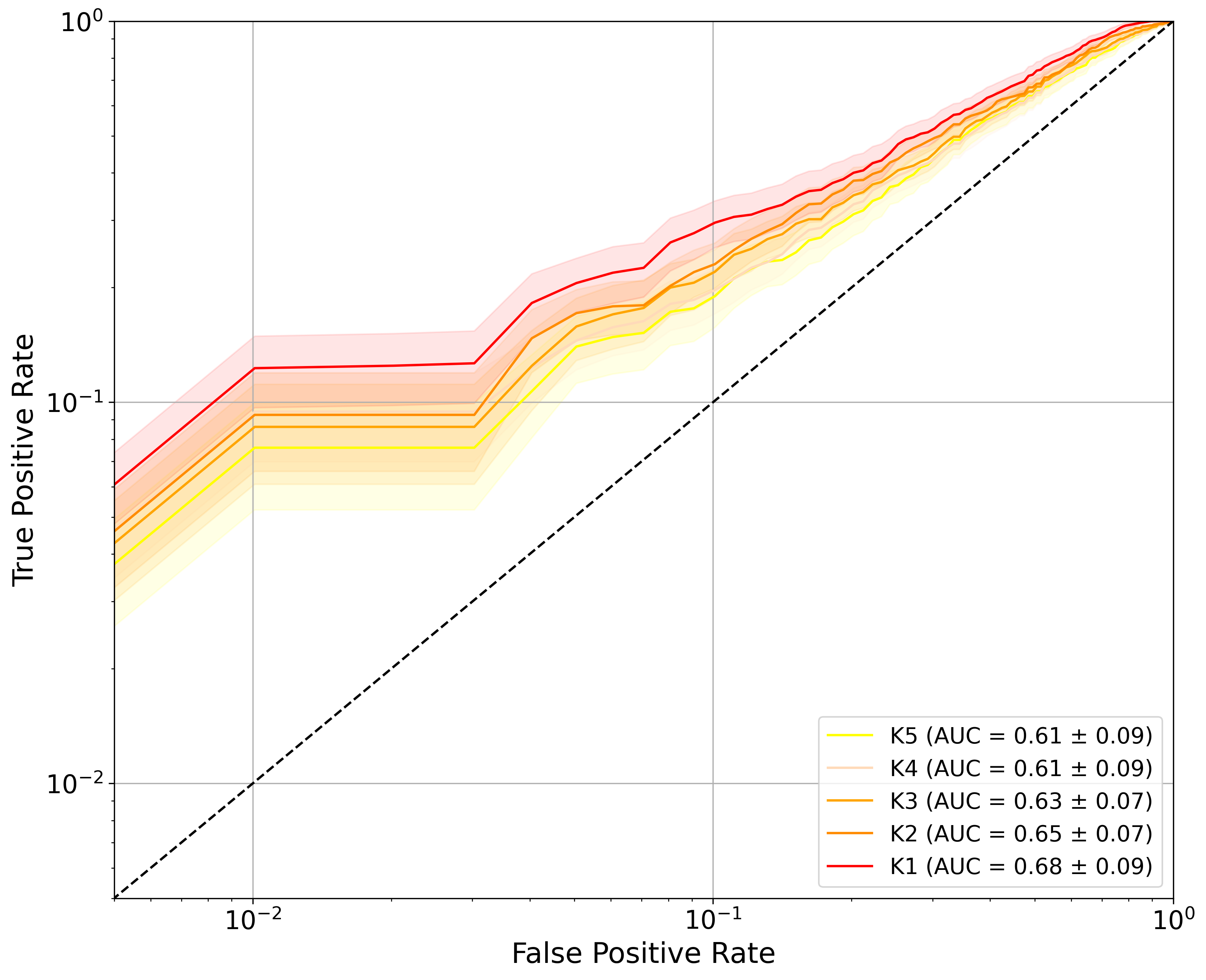}};
    \end{tikzpicture}
    \caption{CIFAR10: Membership Inference Attacks (MIA) AUC and ROC curves (in log-scale) using LiRA, t-WaKA, confidence, and calibrated confidence attacks for $k$ values between 1 and 5.}
    \label{fig:roc_comparison_cifar}
\end{figure*}

\begin{figure*}[htbp]
    \centering
    \begin{tikzpicture}[node distance=0cm]
        \node at (2.2, 3.5) {LiRA};
        \node at (6.7, 3.5) {t-WaKA};
        \node at (11.2, 3.5) {Conf};
        \node at (15.5, 3.5) {Conf-calib};
        
        \node[anchor=north west] (img1) at (0,3) {\includegraphics[width=0.26\linewidth]{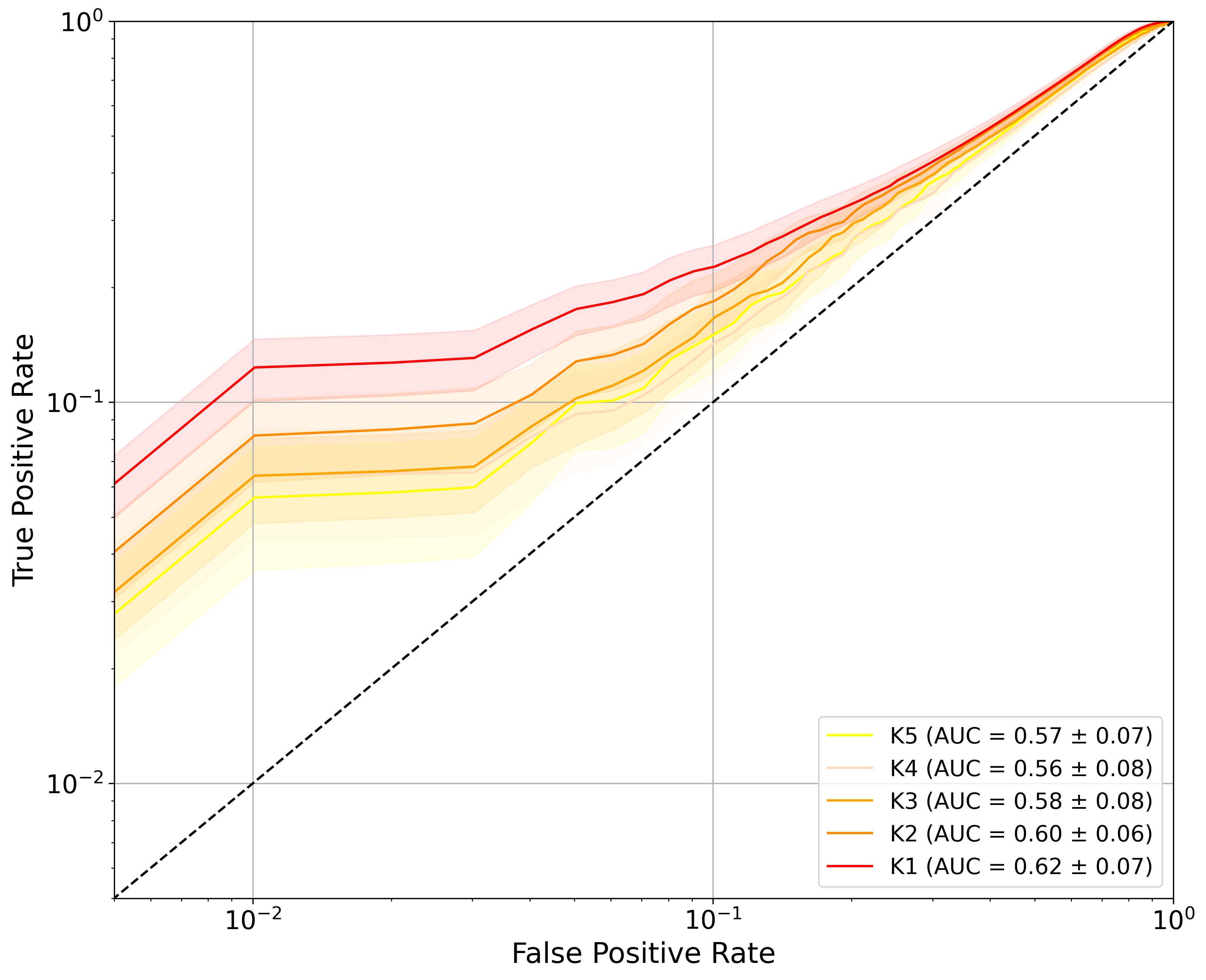}};
        \node[anchor=north west] (img2) at (4.4,3) {\includegraphics[width=0.26\linewidth]{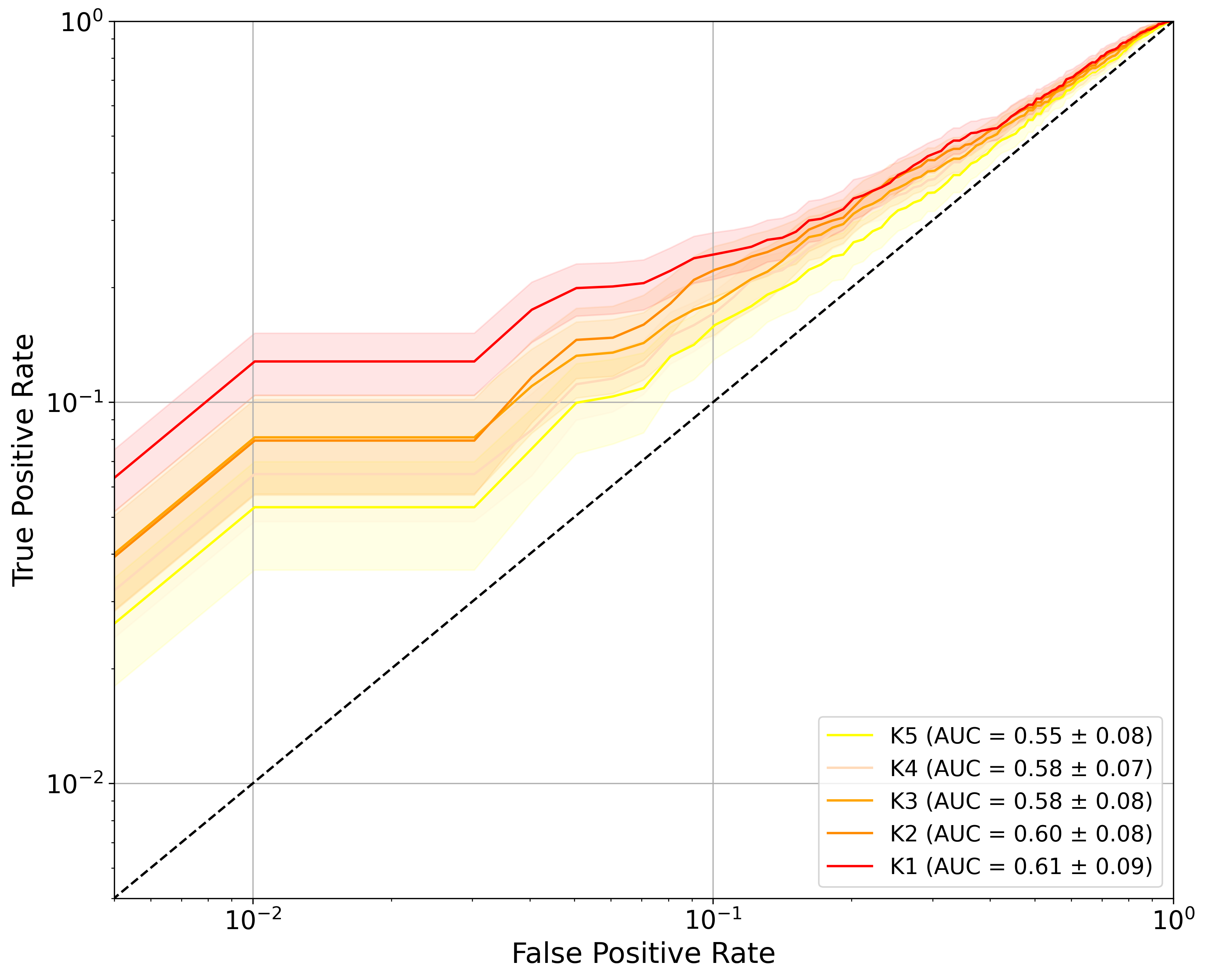}};
        \node[anchor=north west] (img3) at (8.8,3) {\includegraphics[width=0.26\linewidth]{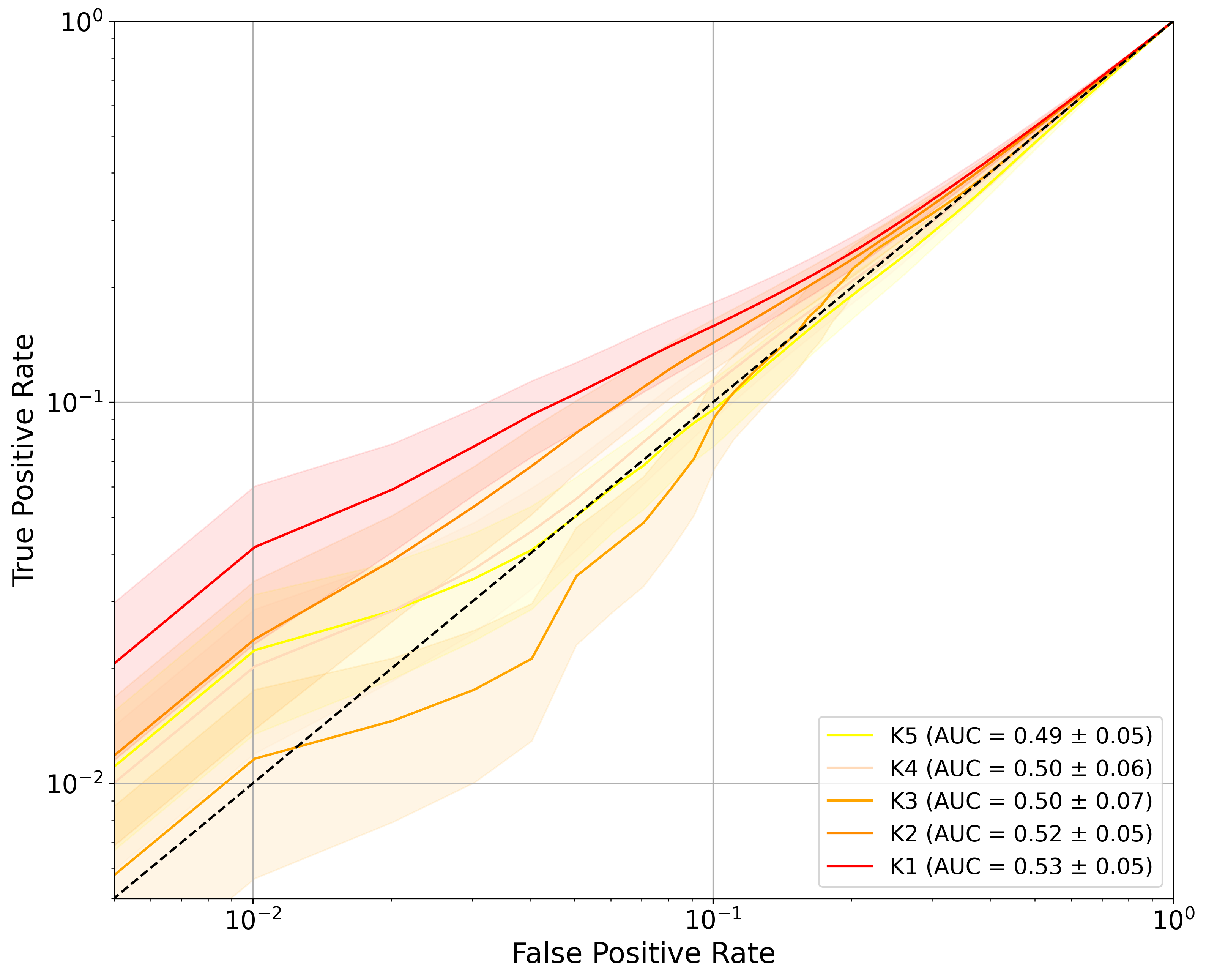}};
        \node[anchor=north west] (img4) at (13.2,3) {\includegraphics[width=0.26\linewidth]{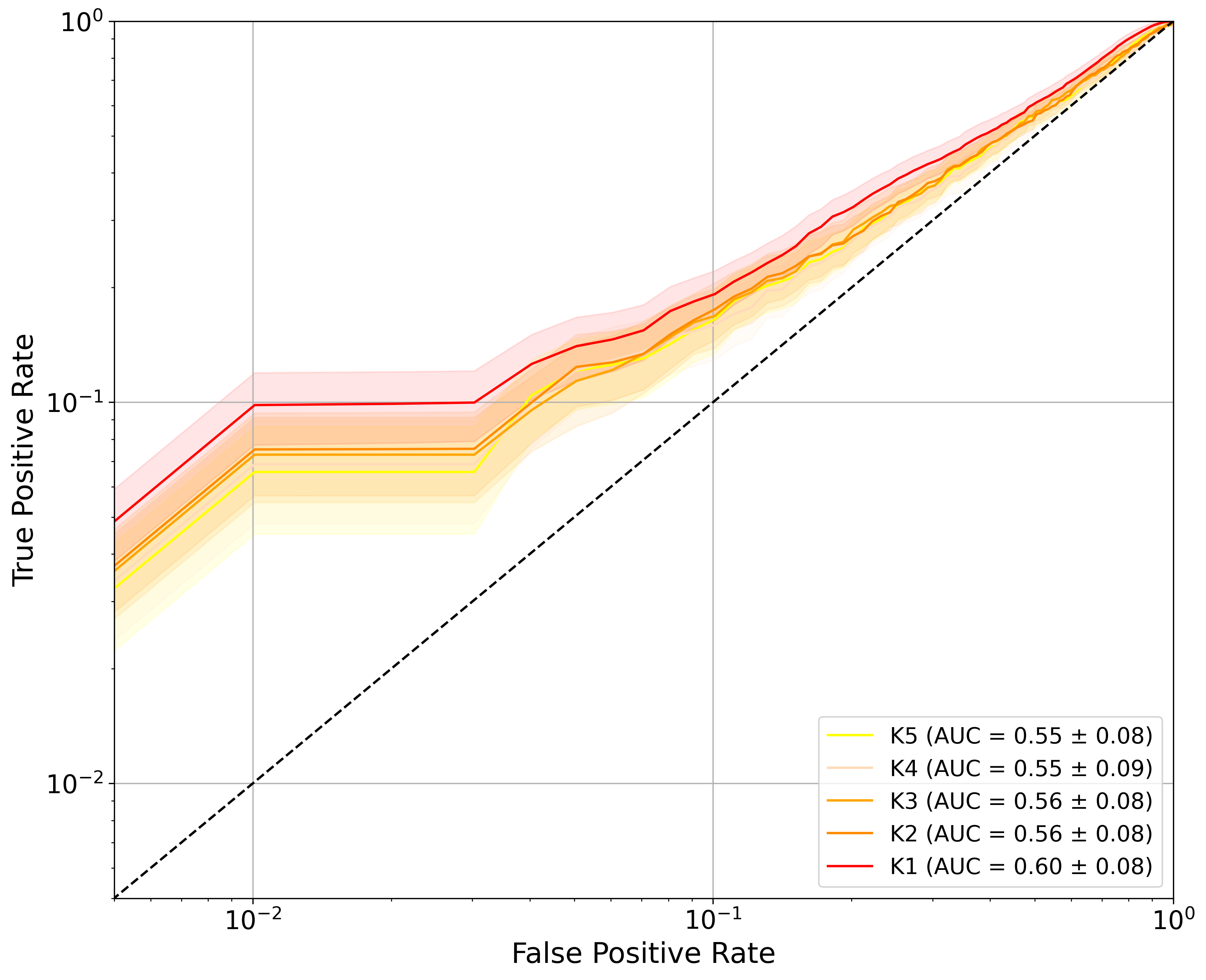}};
    \end{tikzpicture}
    \caption{Bank: Membership Inference Attacks (MIA) AUC and ROC curves (in log-scale) using LiRA, t-WaKA, confidence, and calibrated confidence attacks for $k$ values between 1 and 5.}
    \label{fig:roc_comparison_bank}
\end{figure*}

\begin{figure*}[htbp]
    \centering
    \begin{tikzpicture}[node distance=0cm]
        \node at (2.2, 3.5) {LiRA};
        \node at (6.7, 3.5) {t-WaKA};
        \node at (11.2, 3.5) {Conf};
        \node at (15.5, 3.5) {Conf-calib};
        
        \node[anchor=north west] (img1) at (0,3) {\includegraphics[width=0.26\linewidth]{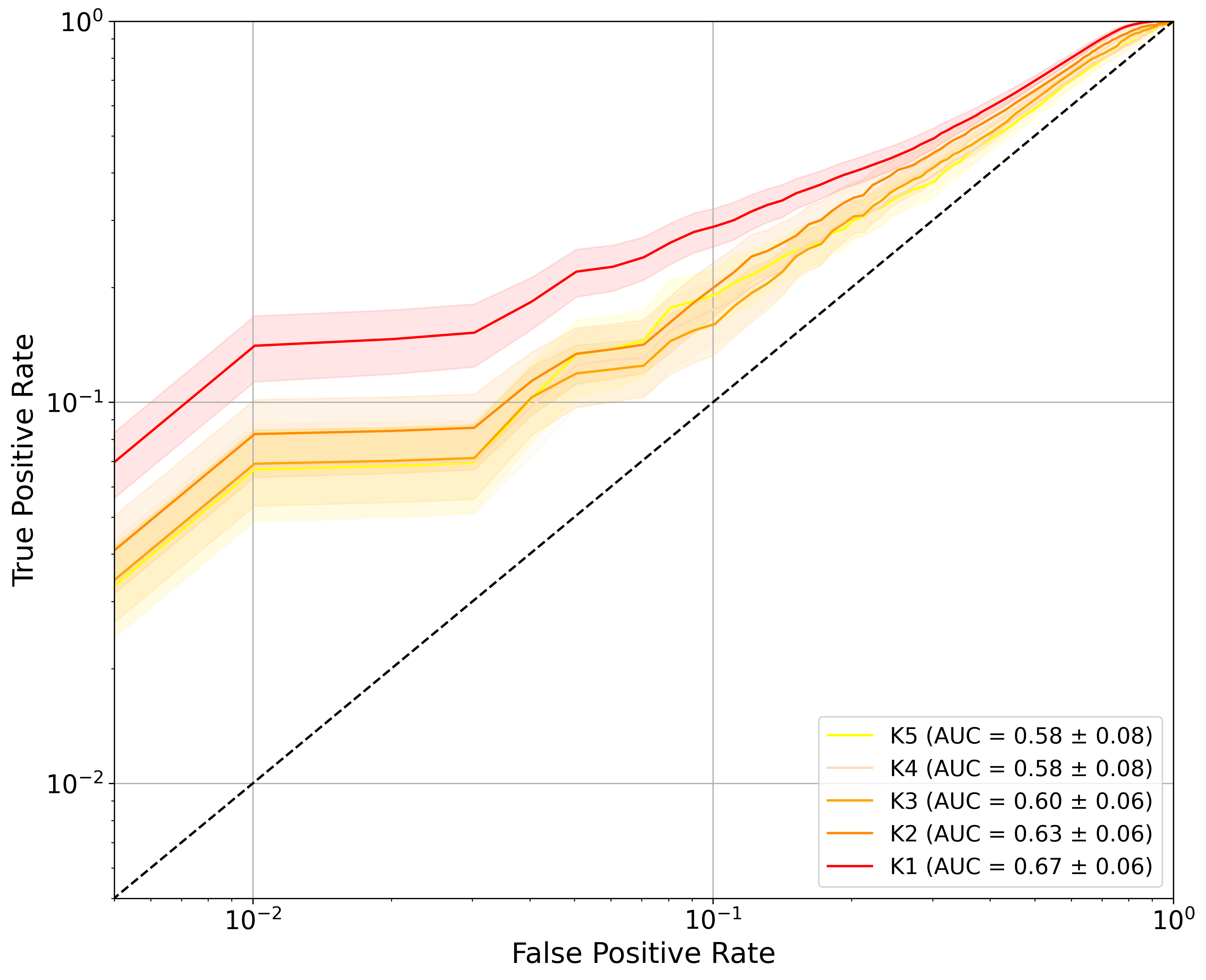}};
        \node[anchor=north west] (img2) at (4.4,3) {\includegraphics[width=0.26\linewidth]{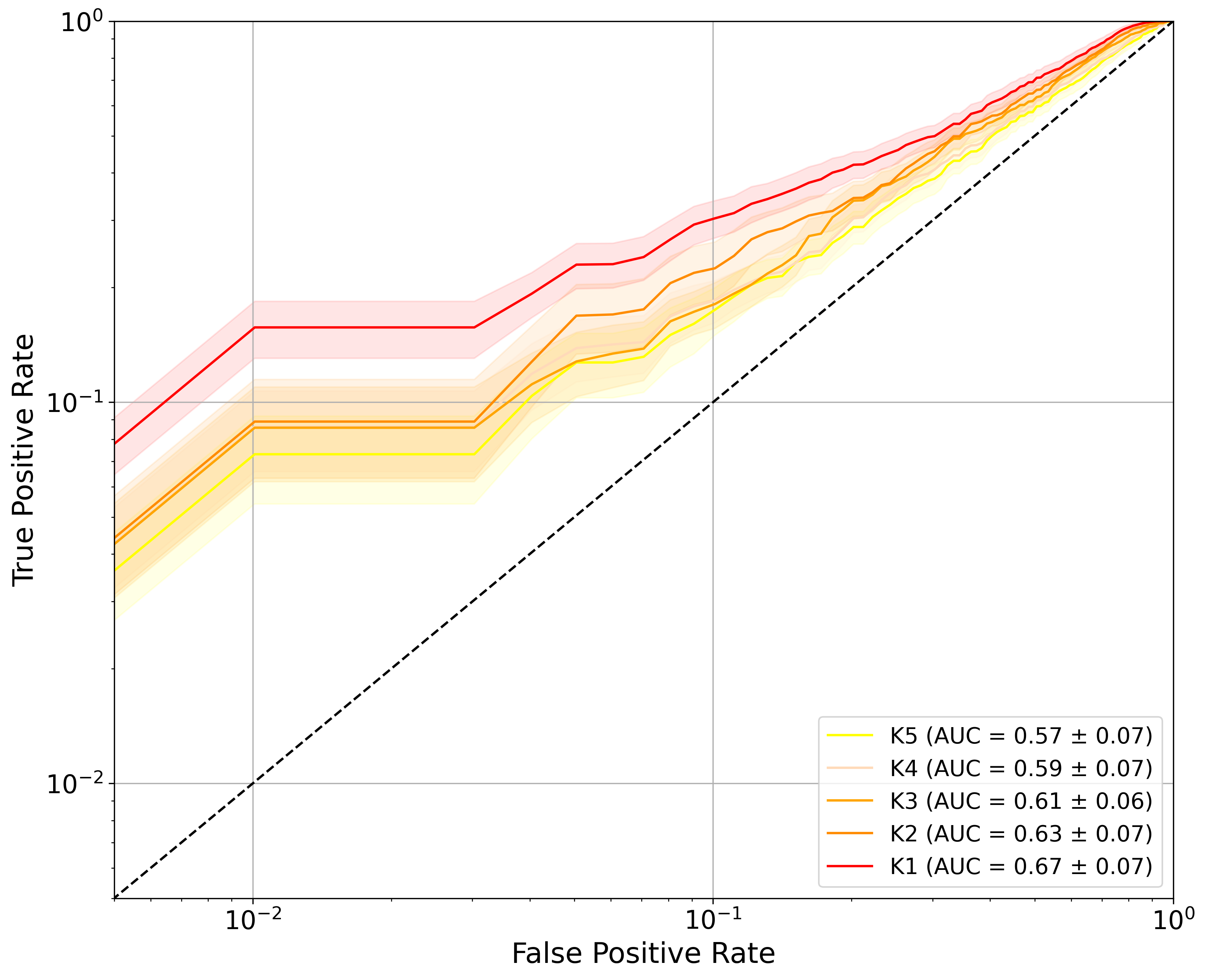}};
        \node[anchor=north west] (img3) at (8.8,3) {\includegraphics[width=0.26\linewidth]{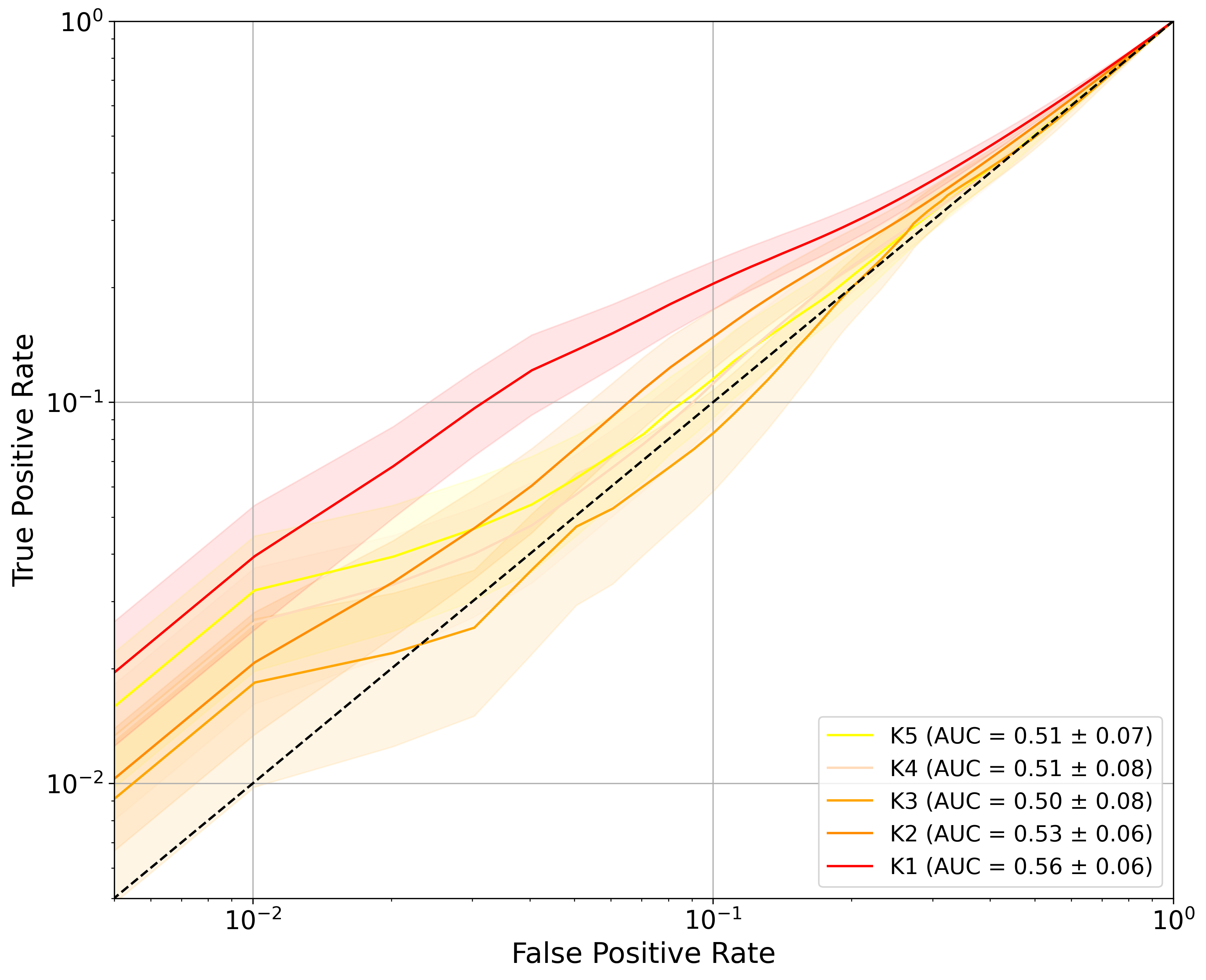}};
        \node[anchor=north west] (img4) at (13.2,3) {\includegraphics[width=0.26\linewidth]{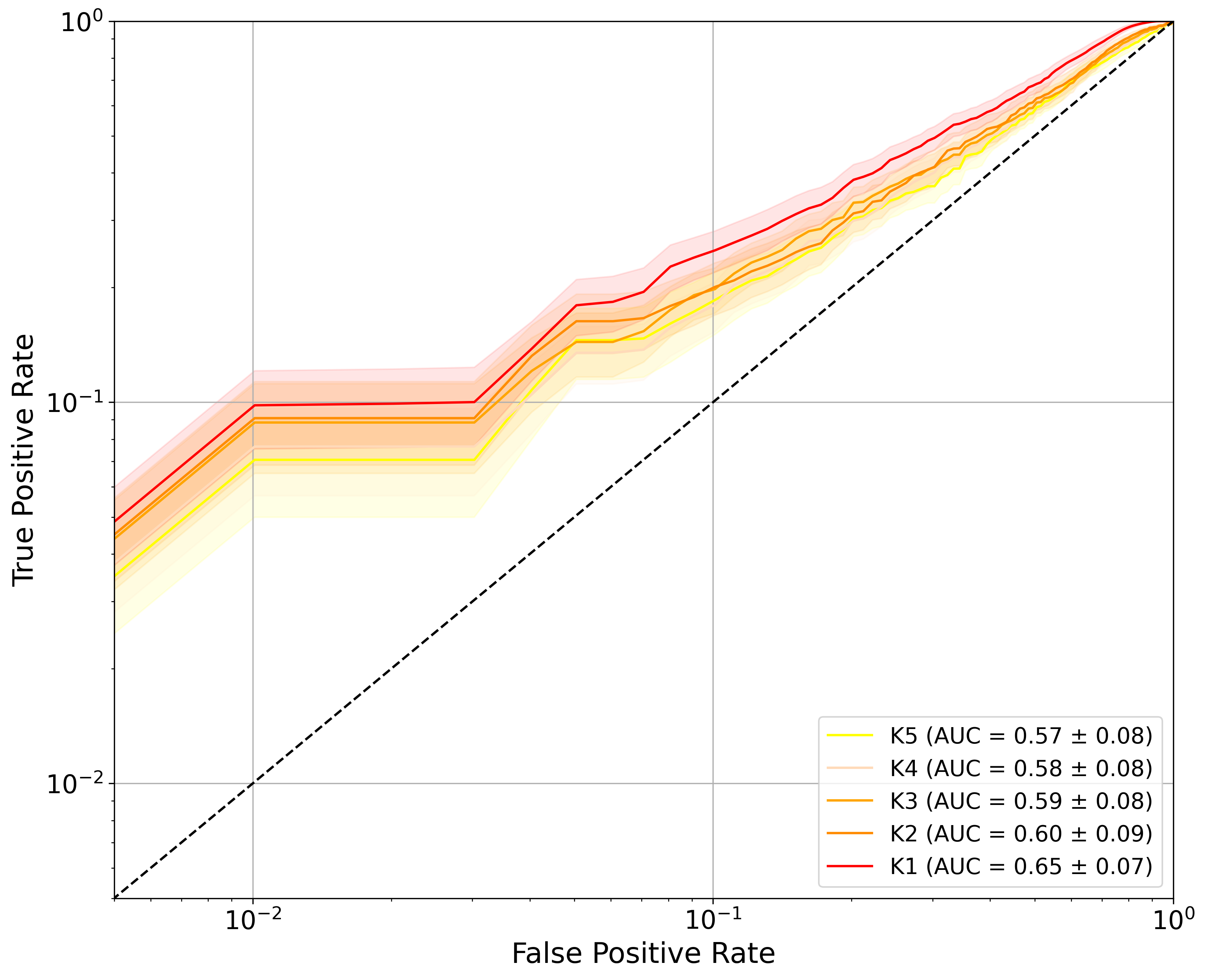}};
    \end{tikzpicture}
    \caption{Adult: Membership Inference Attacks (MIA) AUC and ROC curves (in log-scale) using LiRA, t-WaKA, confidence, and calibrated confidence attacks for $k$ values between 1 and 5.}
    \label{fig:roc_comparison_adult}
\end{figure*}

\begin{figure*}[htbp]
    \centering
    \begin{tikzpicture}[node distance=0cm]
        \node at (2.2, 3.5) {LiRA};
        \node at (6.7, 3.5) {t-WaKA};
        \node at (11.2, 3.5) {Conf};
        \node at (15.5, 3.5) {Conf-calib};
        
        \node[anchor=north west] (img1) at (0,3) {\includegraphics[width=0.26\linewidth]{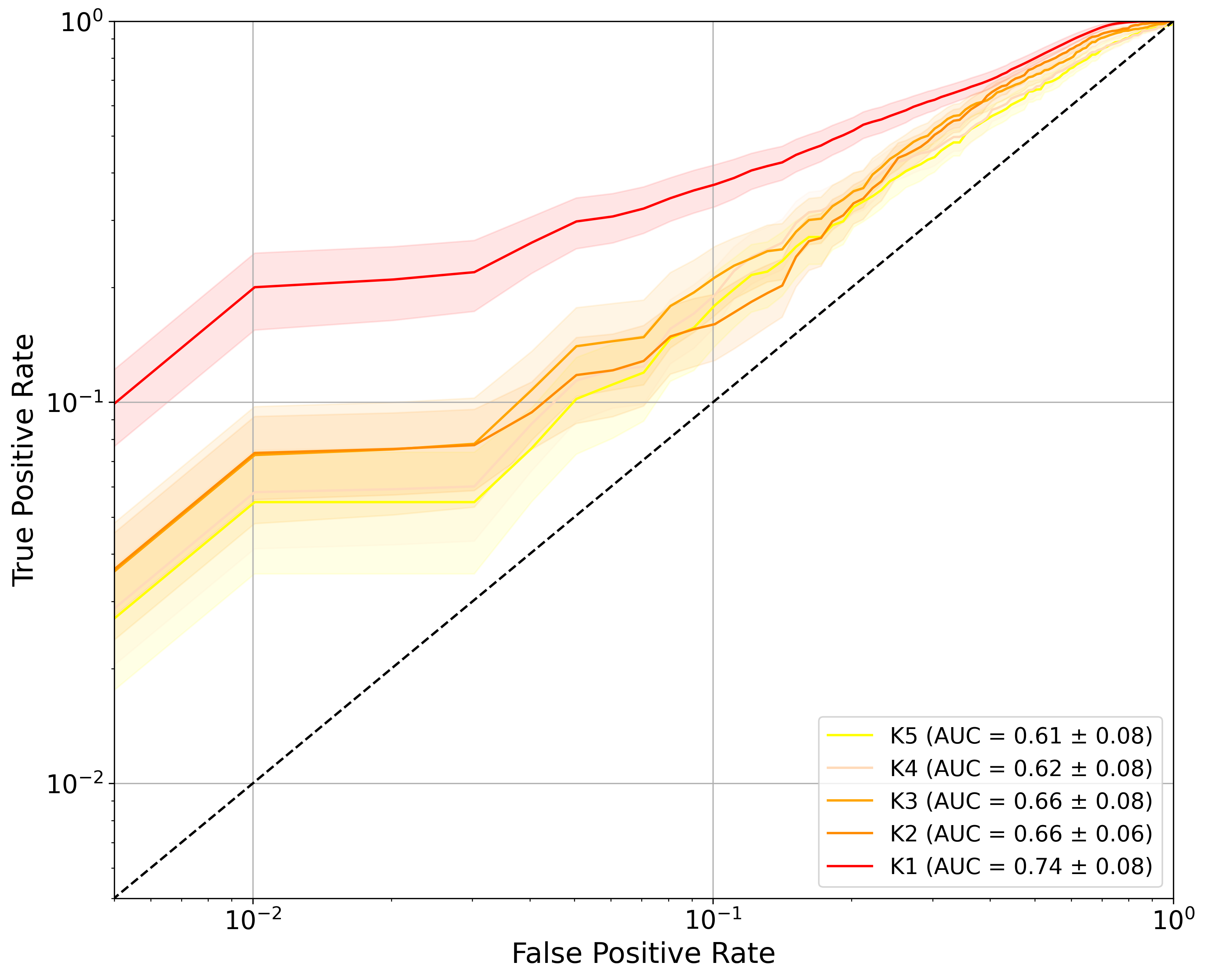}};
        \node[anchor=north west] (img2) at (4.4,3) {\includegraphics[width=0.26\linewidth]{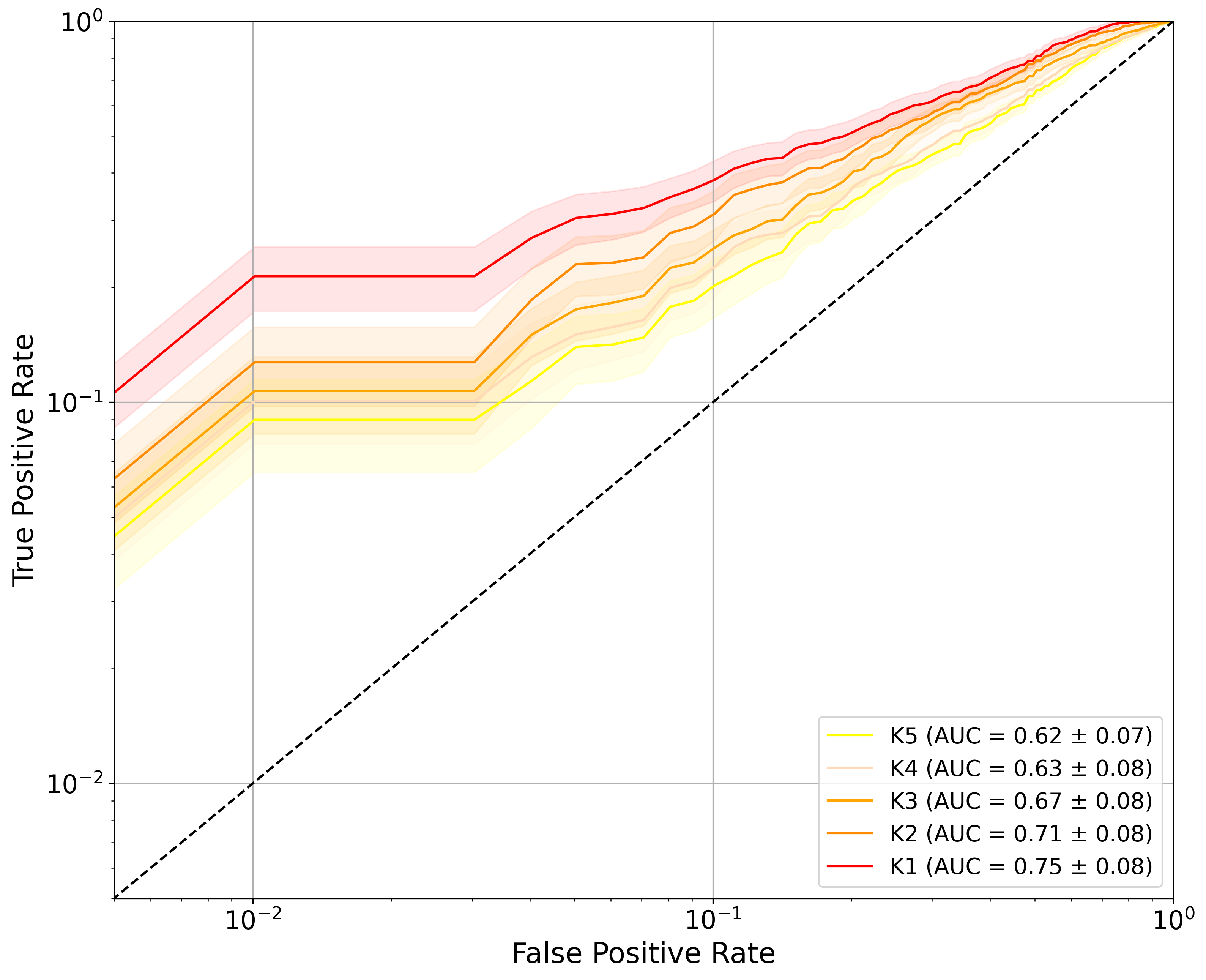}};
        \node[anchor=north west] (img3) at (8.8,3) {\includegraphics[width=0.26\linewidth]{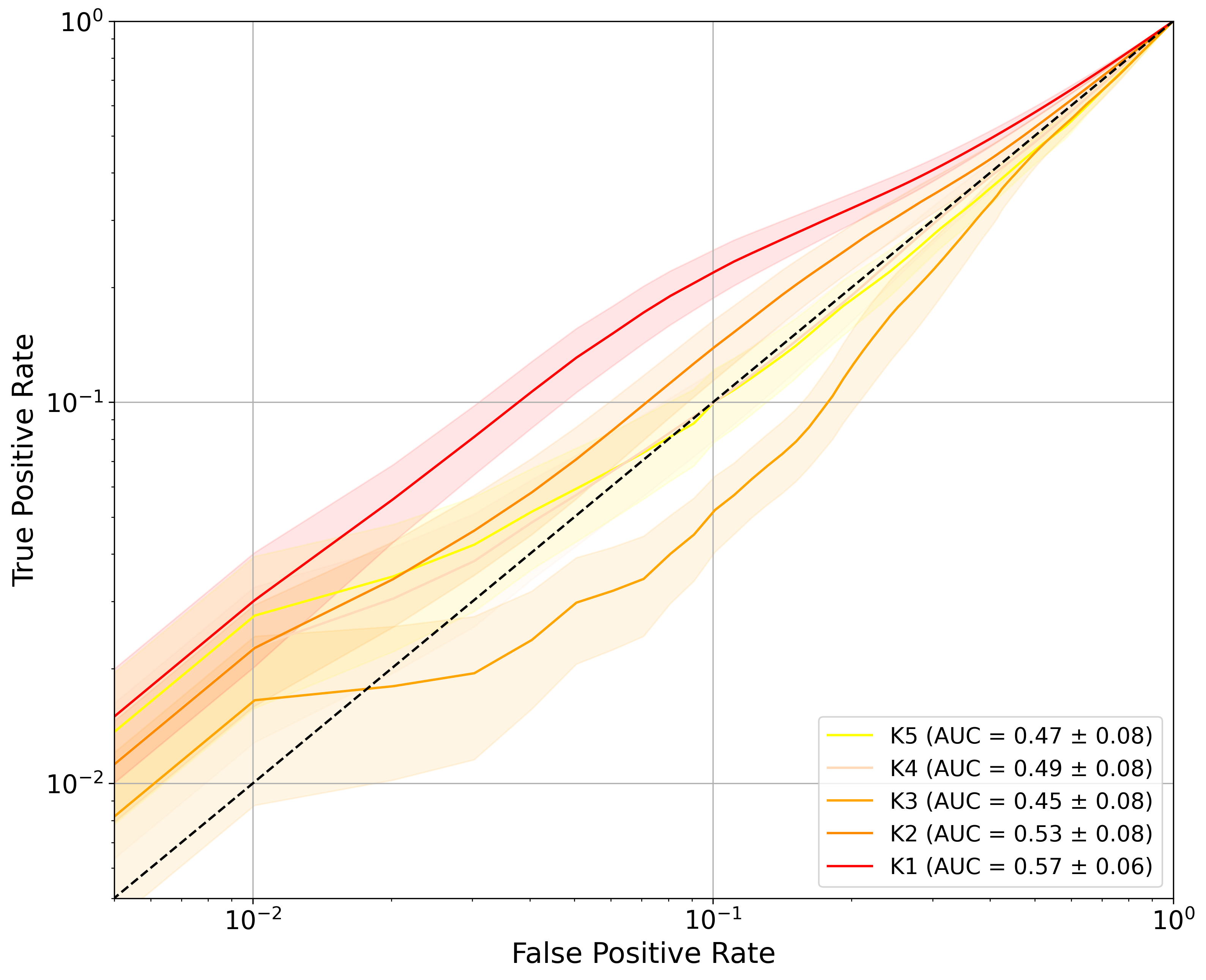}};
        \node[anchor=north west] (img4) at (13.2,3) {\includegraphics[width=0.26\linewidth]{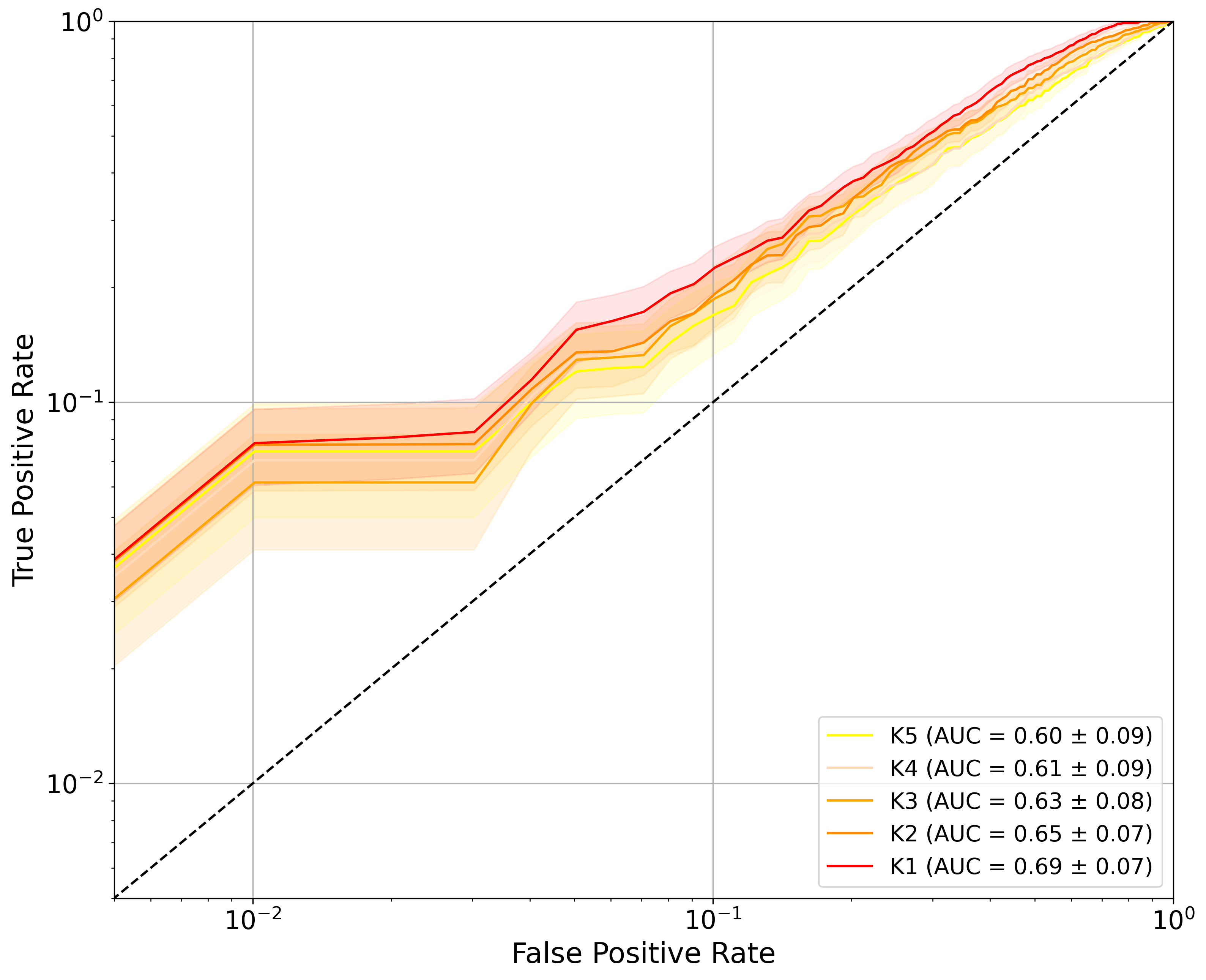}};
    \end{tikzpicture}
    \caption{IMDB: Membership Inference Attacks (MIA) AUC and ROC curves (in log-scale) using LiRA, t-WaKA, confidence, and calibrated confidence attacks for $k$ values between 1 and 5.}
    \label{fig:roc_comparison_imdb}
\end{figure*}

\begin{figure*}[htbp]
    \centering
    \begin{tikzpicture}[node distance=0cm]
        \node at (2.2, 3.5) {LiRA};
        \node at (6.7, 3.5) {t-WaKA};
        \node at (11.2, 3.5) {Conf};
        \node at (15.5, 3.5) {Conf-calib};
        
        \node[anchor=north west] (img1) at (0,3) {\includegraphics[width=0.26\linewidth]{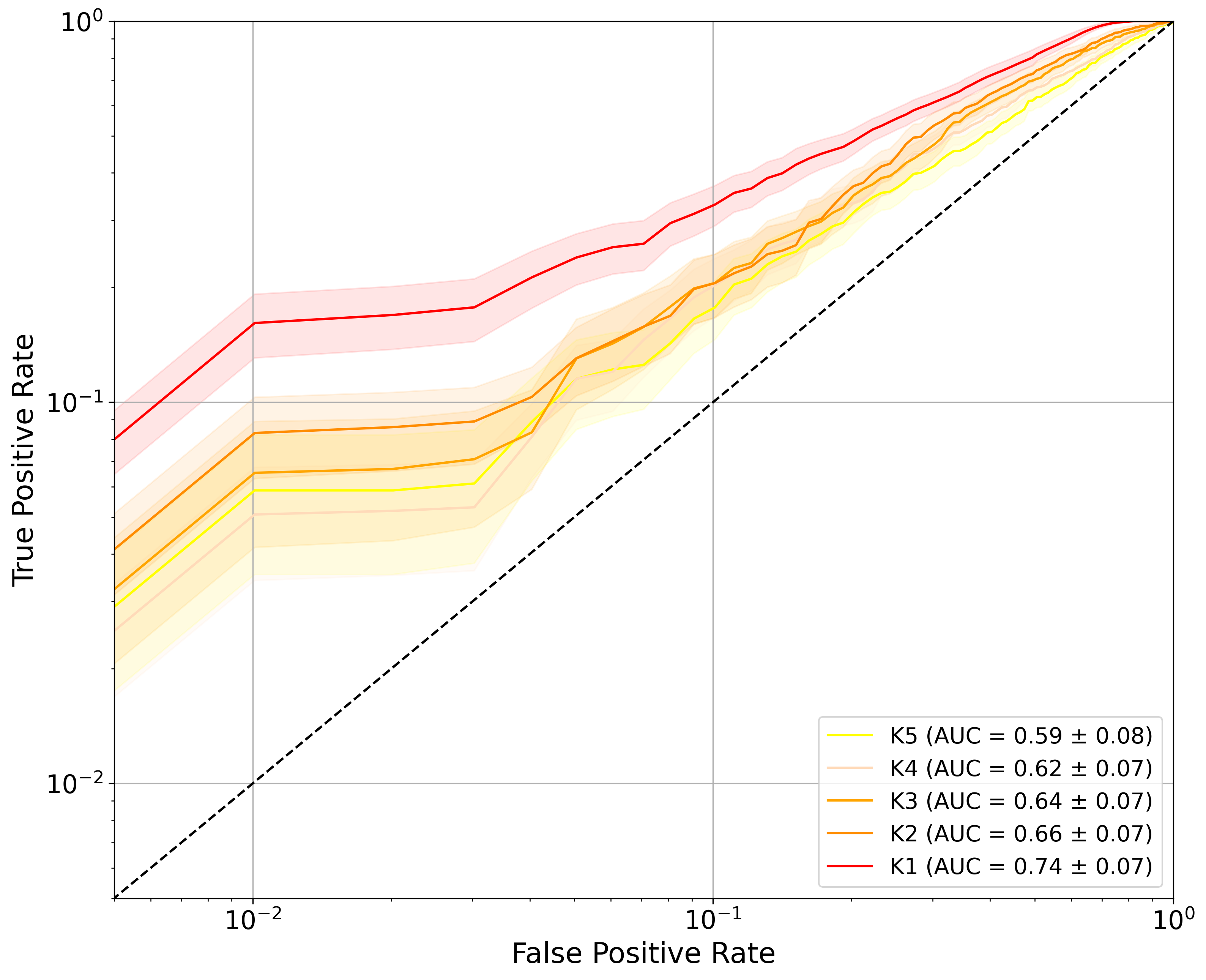}};
        \node[anchor=north west] (img2) at (4.4,3) {\includegraphics[width=0.26\linewidth]{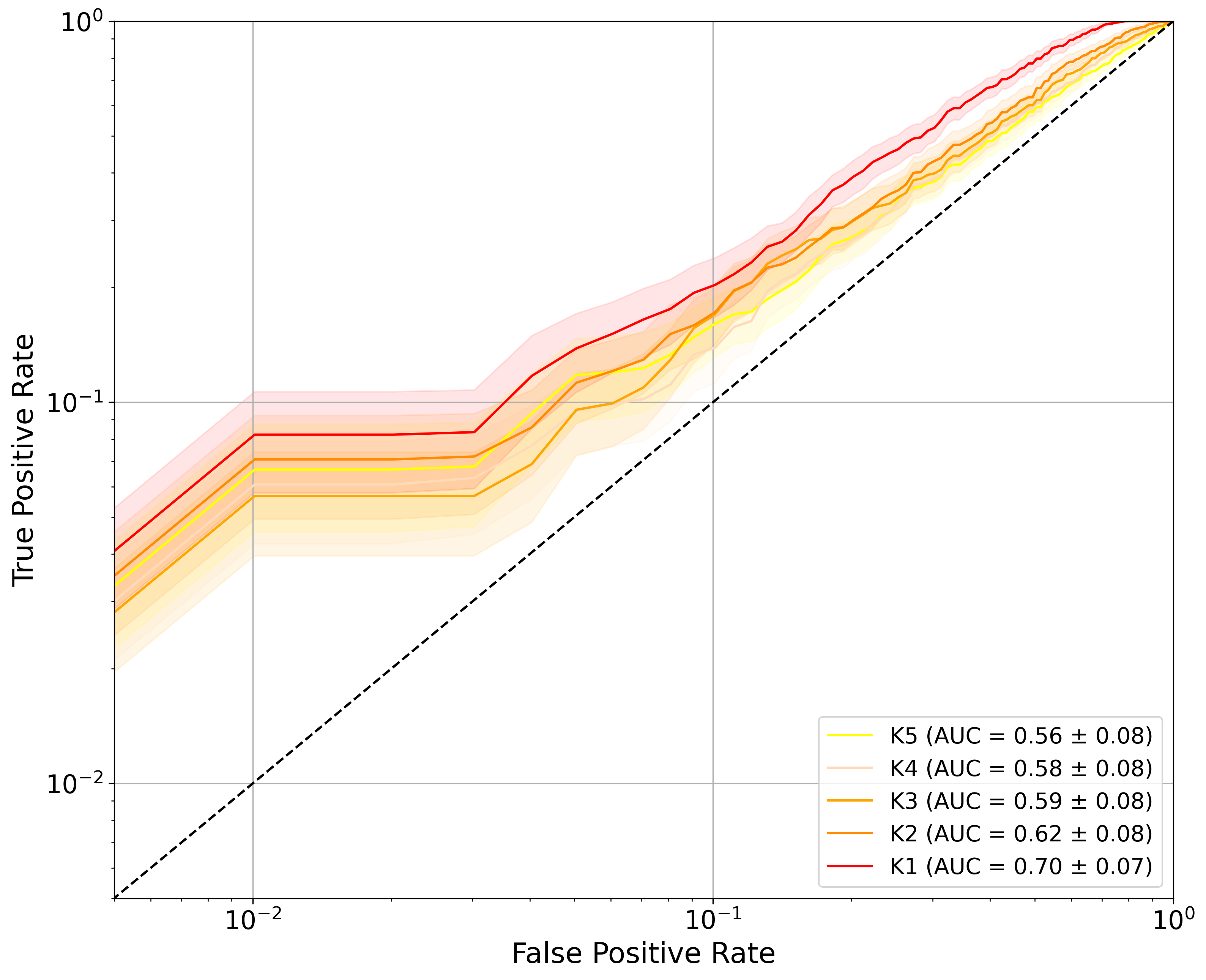}};
        \node[anchor=north west] (img3) at (8.8,3) {\includegraphics[width=0.26\linewidth]{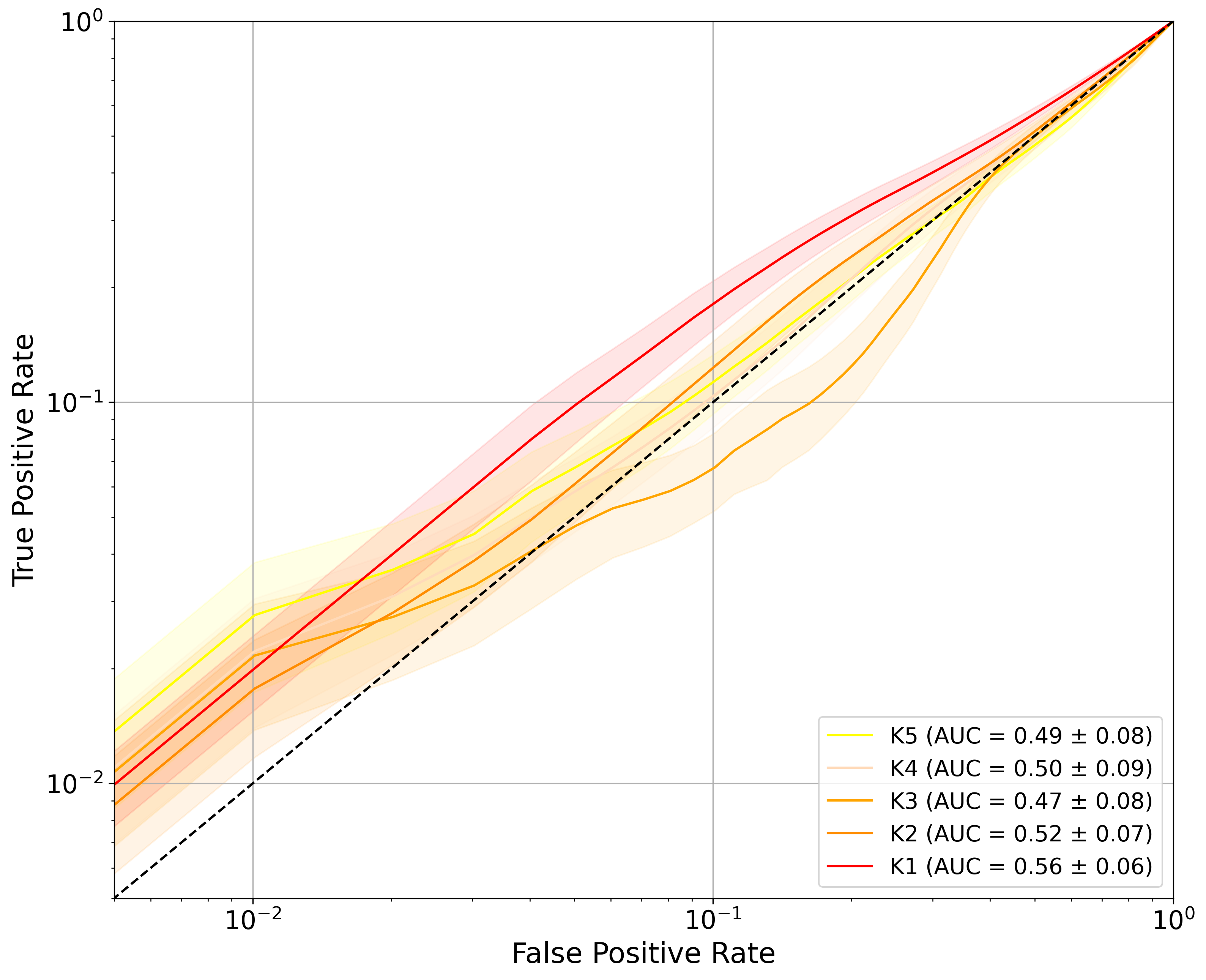}};
        \node[anchor=north west] (img4) at (13.2,3) {\includegraphics[width=0.26\linewidth]{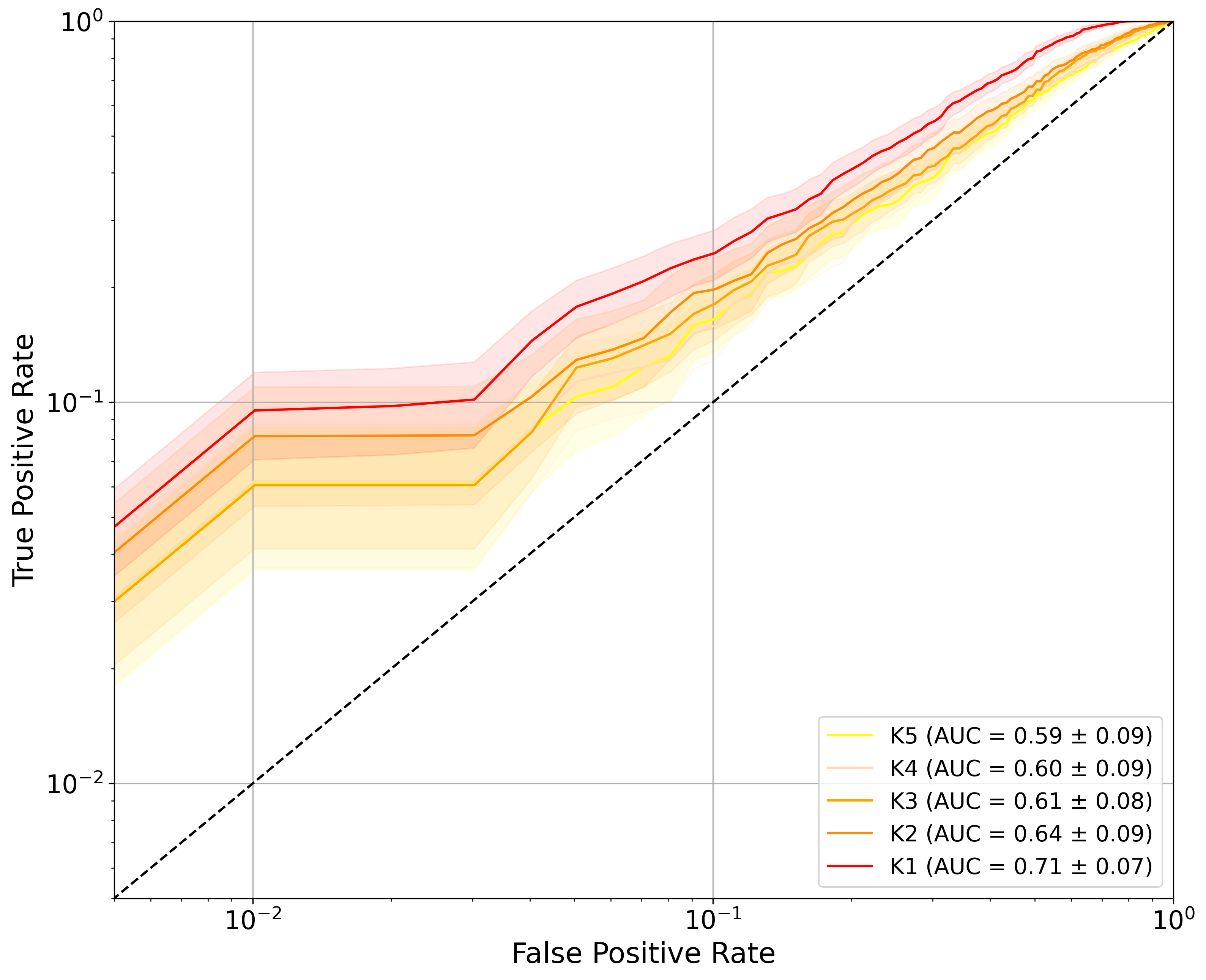}};
    \end{tikzpicture}
    \caption{Yelp: Membership Inference Attacks (MIA) AUC and ROC curves (in log-scale) using LiRA, t-WaKA, confidence, and calibrated confidence attacks for $k$ values between 1 and 5.}
    \label{fig:roc_comparison_adult}
\end{figure*}

\begin{figure*}[htbp]
    \centering
    \begin{tikzpicture}[node distance=0cm]
        \node at (2.2, 3.5) {LiRA};
        \node at (6.7, 3.5) {t-WaKA};
        \node at (11.2, 3.5) {Conf};
        \node at (15.5, 3.5) {Conf-calib};
        
        \node[anchor=north west] (img1) at (0,3) {\includegraphics[width=0.26\linewidth]{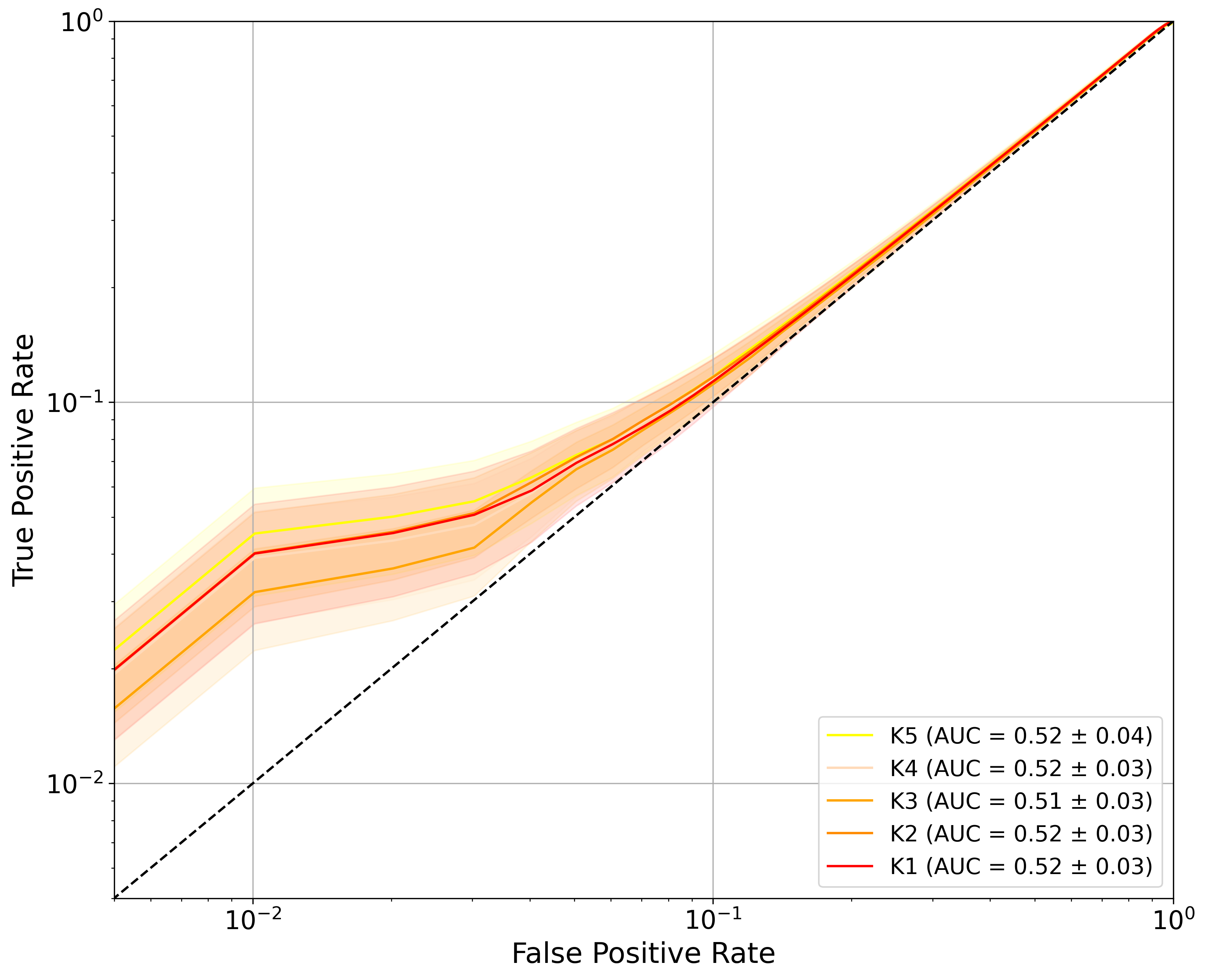}};
        \node[anchor=north west] (img2) at (4.4,3) {\includegraphics[width=0.26\linewidth]{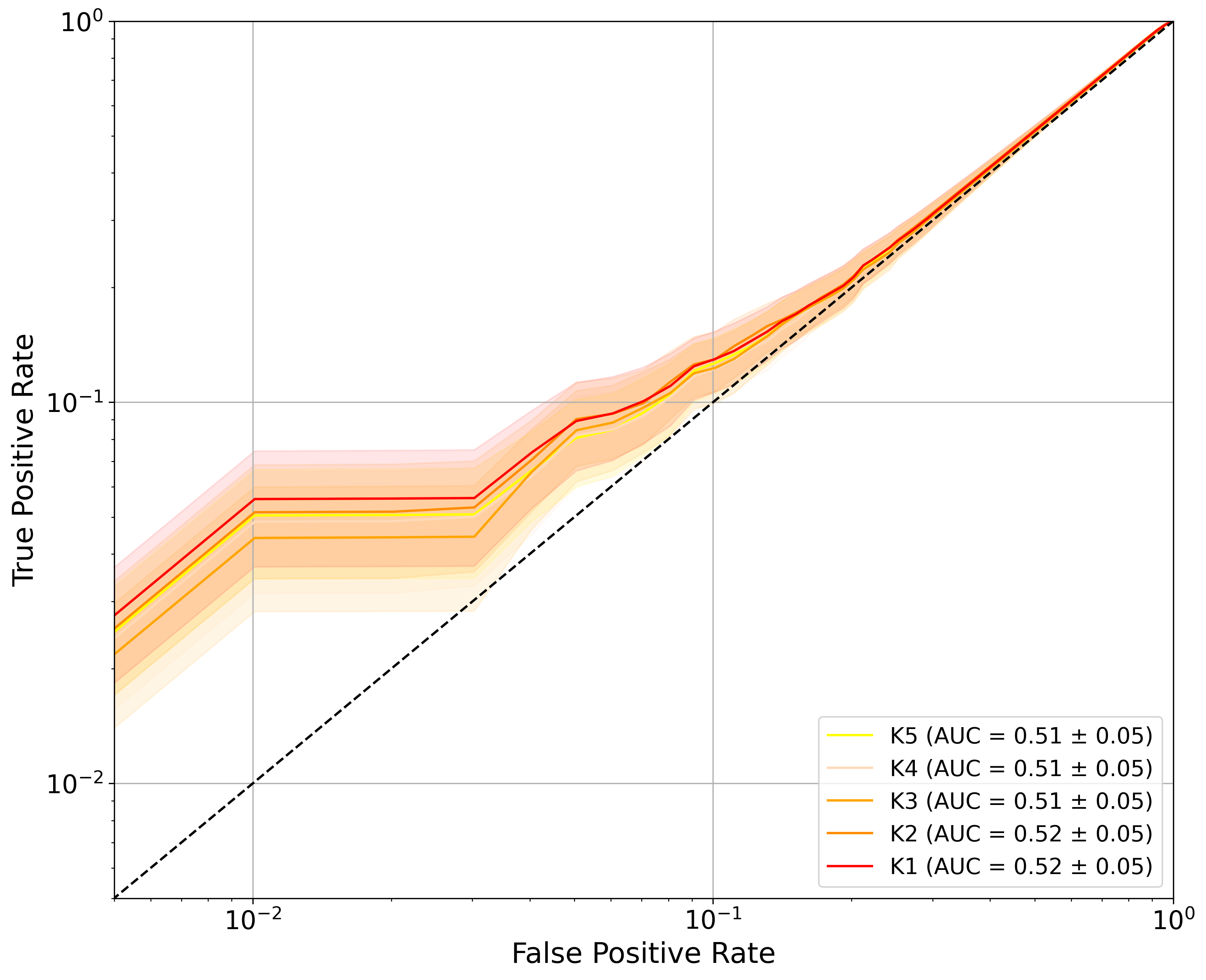}};
        \node[anchor=north west] (img3) at (8.8,3) {\includegraphics[width=0.26\linewidth]{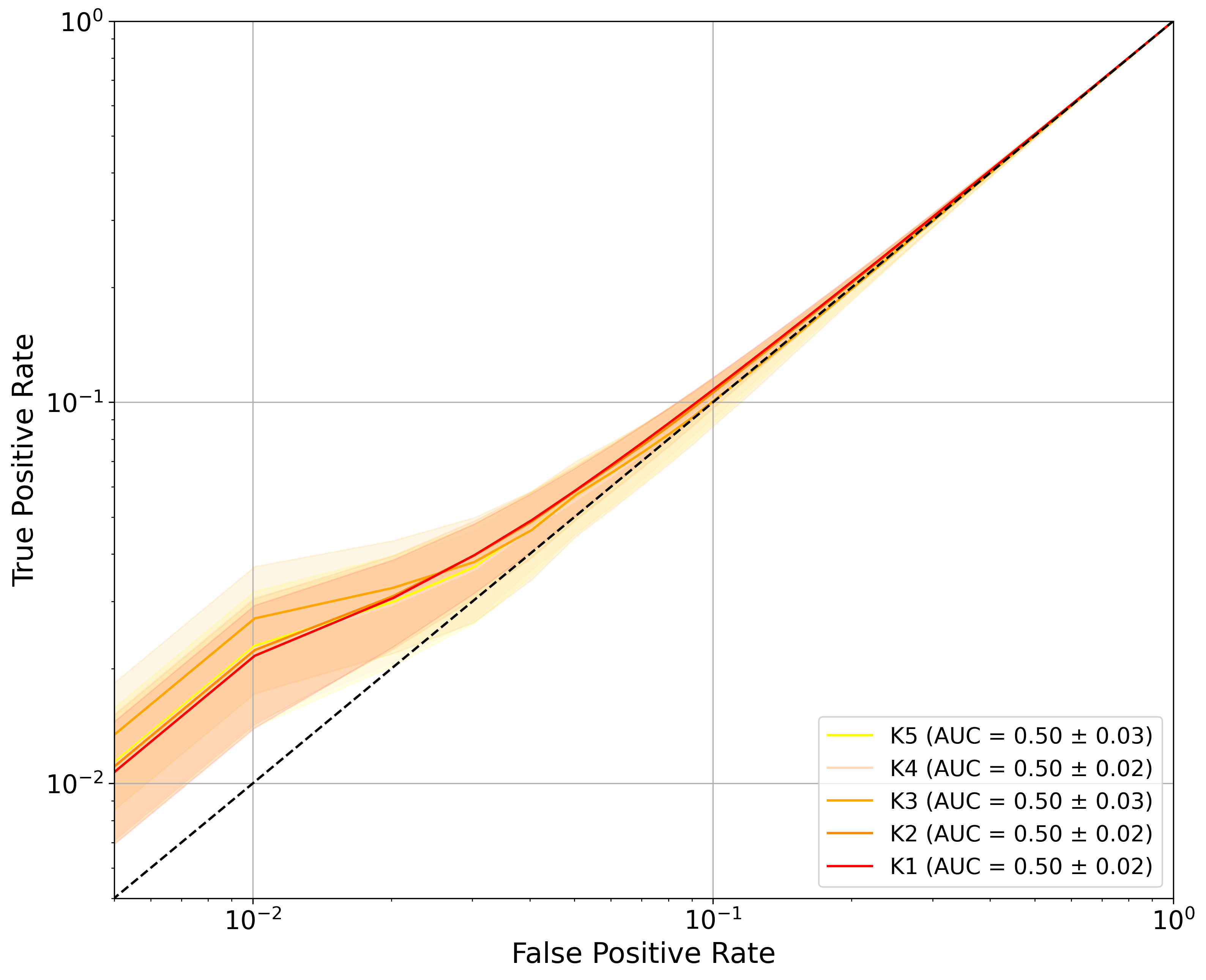}};
        \node[anchor=north west] (img4) at (13.2,3) {\includegraphics[width=0.26\linewidth]{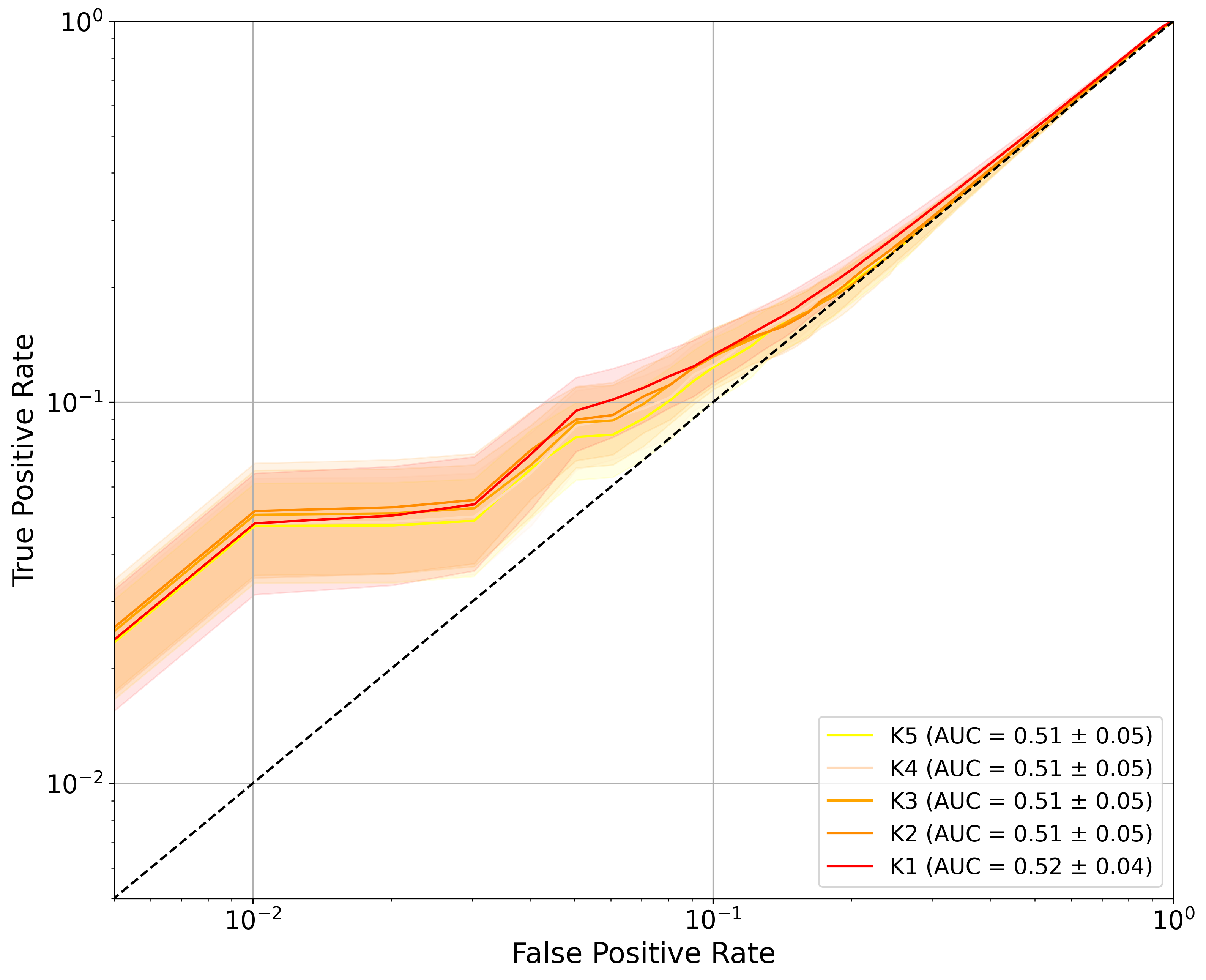}};
    \end{tikzpicture}
    \caption{Celeba: Membership Inference Attacks (MIA) AUC and ROC curves (in log-scale) using LiRA, t-WaKA, confidence, and calibrated confidence attacks for $k$ values between 1 and 5.}
    \label{fig:roc_comparison_imdb}
\end{figure*}

    \end{document}